\documentclass[twoside]{article}

\usepackage[accepted]{aistats2025}
\usepackage{graphicx} 
\usepackage{placeins}
\usepackage{bm}
\usepackage{amsmath}
\usepackage{amssymb}
\usepackage{amsthm}
\usepackage{xcolor}
\usepackage{hyperref}
\usepackage{caption}
\usepackage{subcaption}
\DeclareMathOperator*{\argmin}{arg\,min}
\usepackage{booktabs}
\usepackage{tcolorbox}
\usepackage{placeins}  
\usepackage{neuralnetwork}
\usepackage{amsfonts}
\usepackage{mathtools}
\usepackage{bbm}
\usepackage{tikz}
\usepackage{accents}
\usepackage{multirow}
\usepackage{makecell}
\usepackage{hyperref} 
\usepackage{algorithm}

\usepackage{amsthm}

\newtheorem{proposition}{Proposition} 
\newtheorem{definition}{Definition}
\newtheorem{assumption}{Assumption}
\usepackage[american]{babel}
\usepackage{xcolor}

\usepackage[round]{natbib}

\begin{document}

%

%

\twocolumn[

\aistatstitle{Score-Based Metropolis-Hastings Algorithms}

\aistatsauthor{ Ahmed Aloui \And Ali Hasan \And  Juncheng Dong \And Zihao Wu \And Vahid Tarokh}

\aistatsaddress{ Duke University \And  Morgan Stanley \And Duke University \And Duke University \And Duke University} ]

\begin{abstract}
In this paper, we introduce a new approach for integrating score-based models with the Metropolis-Hastings algorithm. Score-based diffusion models have emerged as state-of-the-art methods for generative modeling, achieving remarkable performance in accurately learning the score function from data. However, these models lack an explicit energy function, making the Metropolis-Hastings adjustment step inaccessible. Consequently, the unadjusted Langevin algorithm is often used for sampling using estimated score functions. 
The lack of an energy function prevents the application of the Metropolis-adjusted Langevin algorithm and other Metropolis-Hastings methods, restricting the use of acceptance-function-based algorithms. We address this limitation by introducing a new loss function based on the \emph{detailed balance condition}, allowing the estimation of the Metropolis-Hastings acceptance probabilities given a learned score function. We demonstrate the effectiveness of the proposed method for various scenarios, including sampling from score-based diffusion models and heavy-tail distributions.
\end{abstract}

\section{Introduction}\label{sec:intro}


Sampling from probability distributions with access only to samples is a longstanding challenge that has received significant attention in machine learning and statistics~\citep{goodfellow2020generative,kingma2013auto,ho2020denoising,dinh2016density,song2020score}.
In particular, score-based models have emerged as a powerful class of generative algorithms, underpinning state-of-the-art (SOTA) models such as DALL·E 2 and Stable Diffusion~\citep{rombach2022high}. Given samples from a target distribution, these models first estimate the score function ---the gradient of the log-density of the data distribution~\citep{hyvarinen2005estimation}---using techniques like sliced score matching~\citep{song2020sliced} and denoising score matching~\citep{vincent2011connection,song2019generative,reusmooth}. Subsequently, they leverage the learned score function and score-based sampling methods, primarily the Unadjusted Langevin Algorithm (ULA), to sample from the target distribution~\citep{grenander1994representations,roberts1996exponential}.


In parallel, a wealth of literature focuses on sampling from densities known only up to a normalizing constant~\citep{hastings1970monte,wang2024cyclical}.
A particular instance of this is the Metropolis-Hastings (MH) method, which provides a condition for accepting or rejecting new samples based on knowledge of a function proportional to the density~\citep{robert2004metropolis}. 
The MH method construct a Markov chain based on a proposal distribution and an acceptance function, ensuring convergence when the detailed balance condition is satisfied. 
Various MH algorithms have been introduced, defined by their choice of proposal distributions, such as the Random Walk Metropolis (RW) algorithm~\citep{metropolis1953equation}, the Metropolis-adjusted Langevin algorithm (MALA)~\citep{roberts1996exponential,roberts2002langevin}, and the preconditioned Crank-Nicolson (pCN) algorithm~\citep{hairer2014spectral}. While these methods are theoretically well-grounded, they require knowledge of the unnormalized density, making them impractical when only samples or an estimated score function are available.  

A natural observation arises when considering score-based modeling: despite the extensive research on MH sampling, its integration with score-based methods remains largely unexplored. The key technical challenge lies in the absence of an unnormalized density when only the score function (or its estimation) is available.
This work proposes a unifying framework that bridges this gap, enabling the application of MH algorithms using only samples and a learned score function. Consequently, this framework can leverage the strengths of both MH and score-based methods. Our analysis is motivated by two key questions:
\begin{tcolorbox}
A). Can we estimate the acceptance function of an MH algorithm \emph{solely} from an (estimated) score function and samples?

B). Does including an MH adjustment step \emph{improve sample quality} for score-based models?
\end{tcolorbox}
We answer these questions affirmatively by developing an approach to estimate an acceptance function from data. Our contributions are as follows:
\begin{itemize}
    \item We introduce Score Balance Matching, a method for estimating the acceptance function in the MH algorithm using only an estimated score function.
    \item We use the learned acceptance function to incorporate MH steps into score-based samplers like ULA, enhancing the quality and diversity of generated samples. 
    \item We evaluate our method's effectiveness and its robustness to hyperparameters by comparing it to ULA across diverse scenarios. 
\end{itemize}

\section{Related Work} 
A significant area of related work focuses on estimating a score function and developing efficient sampling strategies based on it. 
While most approaches aim to accelerate convergence to the target distribution (e.g. through improved numerical integrators), we consider our work as a parallel approach, incorporating an acceptance function to refine samples. 
This work lies at the intersection of the literature on score-based models and Metropolis-Hastings methods. We first review recent advances in both areas.

\paragraph{Score-Based Models.}
Many successful generative models sample from an estimated score function~\citep{song2019generative, song2020score}. 
These methods are trained using by minimizing the Fisher divergence between the true and predicted score~\citep{hyvarinen2005estimation}. 
A significant research effort is concerned with understanding the efficacy of these sampling algorithms, with some papers investigating the asymptotic error of score-based diffusion models, as in~\citep{chen2023sampling, pmlr-v235-zhang24bv}. 
Different sampling schemes have been proposed to improve sampling speed and quality~\citep{dockhorn2021score, lu2022dpm, chen2024adaptive}. 
These methods have relied heavily on various score-based samplers such as ULA or the reverse-time SDE of a diffusion process~\citep{song2020score}.

\paragraph{Metropolis-Hastings.}
Metropolis-Hastings algorithms are a class of MCMC algorithms that use an acceptance function to sample from target distributions~\citep{metropolis1953equation, hastings1970monte}. 
These algorithms generate a sequence of samples by proposing a candidate state from a proposal distribution and accepting or rejecting the candidate based on an acceptance criterion that ensures convergence to the target distribution. 
The efficiency of MH algorithms heavily depends on the choice of the proposal distribution, and significant research effort has focused on designing efficient proposal mechanisms~\citep{rosenthal2011optimal,song2017nice,titsias2019gradient,davies23a,lew23a}.

Recent developments in MH algorithms have explored the use of adaptive proposals to improve convergence rates and reduce autocorrelation in the generated samples~\citep{andrieu2008tutorial, haario2001adaptive}. These adaptive MH methods adjust the proposal distribution during the sampling process to better capture the geometry of the target distribution, leading to faster and more accurate sampling~\citep{roberts2009examples, brooks2011handbook,hirt2021entropy,biron2024automala}.

\begin{figure*}[ht!]
    \centering
    \begin{subfigure}[b]{0.23\textwidth}
        \includegraphics[width=\textwidth]{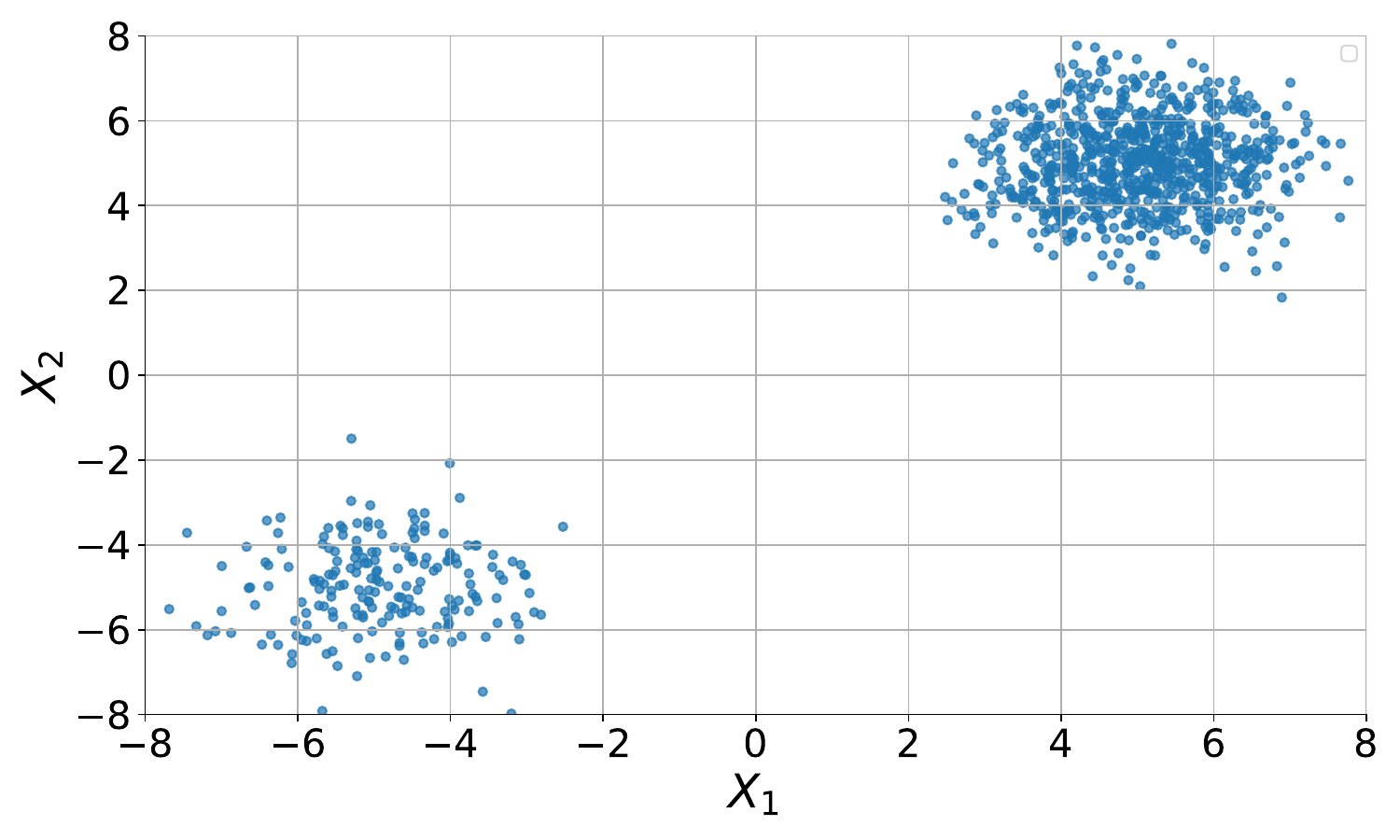}
        \caption{Original samples, weights \((0.8, 0.2)\).}
    \end{subfigure}
    \hfill
    \begin{subfigure}[b]{0.23\textwidth}
        \includegraphics[width=\textwidth]{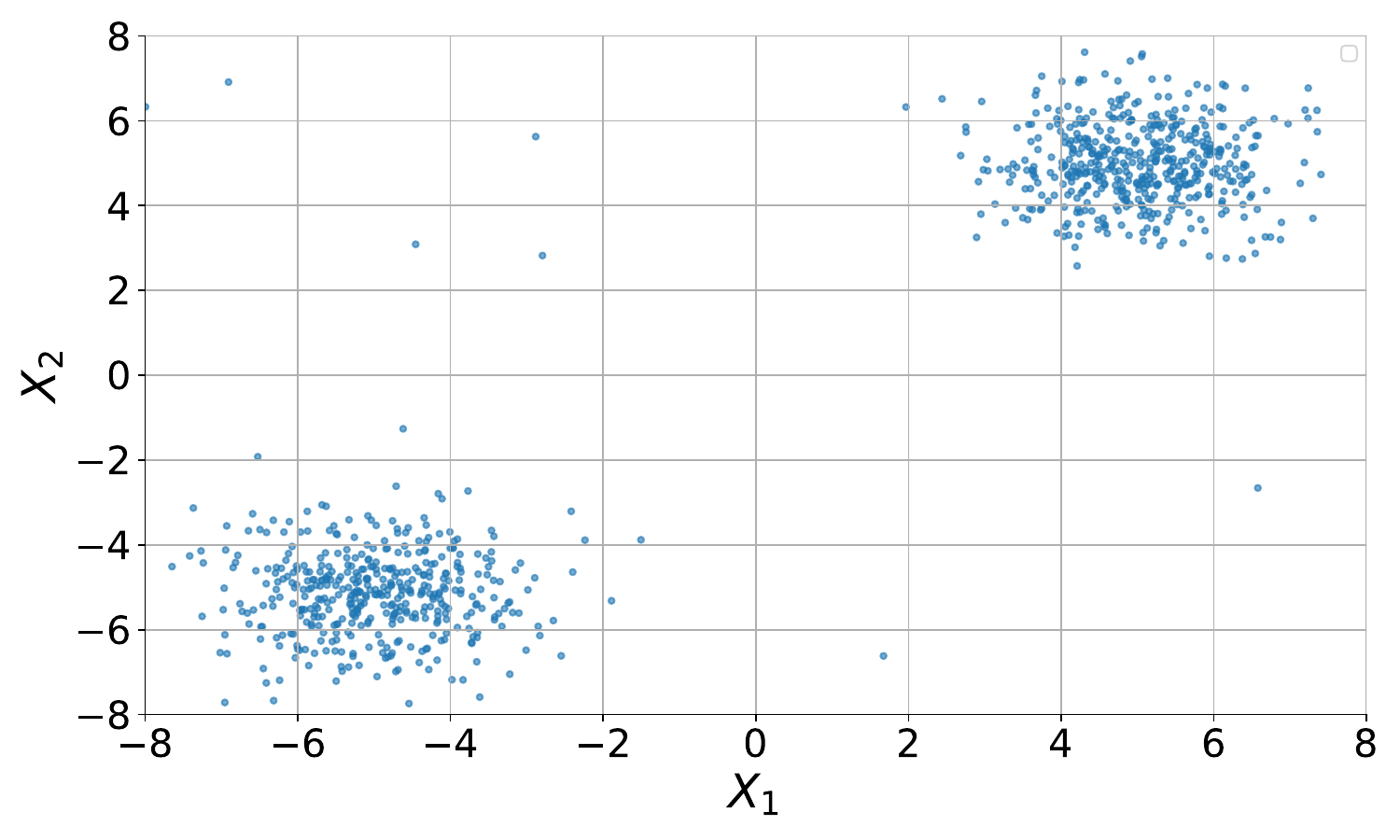}
        \caption{ULA samples, weights \((0.52, 0.48)\).}
    \end{subfigure}
    \hfill
    \begin{subfigure}[b]{0.23\textwidth}
        \includegraphics[width=\textwidth]{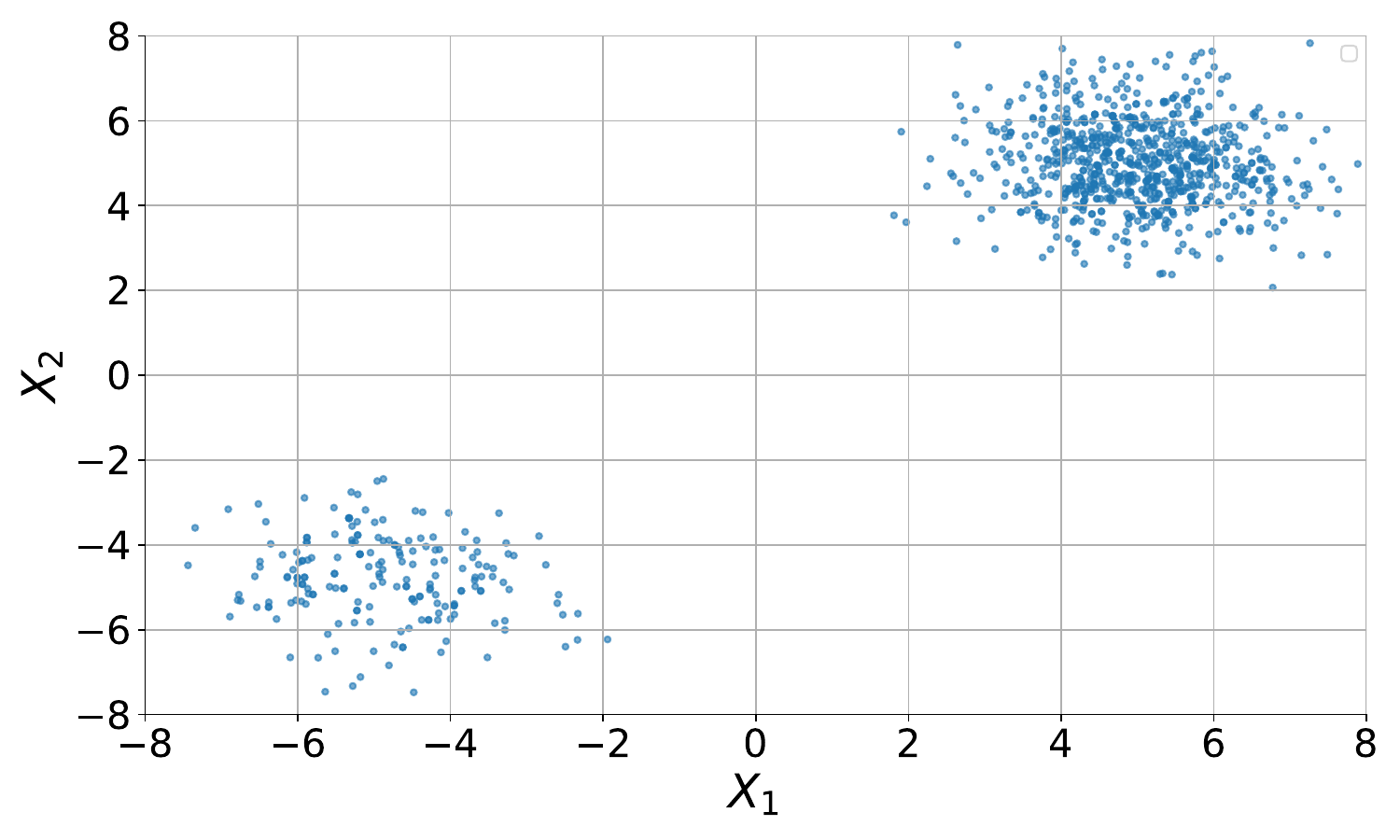}
        \caption{RW samples, weights \((0.79, 0.21)\).}
    \end{subfigure}
    \hfill
    \begin{subfigure}[b]{0.23\textwidth}
        \includegraphics[width=\textwidth]{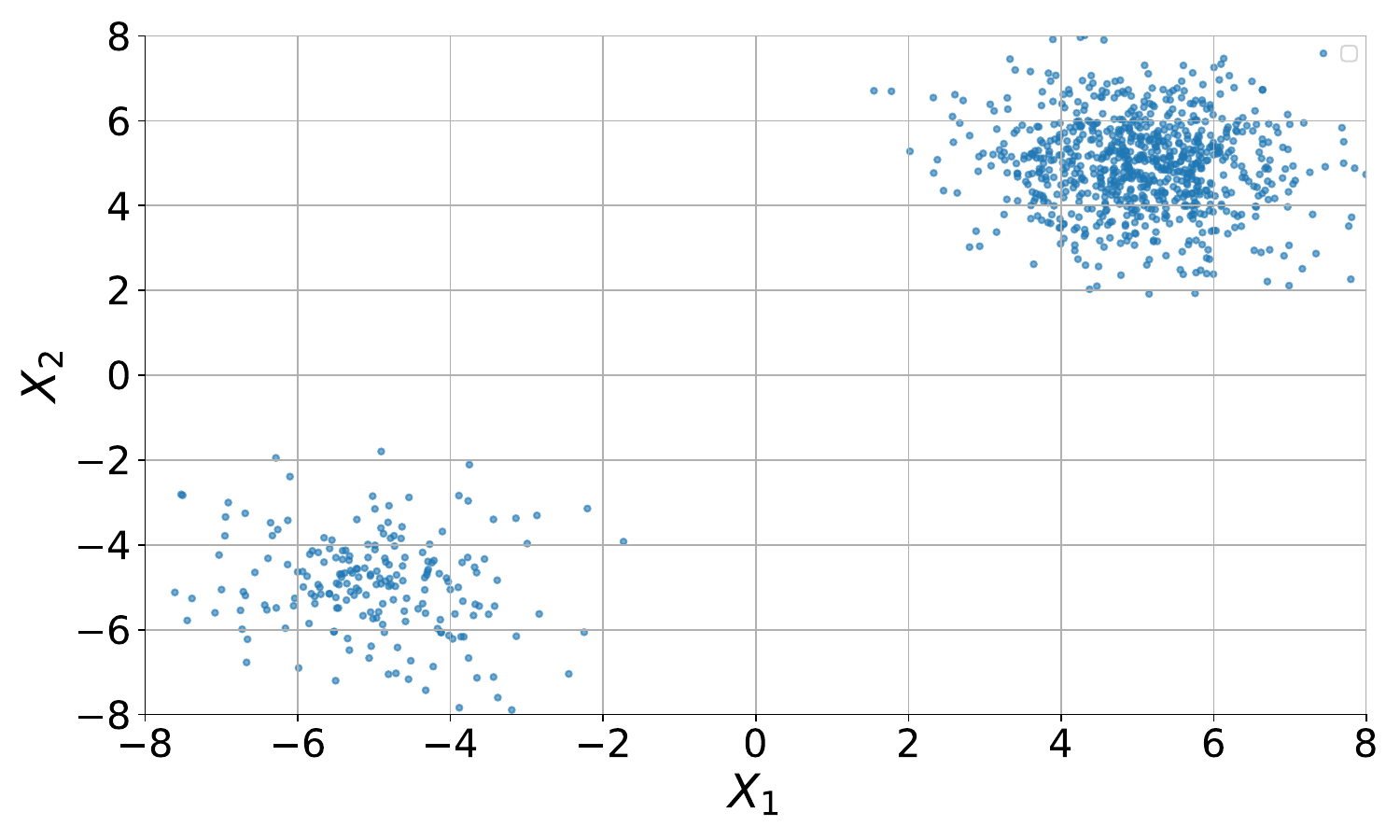}
        \caption{MALA samples, weights \((0.79, 0.21)\).}
    \end{subfigure}
    
    \caption{Comparison of sampling methods for a mixture of two Gaussians: (a) Original distribution, (b) ULA with the true scores, (c) RW with the true distribution, and (d) MALA with true scores and distribution. The plots demonstrate the impact of slow mixing between modes, highlighting the challenges in low-density regions. We report the empirical weights of the mixture given by each sampling method.}
    \label{fig:slow_mixing_examples}
\end{figure*}

\section{Score-Based Metropolis-Hastings}
In this section, we first review score matching and MH algorithms in Sections~\ref{sec:sm} and~\ref{sec:mh} before we introduce Score Balance Matching in Section~\ref{sec:sbm}, a novel approach for learning the acceptance function from the score function. Let \( X \sim p \) be a random variable, with  support \(\mathrm{supp}(X) = \mathcal{X}\subset \mathbb{R}^d \) and \( p \) be its probability density function. 
We assume that \( p\) is unknown, and only samples of \( X \) are observed. 
Denote these observed samples by \( \{x^{\left(i\right)}\}_{i=1}^N \), where \( N \) is the number of observed samples.

\subsection{Score Matching}\label{sec:sm}
Score-based models are grounded in score matching~\citep{hyvarinen2005estimation}, which minimizes the Fisher divergence between the estimated and true gradients of the log data distribution, defined as follows.

\begin{definition}[Fisher Divergence]
    The Fisher divergence \(D_{\nabla}(p_1 \| p_2)\) between two probability distributions \( p_1 \) and \( p_2 \) is defined as:
    \[
     D_{\nabla}(p_1 \| p_2) = \mathbb{E}_{x \sim p_1} \left[ \|\nabla_x \log p_1(x) - \nabla_x \log p_2(x) \|^2 \right].
    \]
\end{definition}
Therefore, given a class of hypothesis function $\mathcal{S} \subset \{s: \mathcal{X} \rightarrow \mathbb{R}^d\}$, score matching aims to find $s^*\in \mathcal{S}$ that minimizes the Fisher divergence to the data distribution $p$: 
\begin{equation}
\label{eq:fisher_loss}
s^* \in \argmin_{s\in \mathcal{S}} \mathbb{E}_{x \sim p} \left[ \|\nabla_x \log p(x) - s(x) \|^2 \right]
\end{equation}
Since the score function is not directly observed, Hyvärinen proved that minimizing in the loss funcition in Equation \ref{eq:fisher_loss} is equivalent to minimizing the following loss function:
\begin{equation}
\label{eq:hyv_loss}
    \mathbb{E}_{x \sim p} \left[ \frac{1}{2} \| s(x)\|^2 + \operatorname{tr}\left(\nabla_x s(x)\right) \right].
\end{equation}
Several variants of this loss have been proposed to circumvent the cumbersome estimation of the Hessian term in Equation \ref{eq:hyv_loss}, including sliced score matching and denoising score matching, defined as follows:
\begin{definition}[Sliced Score Matching]
    Sliced score matching (SSM) projects the score function onto random directions \( v \sim \mathcal{N}(0, I_d) \):
    \begin{equation}
        \mathbb{E}_{x \sim p, v \sim \mathcal{N}(0, I_d)} \left[ \frac{1}{2} \| s(x) \|^2 + v^\top \nabla_x s(x) v \right].
    \end{equation}
\end{definition}

\begin{definition}[Denoising Score Matching]
     Given a noise distribution \( p_\sigma(\tilde{x} | x) \), the loss function is:
    \begin{equation}
        \mathbb{E}_{x \sim p, \tilde{x} \sim p_\sigma(\tilde{x} | x)} \left[ \|\nabla_{\tilde{x}} \log p_\sigma(\tilde{x} | x) - s(\tilde{x})\|^2 \right].
    \end{equation}
\end{definition}

\subsection{Metropolis-Hastings}\label{sec:mh}

MH involves an \emph{acceptance function} \( a: \mathcal{X} \times \mathcal{X} \rightarrow [0,1] \), which defines the probability \( a(x', x) \) of transitioning from the current \( x \in \mathcal{X} \) to a proposal \( x' \in \mathcal{X} \). 
The common choice of the acceptance function is given by:
\begin{equation}
\label{eqn:ratio_acceptance}
 a(x', x) = \min \left\{ 1, \frac{p(x') q(x | x')}{p(x) q(x' | x)} \right\}   
\end{equation}
where \( p(x) \) is the target distribution and \( q(x | x') \) is the proposal distribution. The MH algorithm proceeds as follows:
\begin{itemize}
    \item \textbf{Initialize:} Set \( x_1 \in \mathcal{X} \), number of iterations \( T \), and proposal distribution \( q(x'|x) \).
    \item \textbf{For} \( t = 1, \dots, T \):
    \begin{itemize}
        \item Propose \( x' \sim q(x'|x_t) \).
        
        \item Compute \(a(x', x_t)\) in~\eqref{eqn:ratio_acceptance} as the acceptance probability.
        
        \item Draw \( u \sim \text{Uniform}(0, 1) \). If \( u \leq a(x', x_t) \), accept \( x_{t+1} = x' \), otherwise set \( x_{t+1} = x_t \).
    \end{itemize}
\end{itemize}
Note that for computing \(a(x', x_t)\) in Equation~\eqref{eqn:ratio_acceptance}, a function proportional to $p(x)$ is needed.

\paragraph{Detailed Balance.} Although the acceptance function is often defined by Equation~\eqref{eqn:ratio_acceptance}, a sufficient condition for ensuring convergence to the target distribution is that the acceptance ratio satisfies the detailed balance condition:
\begin{equation}
\label{eqn:detailed_balance}
\forall x,x' \in \mathcal{X}, \quad \frac{a(x', x)}{a(x, x')} = \frac{p(x') q(x \mid x')}{p(x) q(x' \mid x)}.
\end{equation}
In particular, any acceptance function satisfying the detailed balance condition above ensures that the Markov chain has the target distribution $p(x)$ as its unique stationary distribution. Note that the standard acceptance function in Equation~\eqref{eqn:ratio_acceptance} is one specific solution that satisfies this condition. 

\begin{figure*}[t]
    \centering
    \begin{subfigure}[b]{0.32\textwidth}
        \includegraphics[width=\textwidth]{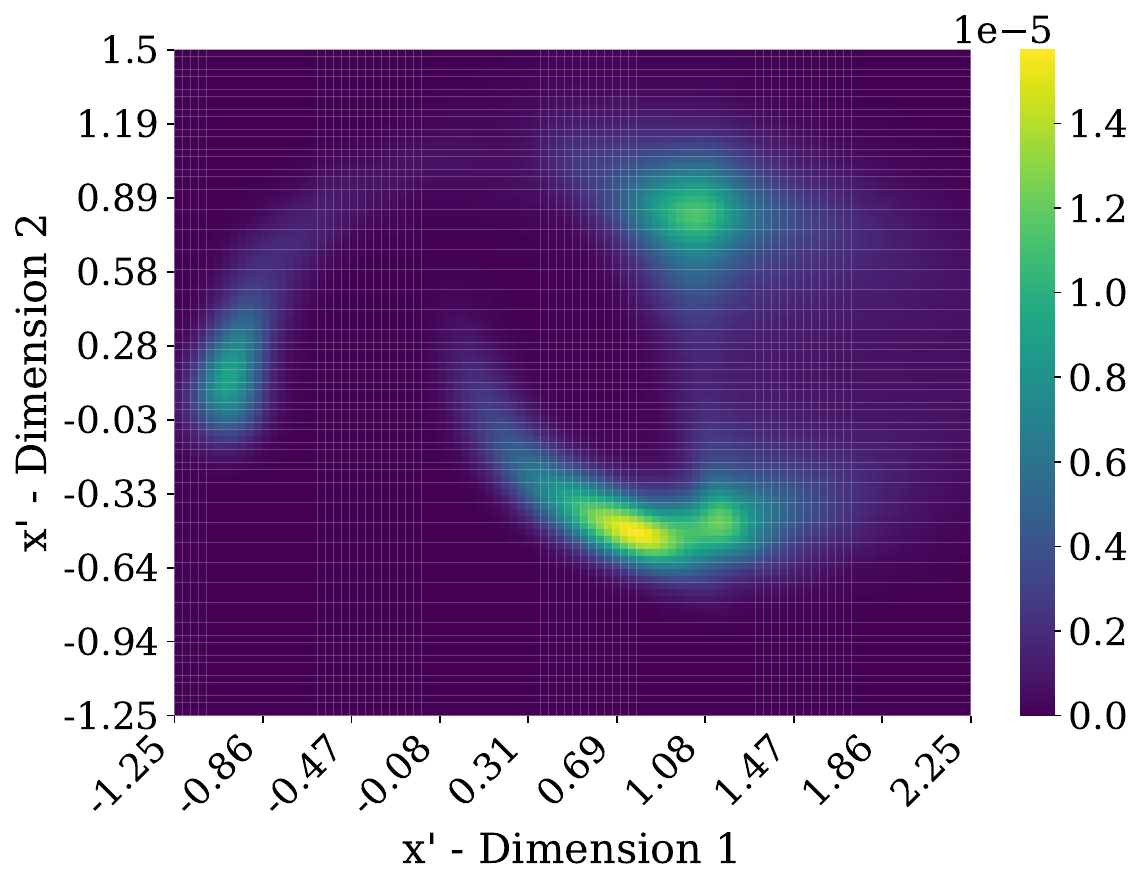}
        \caption{$\lambda = 0$}
    \end{subfigure}
    \hfill
    \begin{subfigure}[b]{0.32\textwidth}
        \includegraphics[width=\textwidth]{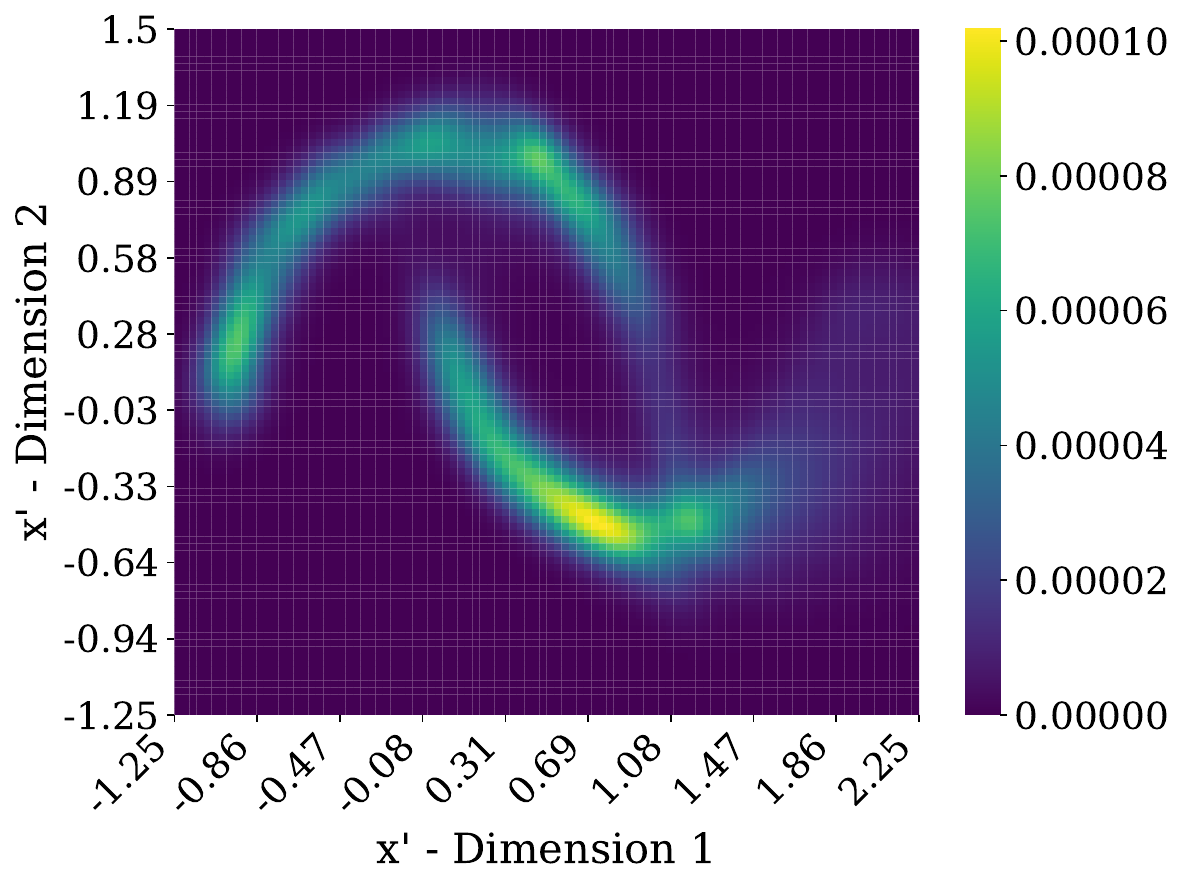}
        \caption{$\lambda = 0.1$}
    \end{subfigure}
    \hfill
    \begin{subfigure}[b]{0.32\textwidth}
        \includegraphics[width=\textwidth]{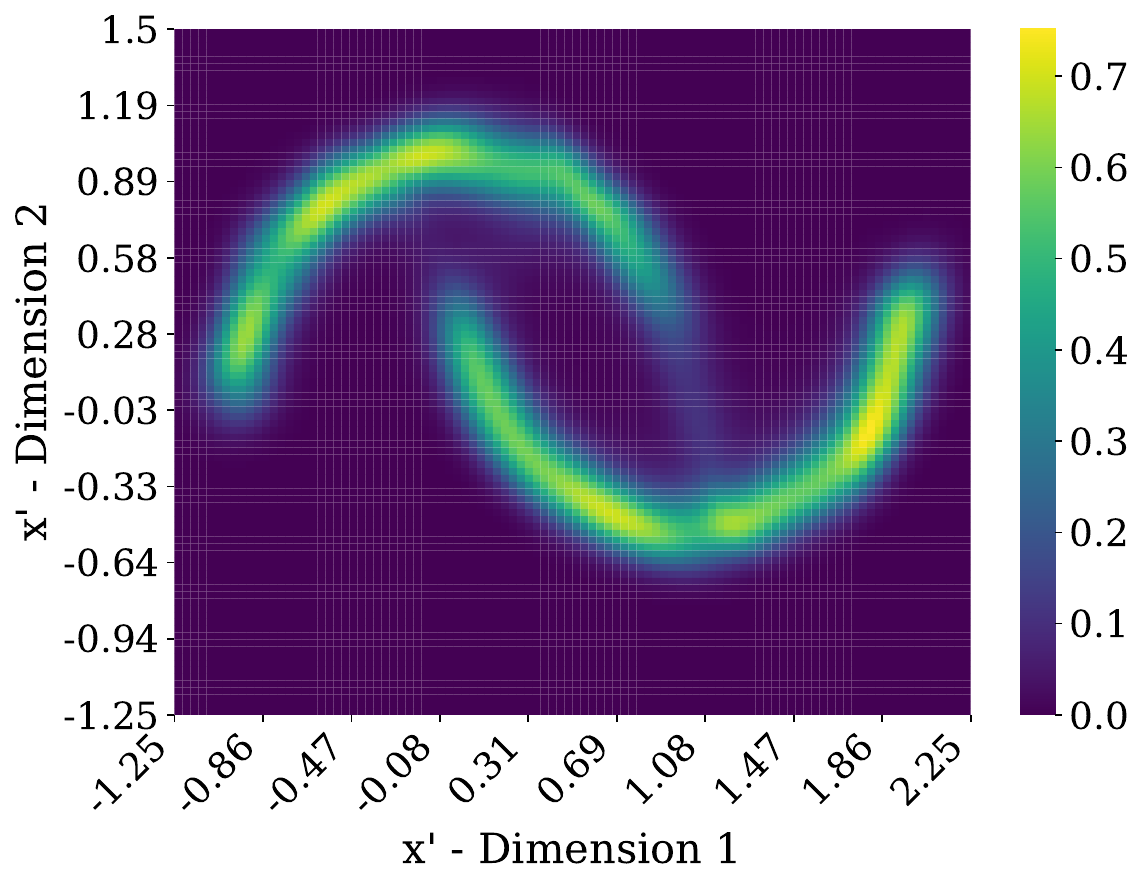}
        \caption{$\lambda = 1.0$ }
    \end{subfigure}
    \caption{Visualizing the acceptance probabilities for different values of $x'$ for a given initial state $x=(0.0,0.0)$. We plot the results for training an acceptance network with three different regularization values $\lambda\in\{0,0.1,1.0\}$}.
    \label{fig:moons_entropy}
\end{figure*}

\subsection{Score Balance Matching}\label{sec:sbm}
We now introduce our new objective function called Score Balance Matching (SBM). Let $q$ be a proposal distribution and let $\mathcal{A}$ be the set of valid acceptance functions for a given proposal $q$ and target $p$. 
\[
 \mathcal{A} = \{a: \mathcal{X} \times \mathcal{X} \rightarrow [0,1] \;|\; \text{\(a\) satisfies~\eqref{eqn:detailed_balance} almost surely}\}.
\]
The SBM objective is defined as follows:
\begin{align}
\label{eqn:ideal}
    \mathcal{L}(a) & = \mathbb{E}_{x \sim p,x'\sim q(\cdot|x)}\bigg[ \left\| \nabla \log a(x', x) - \nabla \log a(x, x') \right. \nonumber \\
    & \quad - \left. \nabla \log p(x') + \nabla \log p(x) \right. \nonumber \\
    & \quad  \left. - \nabla \log q(x \mid x') + \nabla \log q(x' \mid x) \right\|^2 \bigg],
\end{align}
where $\nabla$ denotes the gradient with respect to \(x, x' \in \mathbb{R}^d\), i.e., \(\nabla = \nabla_{x, x'}\). The terms \( \nabla \log p(x) \) and \( \nabla \log p(x') \) represent the score functions at \( x \) and \( x' \), which can be approximated with a score model (e.g., obtained from score matching).
We can express these score functions as:
\[
\nabla \log p(x) = \nabla_{x,x'} \log p(x) = \left( \nabla_x \log p(x), \mathbf{0} \right)^\top,
\]
and similarly,
\[\nabla \log p(x') = \left( \mathbf{0}, \nabla_{x'} \log p(x') \right)^\top,
\]
where \( \mathbf{0} \in \mathbb{R}^d \) is the zero vector. 
Finally, we note that the gradients of the proposal distribution \( q(x'|x) \) are typically known by design when the proposal function is specified. 

\begin{proposition}
    We have that, for every function $a:\mathcal{X} \times \mathcal{X} \rightarrow [0,1]$
\begin{equation}
    \mathcal{L}(a) = 0 \iff a \in \mathcal{A}.
\end{equation}
\end{proposition}
Therefore, if we minimize SBM to zero, we are guaranteed to have a valid acceptance function.   
In practice, the acceptance function \(a\) can be parameterized using an expressive class of functions and optimized to minimize SBM. 
Since we estimate the acceptance function from finite samples, we establish a generalization bound to quantify the approximation error when learning \( a(x', x) \) using a function class \( \mathcal{F} \) (e.g., neural networks). We start by making the following assumptions.

\begin{assumption}\label{ass:compact}$\mathcal{X}$ is compact.
\end{assumption}
\begin{assumption}\label{ass:cont}$\nabla \log p$ and  $\nabla \log q$ are continuous on $\mathcal{X}\times\mathcal{X}$.
\end{assumption}
\begin{assumption}\label{ass:hyp_bound} $\forall a \in \mathcal{F}, \nabla \log a$ is continuous on $\mathcal{X}\times\mathcal{X}$.    
\end{assumption}
We note that Assumption~\ref{ass:hyp_bound} can be achieved by design. However, for many applications, Assumption~\ref{ass:compact} may not hold. To this end, we propose to employ the fact that the score function is scale invariant:
\[
\nabla_x \log p(x) = \nabla_x \log p^*(x), \quad \forall x \in \mathcal{X}.
\]
where, \(\mathcal{X} \subset \mathbb{R}^d\) is chosen as a large compact subset of the support and $p^*$ is a truncated density:
\[
p^*(x) = 
\frac{p(x)}Z  \mathbbm{1}\left(x \in \mathcal{X}\right),
\]
where \(Z = \int_{x \in \mathcal{X}} p(x) dx\). In other words, by replacing the true target $p$ with its truncated version $p^*$ for SBM, Assumption~\ref{ass:compact} can also hold. Next we show the finite sample guarantee.

\begin{proposition}[Finite-Sample Generalization Bound]
    Let \( \widehat{\mathcal{L}_N}(a) \) be the empirical SBM loss computed over \( N \) i.i.d. training samples. Then, under Assumptions~\ref{ass:compact},~\ref{ass:cont}, and~\ref{ass:hyp_bound}, we have that with probability at least \( 1 - \delta \),
    \begin{equation}
        \sup_{a \in \mathcal{F}} \left| \mathcal{L}(a) - \widehat{\mathcal{L}_N}(a) \right| \leq 2 \mathcal{R}_N(\mathcal{F}) +\mathcal{O}\left(\sqrt{\frac{\log(1/\delta)}{N}}\right) ,
    \end{equation}
    where \( \mathcal{R}_N(\mathcal{F}) \) is the empirical Rademacher complexity of class \( \mathcal{F} \).
\end{proposition}

For the scope of this paper, we will instantiate our method on three representative MH algorithms, specified by their proposal functions: the \emph{Random Walk Metropolis-Hastings} (RW), the \emph{Metropolis-Adjusted Langevin Algorithm} (MALA), and the \emph{Preconditioned Crank-Nicolson} (pCN) algorithm. We next briefly review these methods.

\textbf{RW:} The proposal function for RW is a Gaussian centered at the current state, i.e., \( q(x'|x) = \mathcal{N}\left(x, \sigma^2 I \right) \).

\textbf{MALA:} 
The proposal distribution in MALA uses both the current state and the gradient of the log-probability, i.e., \( q(x'|x) = \mathcal{N}\left(x + \frac{\varepsilon^2}{2} \nabla \log p(x), \varepsilon^2 I\right ) \).

\textbf{pCN:} The proposal function for pCN is given by \( x' = \sqrt{1 - \beta^2} x + \beta \xi \), where \(\xi \sim \mathcal{N}(0, I)\). 

When the acceptance function is learned using SBM, we refer to these methods as \textbf{Score RW}, \textbf{Score MALA}, and \textbf{Score pCN}, respectively.
While we illustrate our results on these three MH algorithms, we note that \emph{the proposed framework is general} and can be applied to other proposal functions. In Appendix~\ref{sec:taylor}, we include another approach to approximate the acceptance function. 
\begin{figure*}[t!]
    \centering
    \begin{subfigure}[b]{0.18\textwidth}
        \includegraphics[width=\textwidth]{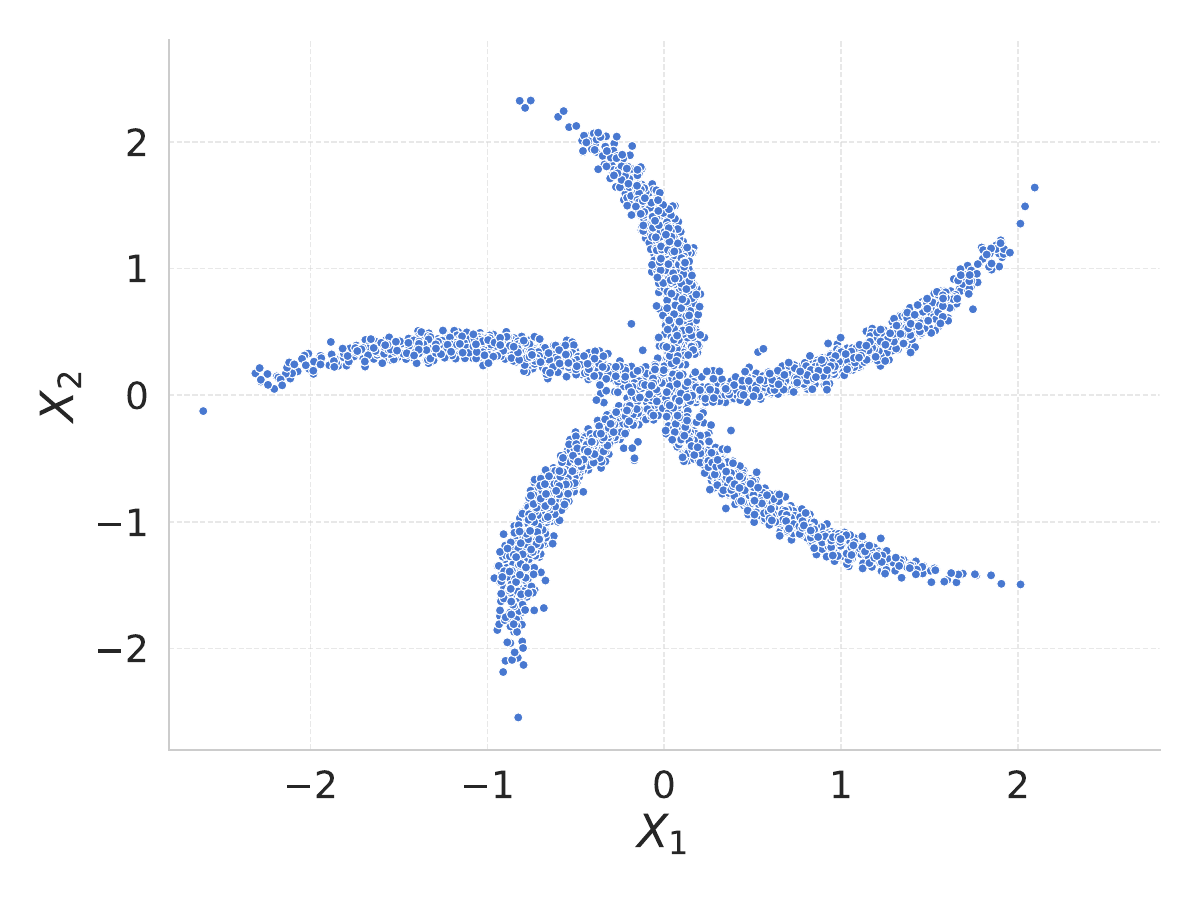}
        \caption{Original samples.}
    \end{subfigure}
    \hfill
    \begin{subfigure}[b]{0.18\textwidth}
        \includegraphics[width=\textwidth]{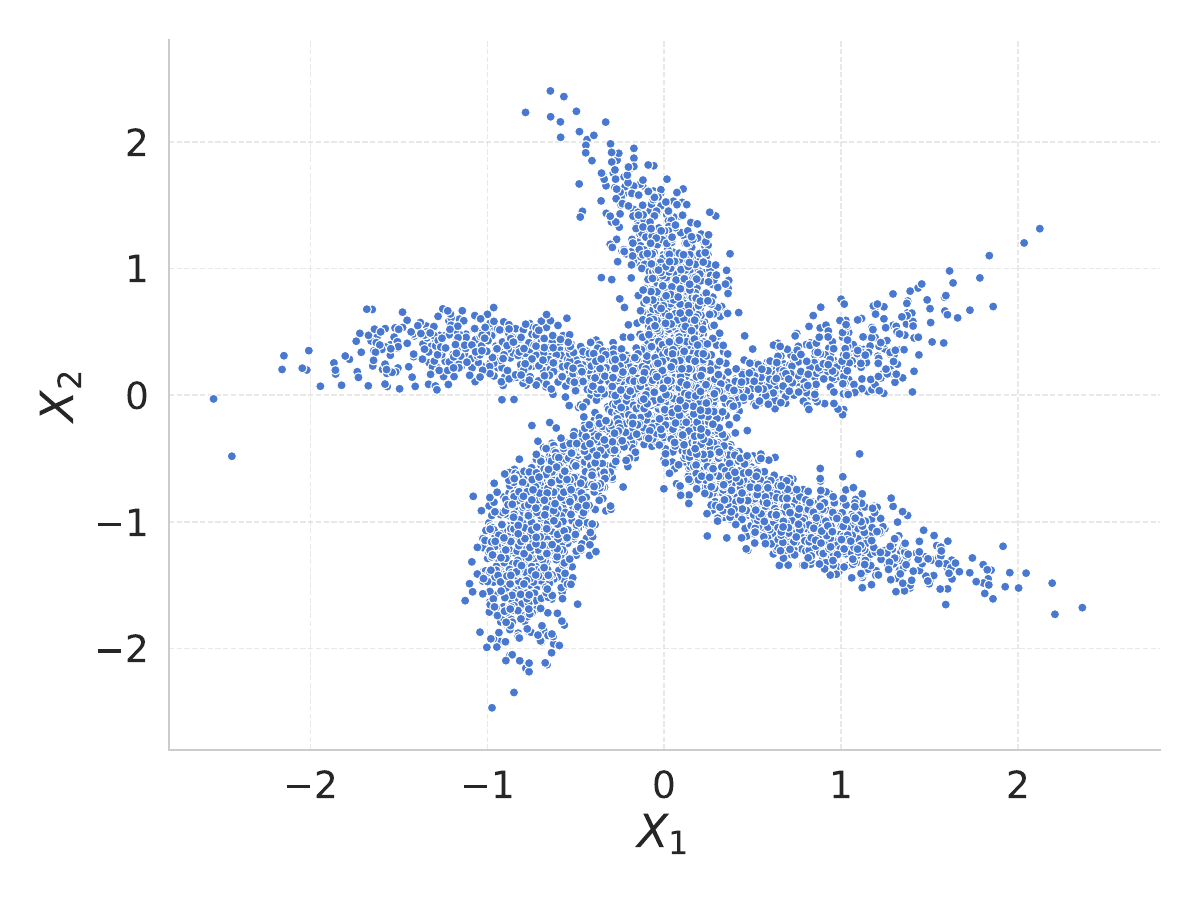}
        \caption{ULA.}
    \end{subfigure}
    \hfill
    \begin{subfigure}[b]{0.18\textwidth}
        \includegraphics[width=\textwidth]{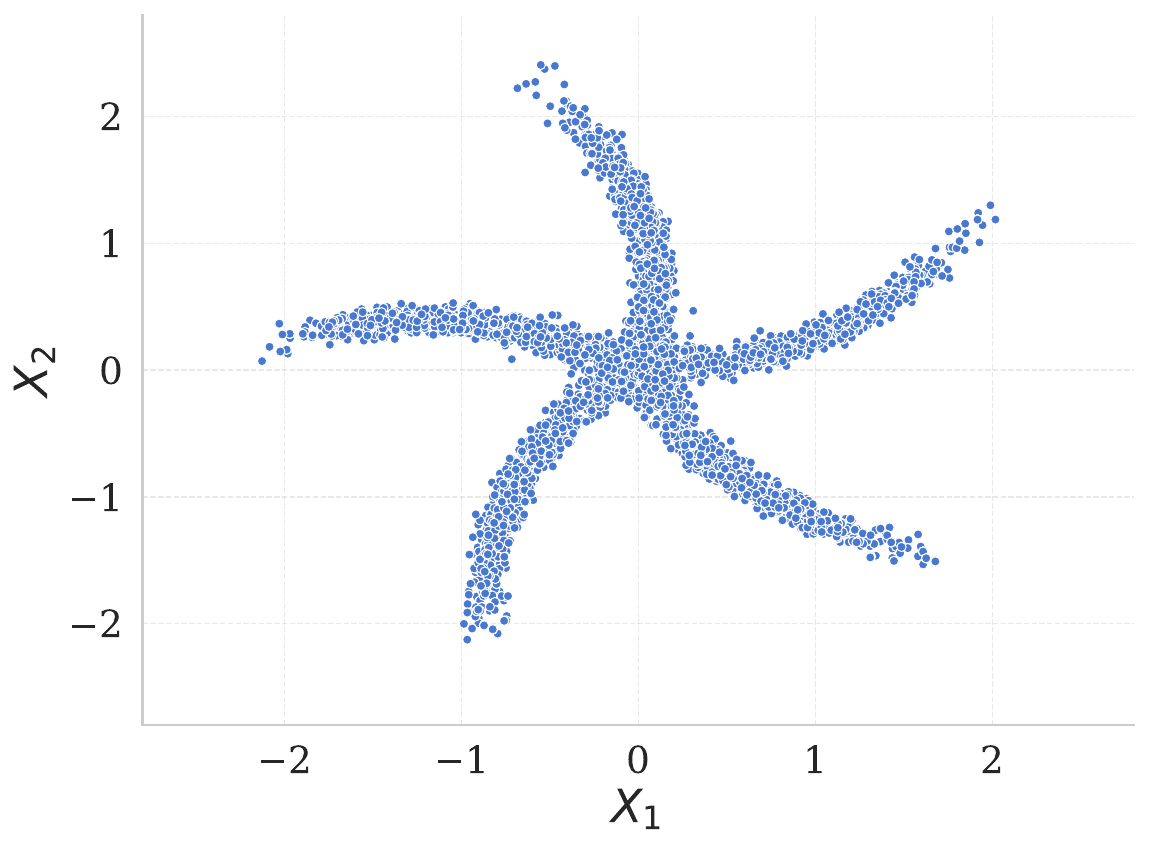}
        \caption{Score RW.}
    \end{subfigure}
    \hfill
    \begin{subfigure}[b]{0.18\textwidth}
        \includegraphics[width=\textwidth]{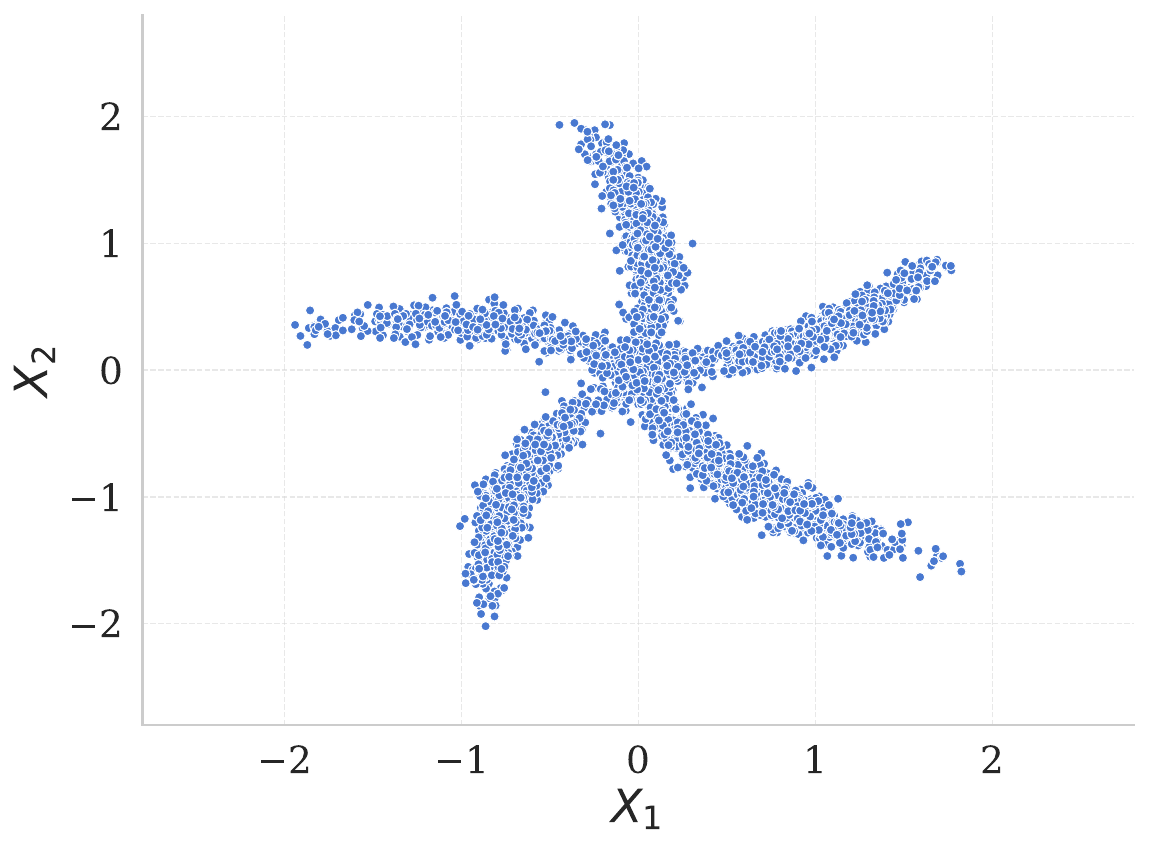}
        \caption{Score MALA}
    \end{subfigure}
    \hfill
    \begin{subfigure}[b]{0.18\textwidth}
        \includegraphics[width=\textwidth]{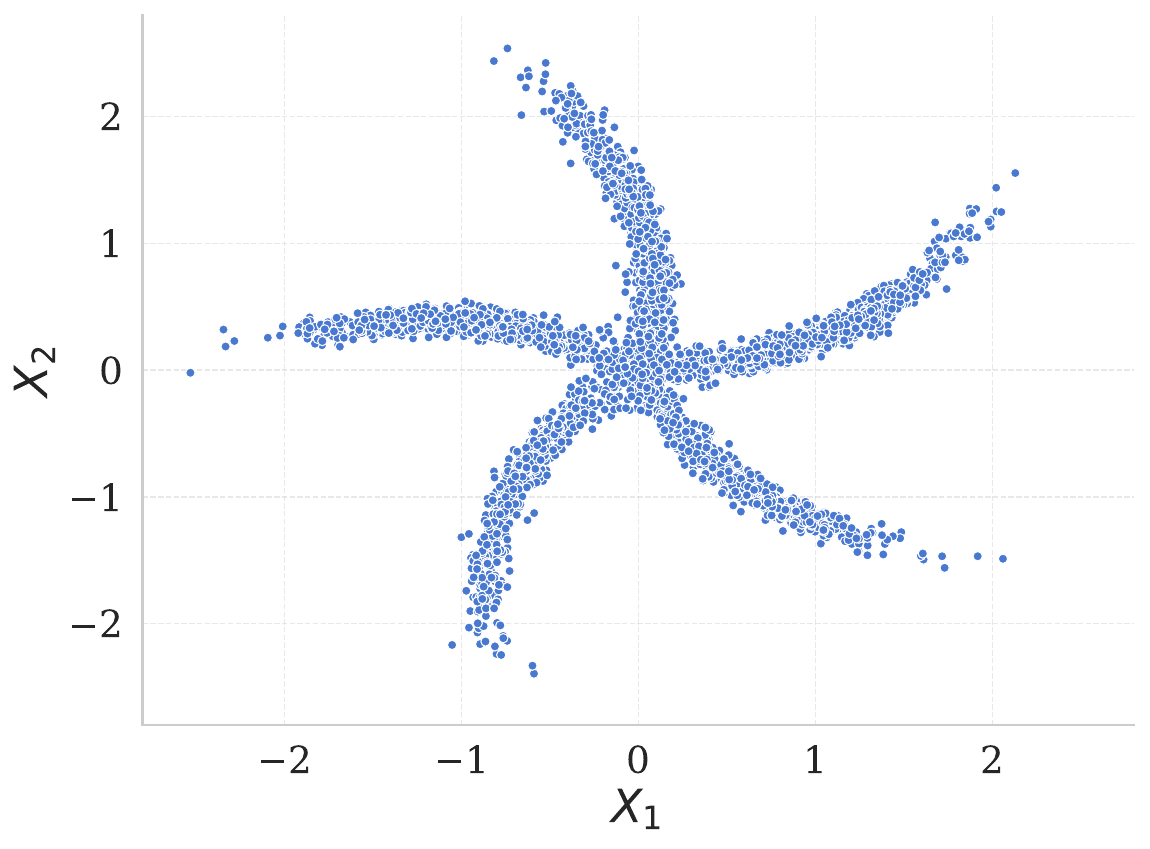}
        \caption{Score pCN.}
    \end{subfigure}
    \caption{Comparison of different methods on the Pinwheel dataset.}
    \label{fig:pinwheel_results}
\end{figure*}

\begin{figure*}[t!]
    \centering
    \begin{subfigure}[b]{0.18\textwidth}
        \includegraphics[width=\textwidth]{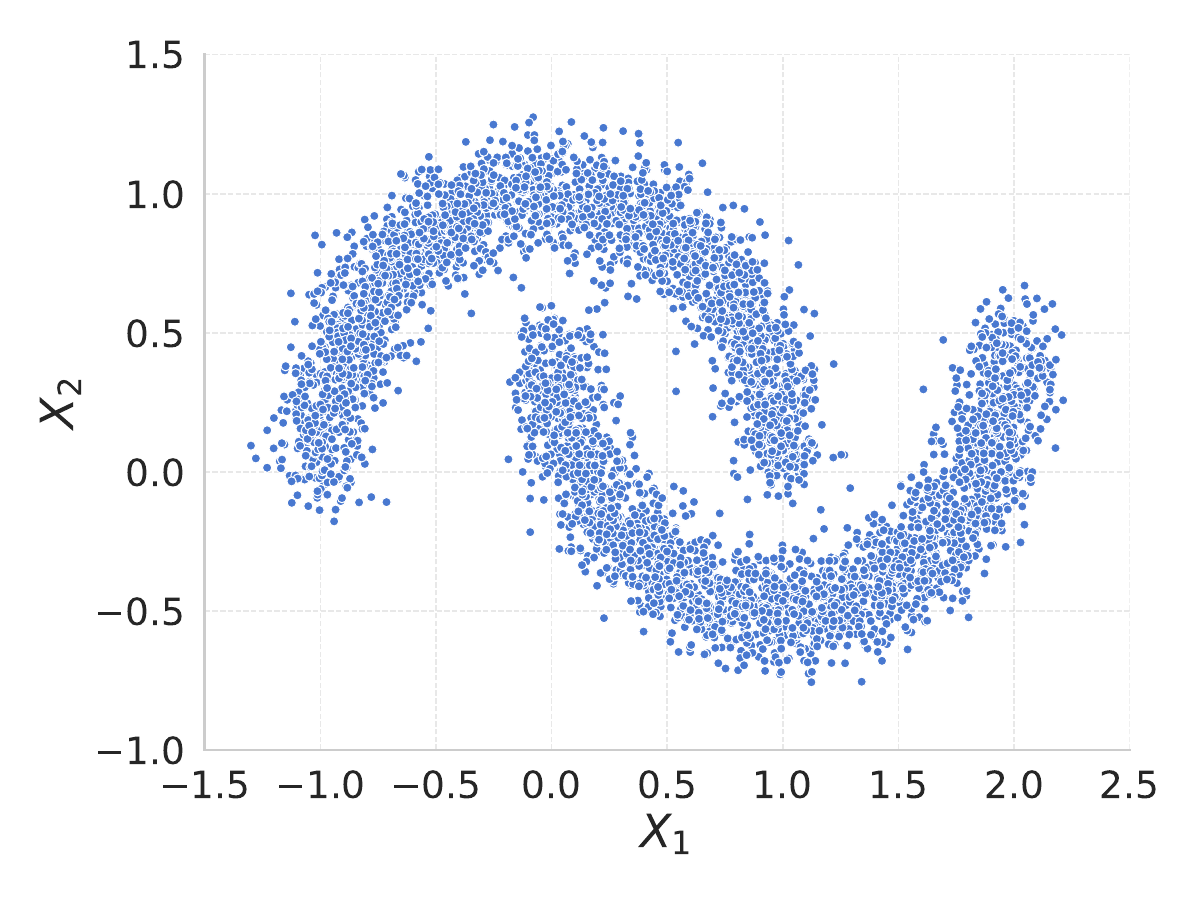}
        \caption{Original samples.}
    \end{subfigure}
    \hfill
    \begin{subfigure}[b]{0.18\textwidth}
        \includegraphics[width=\textwidth]{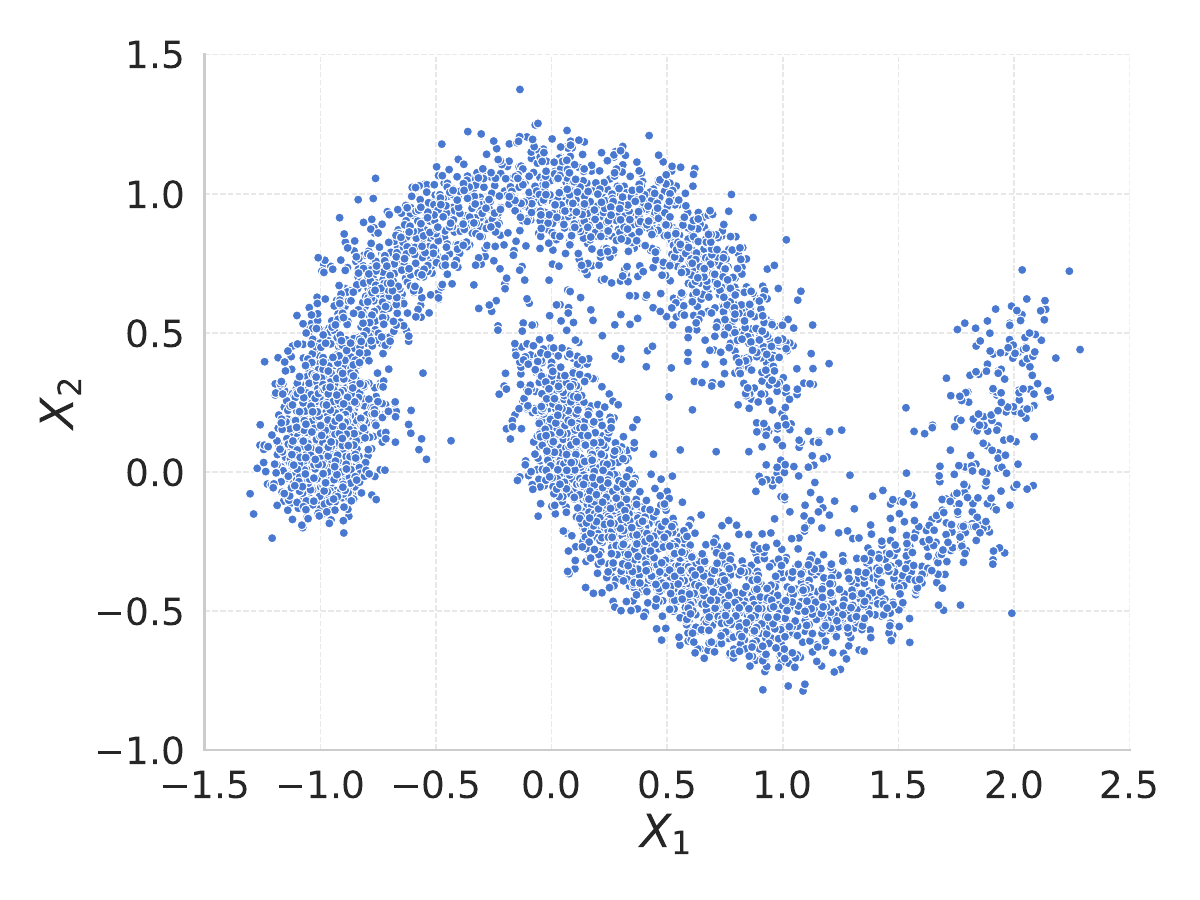}
        \caption{ULA.}
    \end{subfigure}
    \hfill
    \begin{subfigure}[b]{0.18\textwidth}
        \includegraphics[width=\textwidth]{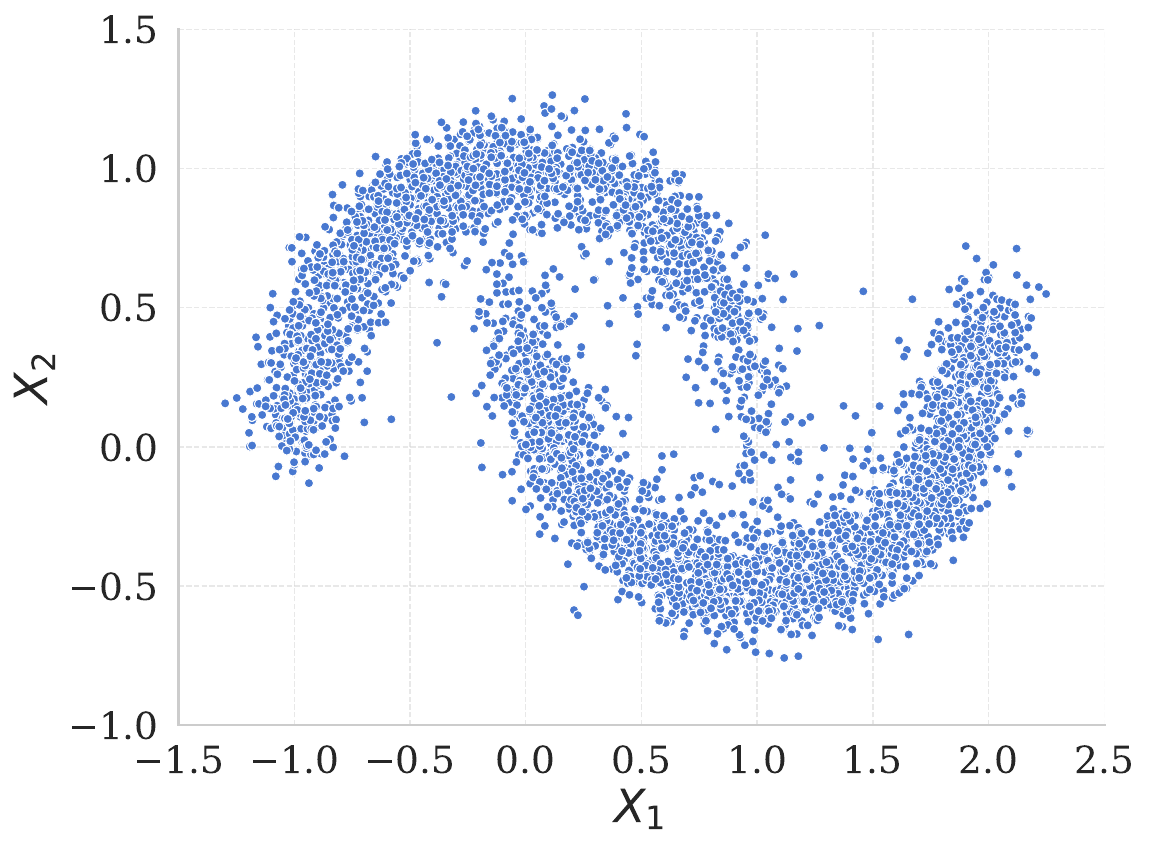}
        \caption{Score RW.}
    \end{subfigure}
    \hfill
    \begin{subfigure}[b]{0.18\textwidth}
        \includegraphics[width=\textwidth]{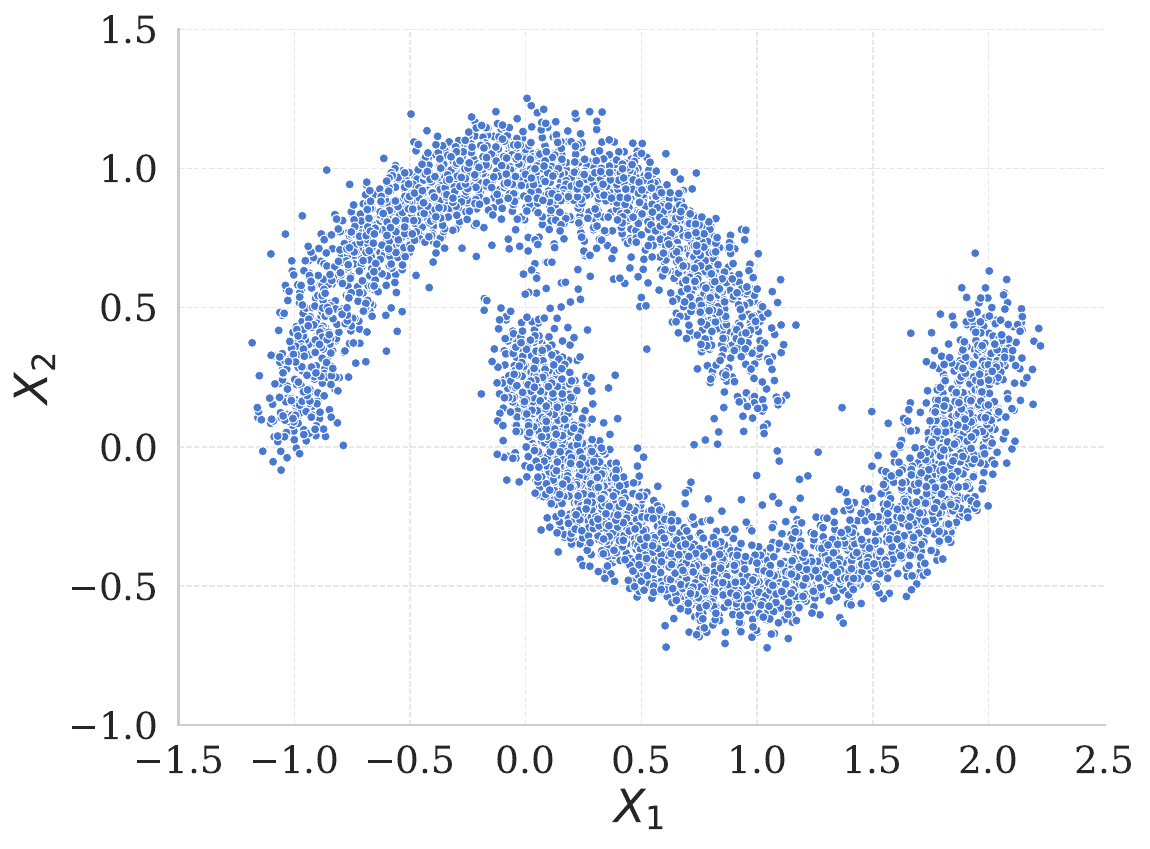}
        \caption{Score MALA.}
    \end{subfigure}
    \hfill
    \begin{subfigure}[b]{0.18\textwidth}
        \includegraphics[width=\textwidth]{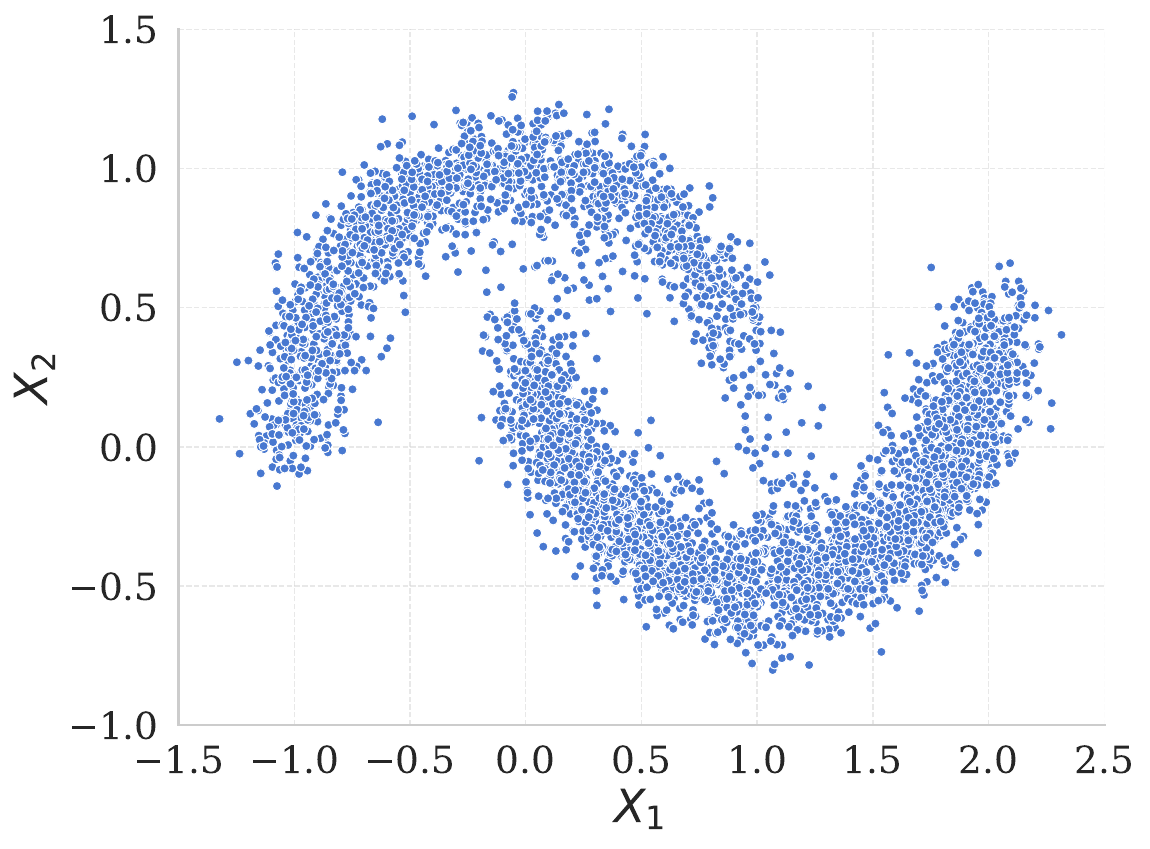}
        \caption{Score pCN.}
    \end{subfigure}
    \caption{Comparison of different methods on the Moons dataset.}
    \label{fig:moons_results}
\end{figure*}

\subsection{Motivation}
Our approach is motivated by three key challenges. First, traditional MH algorithms typically require the knowledge of the unnormalized density, while it is common in generative modeling to assume that only the score function can be accurately estimated. Second, ULA, widely used for sampling from score-based models, suffers from discretization errors that can lead to bias and slow-mixing. Consequently, by incorporating an MH adjustment step, we address a key limitation observed in ULA when sampling from distributions with multiple modes, particularly when these modes are separated by regions of low density. Specifically, when the data distribution is a mixture \( p(x) = \pi p_1(x) + (1 - \pi) p_2(x) \), where \( p_1(x) \) and \( p_2(x) \) are distinct and largely disjoint distributions, the gradient of the log probability density, \( \nabla_x \log p(x) \), becomes misleading. In regions where \( p_1(x) \) dominates, the score is driven solely by \( p_1(x) \), and similarly for \( p_2(x) \). Consequently, ULA, which relies on these score gradients, fails to correctly sample from the mixture distribution as it does not properly account for the mixing proportions \( \pi \) and \( 1 - \pi \). This leads to an incorrect estimation of the relative densities between the modes. The faster mixing time of MALA compared to Langevin has been well-established in the literature as in~\citep{dwivedi2019log,wu2022minimax}. To illustrate this, we conduct experiments to generate a mixture of Gaussian distributions with \( p_1 = \mathcal{N}\left((5,5)^\top, I\right) \) and \( p_2 = \mathcal{N}\left((-5,-5)^\top, I\right) \), with \(\pi = 0.8\). Figure~\ref{fig:slow_mixing_examples} shows that ULA produces samples that \emph{misrepresent the relative weights} of the modes. In contrast, standard MH algorithms \emph{significantly alleviate this problem}, generating samples with mode proportions that closely match the mixture weights. 

\section{Algorithm}\label{sec:loss}
In this section, we formulate the loss function and training algorithm for the acceptance function, which we model using neural networks.
\paragraph{Entropy Regularization.}
\label{sec:entropy}
In addition to the original objective function introduced in Equation~\eqref{eqn:ideal}, our algorithm involves an entropy regularization term. To illustrate its effectiveness, we present the following result as its motivation.
\begin{proposition}\label{prop:infinite}
Let \(a:\mathcal{X}\times\mathcal{X} \rightarrow [0,1]\) and let \(M \geq 1\) be a scaling factor. Then,\[\mathcal{L}(a) = 0 \implies \mathcal{L}(\frac{a}{M}) = 0\]
\end{proposition}It follows from Proposition~\ref{prop:infinite} that there are infinitely many solutions to the detailed balance condition, including cases where the acceptance probabilities can become infinitesimally small. 
While each of these acceptance functions is a mathematically valid solution to the detailed balance condition, in practice, sampling with these low-acceptance probability functions can lead to always rejecting the proposal, resulting in a lack of convergence within a reasonable timeframe. Moreover, these infinitesimal values often cause training instabilities in the neural network that parameterizes the acceptance function.

Therefore, if we \emph{solely} optimize the loss function in~\eqref{eqn:ideal}, we risk converging to minima that are legitimate solutions but impractical due to their very low acceptance probabilities. To address this issue, we propose adding an entropy regularization term to our loss function. 
The modified loss function is given by:
\begin{align}\label{eqn:final}
\mathcal{L}_{r}(a) & = \mathbb{E}\bigg[ \left\| \nabla \log a(X', X) - \nabla \log a(X, X') \right. \nonumber \\
& - \tilde{s}(X') + \tilde{s}(X) - \nabla \log q(X \mid X') \nonumber \\
& \; + \nabla \log q(X' \mid X) \|^2 \bigg] + \lambda \, \mathbb{E}\left[ H(a(X', X)) \right],
\end{align}
where: \begin{align*}
H(a(X', X)) = &\log(a(X', X)) a(X', X) \\ &+ \log(1 - a(X', X)) \left(1 - a(X', X)\right).
\end{align*}
and $\lambda \geq 0$ is the weight of the regularization term. This regularization term encourages acceptance functions with higher entropy acceptance probabilities.

\begin{figure*}[ht]
    \centering
    \begin{subfigure}[b]{0.18\textwidth}
        \includegraphics[width=\textwidth]{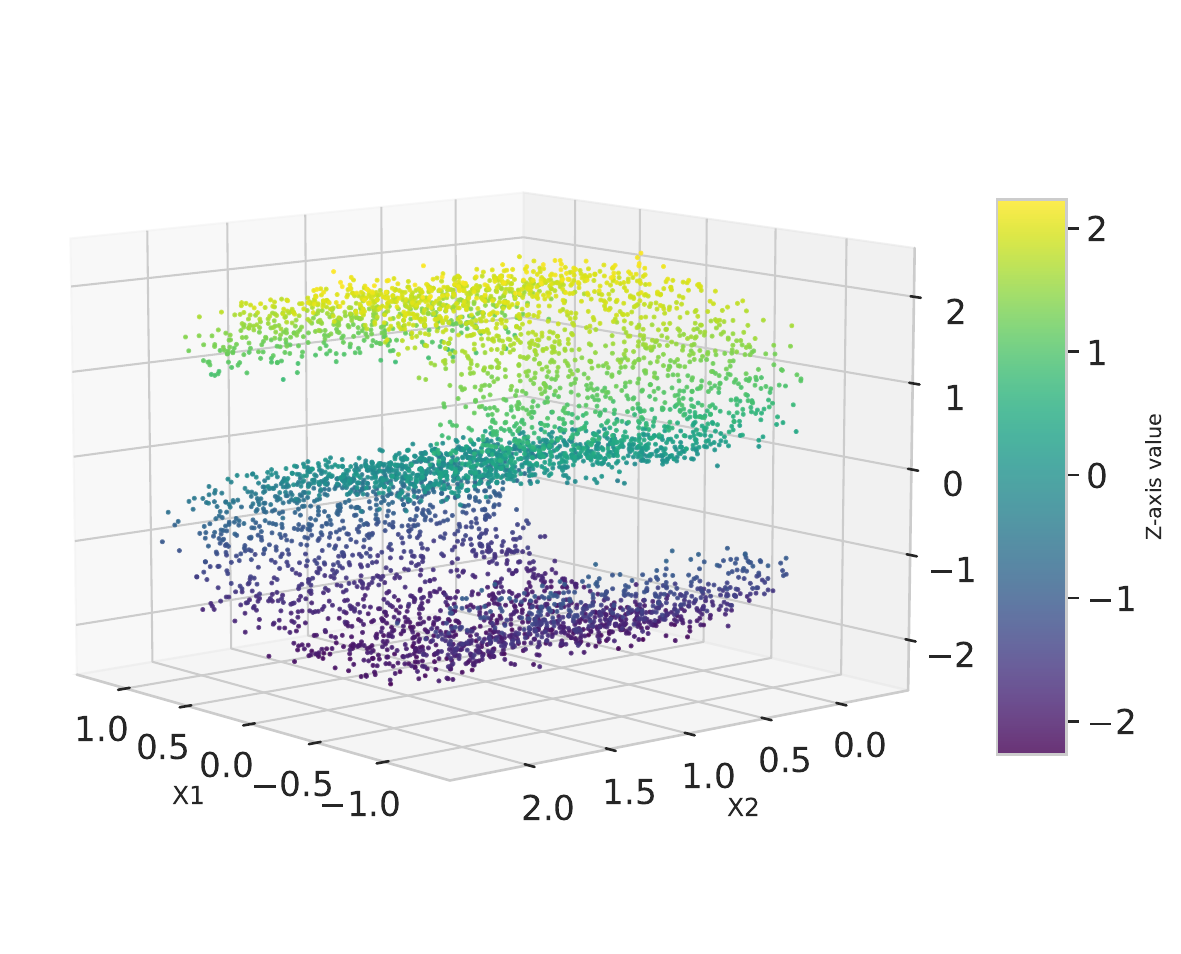}
        \caption{Original samples.}
    \end{subfigure}
    \hfill
    \begin{subfigure}[b]{0.18\textwidth}
        \includegraphics[width=\textwidth]{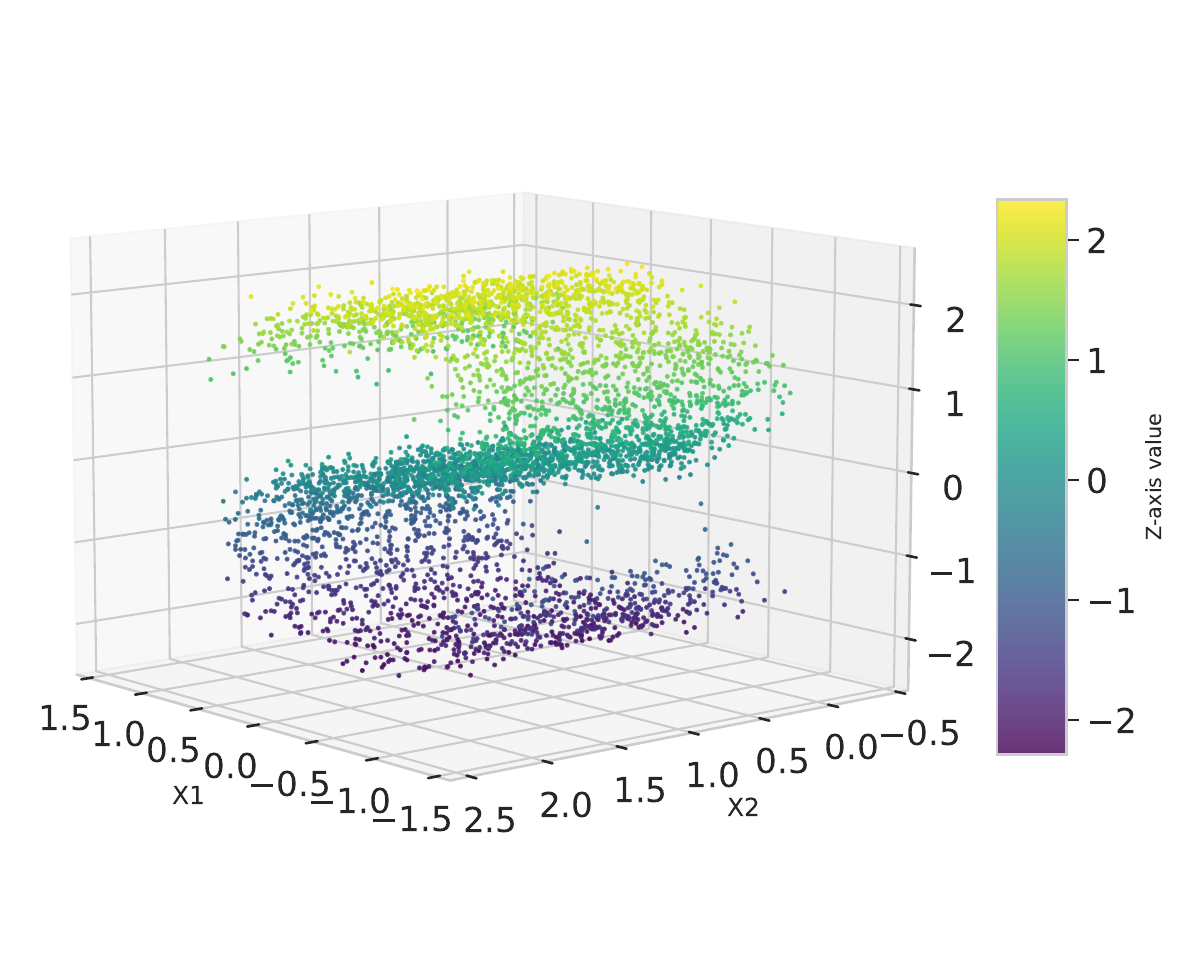}
        \caption{ULA.}
    \end{subfigure}
    \hfill
    \begin{subfigure}[b]{0.18\textwidth}
        \includegraphics[width=\textwidth]{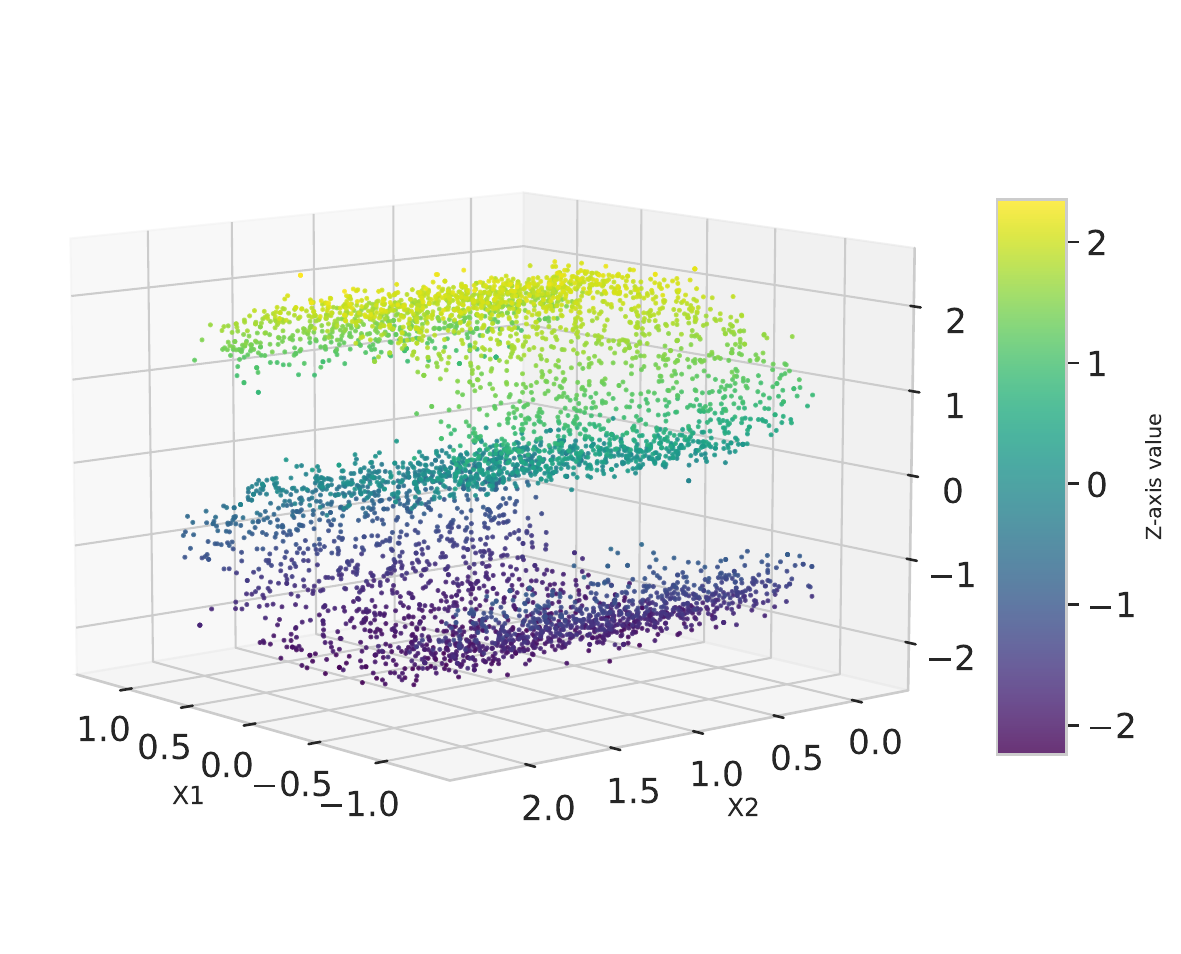}
        \caption{Score RW.}
    \end{subfigure}
    \hfill
    \begin{subfigure}[b]{0.18\textwidth}
        \includegraphics[width=\textwidth]{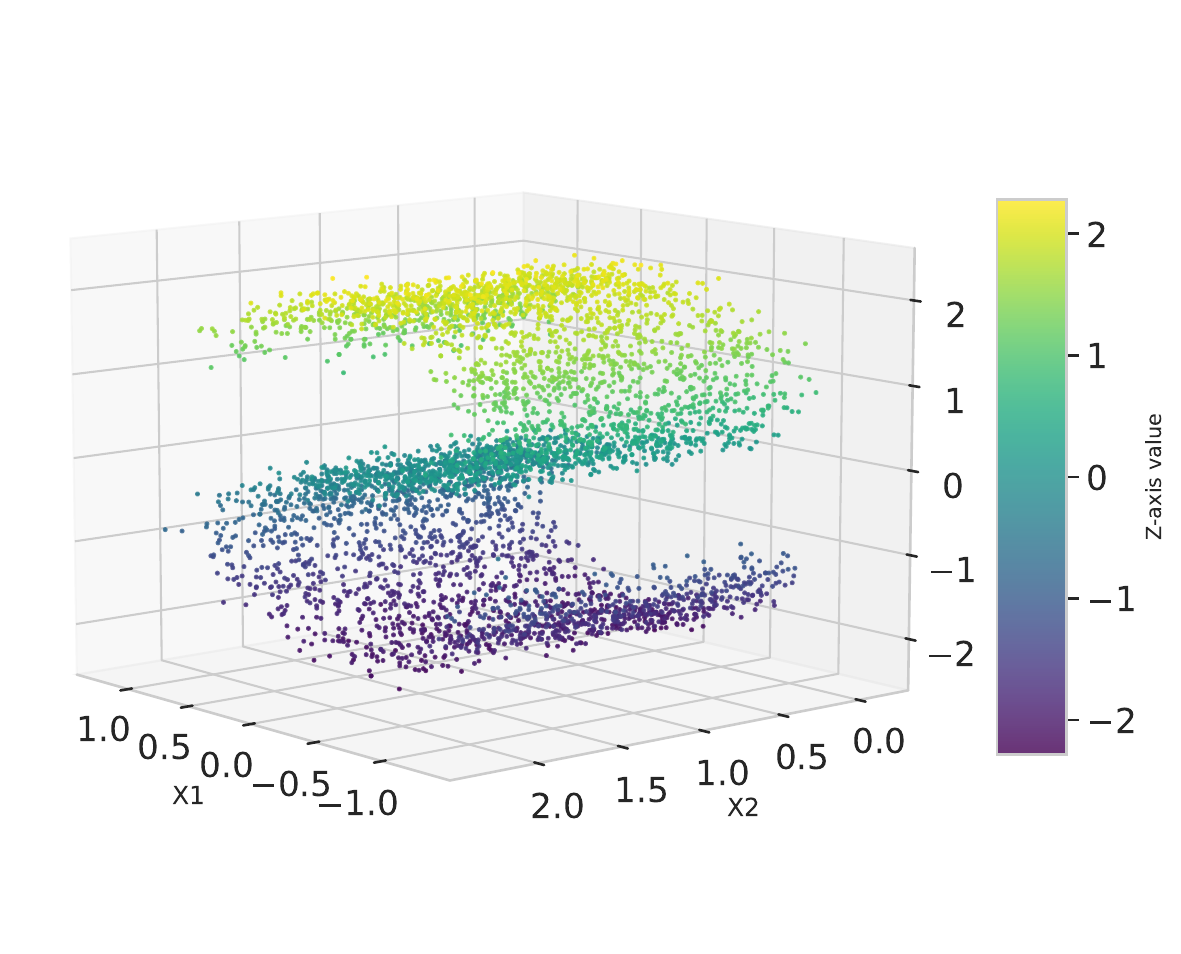}
        \caption{Score MALA.}
    \end{subfigure}
    \hfill
    \begin{subfigure}[b]{0.18\textwidth}
        \includegraphics[width=\textwidth]{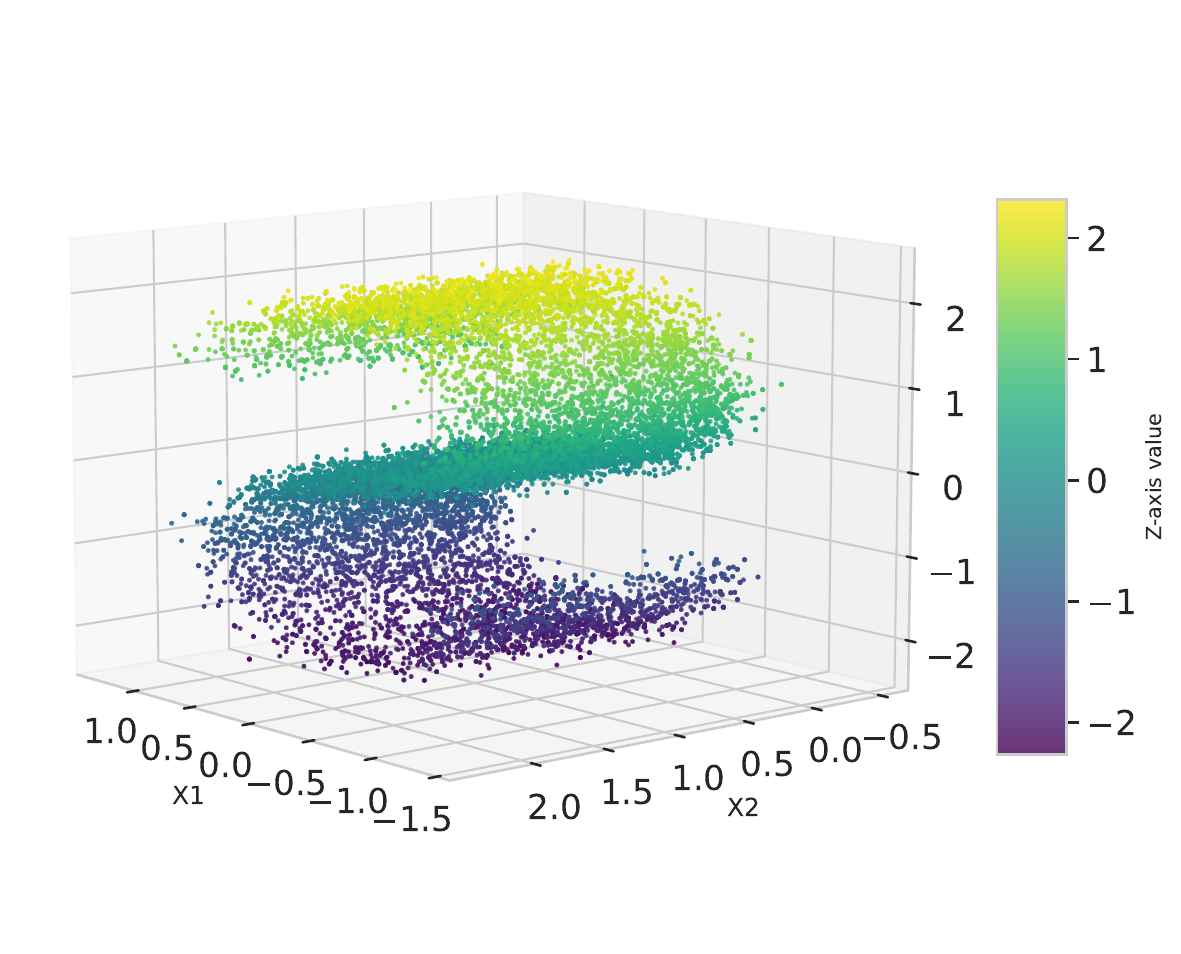}
        \caption{Score pCN.}
    \end{subfigure}
    \caption{Comparison of different methods on the S-curve dataset.}
    \label{fig:curve_results}
\end{figure*}

To illustrate the importance of the regularization term, we present the results using the Moons dataset from \texttt{scikit-learn}~\citep{scikit-learn}. 
Based on an estimated score function, we estimate an acceptance function to sample from the Moons dataset using the proposed Score RW algorithm. 
Training is done for different values of the regularization parameter \(\lambda\), and the results are plotted in Figure~\ref{fig:moons_entropy}. 
It can observed that the acceptance probability is significantly enhanced with the regularization whereas training without the entropy regularization results in acceptance probability close $0$. Although this acceptance probability remains mathematically valid (i.e., it will still converge to the target distribution), the convergence time may become impractically long.
\paragraph{Implementation.}
Typically, only samples from the target distribution $p$ are observed, and the true $\nabla \log p$ is unknown. 
However, score parameterization through a neural network has proven successful in practice. 
We assume that we have an approximation of the score function previously estimated through a model $\tilde{s}\left(x\right) \approx \nabla \log p\left(x\right)$. 
Note that this is often the case in practice where an estimated score function $\tilde{s}$ is used in conjunction with ULA for sampling. 
The final loss function used to train the acceptance network in our score MH framework consists of two components: the primary loss term, which enforces the reversibility condition, and an entropy regularization term. The primary loss is designed to minimize the difference between the gradients of the log acceptance probability and the gradients of the log posterior and proposal distributions.
\begin{align}\label{eqn:empirical_loss}
& \mathcal{L}_{\text{emp}}(a) =\frac{1}{N} \sum_{i=1}^N l_i(a) + \frac{\lambda}{N} \sum_{i=1}^N h_i(a) 
\end{align}
with 
\begin{align*}
l_i(a)  = \bigg\| &\nabla \log a\left(x'^{(i)}, x^{(i)}\right) - \nabla \log a\left(x^{(i)}, x'^{(i)}\right) \\ & - \Tilde{s}\left(x'^{(i)}\right)  
  + \Tilde{s}\left(x^{(i)}\right)   \\ 
  & -\nabla \log q\left(x^{(i)} \mid x'^{(i)}\right) + \nabla \log q\left(x'^{(i)} \mid x^{(i)}\right) \bigg\|^2
\end{align*}
and
\begin{align*}
     h_i(a) & = \log a\left(x'^{(i)}, x^{(i)}\right) \cdot a\left(x'^{(i)}, x^{(i)}\right) \\ & \quad + \log\left( 1 - a\left(x'^{(i)},x^{(i)}\right)\right) \cdot \left( 1 - a\left(x'^{(i)},x^{(i)}\right)\right).
\end{align*}
and where \(x'^{(i)} = \alpha v' + (1 - \alpha) \tilde{x}\), with \(\tilde{x} \sim p\) and \(v' \sim q(\cdot\mid \tilde{x})\), where \(\alpha \in [0,1]\) controls how much \(x'\) deviates from the data distribution towards the proposal. In the first epochs of training, we begin with smaller values of \(\alpha\) to favor data from the true distribution, and then gradually increase \(\alpha\) to estimate the acceptance function around the proposal region. This approach is motivated by the fact that the score function may not be well-estimated outside of the data region, and starting with smaller \(\alpha\) helps prevent biasing the training with poorly learned scores.
We present the pseudo-code for the proposed loss function in Algorithm~\ref{alg:loss} and the training procedure in Algorithm~\ref{alg:acceptance}. 
We made several design choices that we observed to be beneficial for learning stable acceptance probabilities:
\begin{itemize}
    \item Using residual blocks akin to~\citet{he2016deep}.
    \item Employing smooth activation functions, e.g., GeLU.
    \item Clipping the gradients of the log acceptance function and the gradient of the log densities. This enables numerical stability and is also motivated by assumptions ~\ref{ass:cont} and ~\ref{ass:hyp_bound}.
\end{itemize}

\begin{algorithm}[t]
\caption{Acceptance Loss Function}
\label{alg:loss}
\smallskip
\textbf{Input:} Acceptance net \(a(x',x)\), Score Net \(s\), batch of samples \(b_x\), batch of proposals \(b_{x'}\), regularization parameter \(\lambda\), gradient clipping threshold \(C\).

\smallskip
\textbf{Step 1:} Compute gradients \(\nabla \log a(x', x)\) and \(\nabla \log a(x, x')\) using Autograd for every \((x, x') \in b_x \times b_{x'}\).

\smallskip
\textbf{Step 2:} Compute score gradients \(s(x)\) and \(s(x')\) for every \((x, x') \in b_x \times b_{x'}\).

\smallskip
\textbf{Step 3:} Compute proposal gradients \(\nabla q(x|x')\) and \(\nabla q(x'|x)\) for every \((x, x') \in b_x \times b_{x'}\).

\smallskip
\textbf{Step 4:} Clip all gradients to avoid instability using threshold \(C\).

\smallskip
\textbf{Step 5:} Compute the loss \(\mathcal{L}_{\text{emp}}\) as defined in~\eqref{eqn:empirical_loss}.

\smallskip
\textbf{Output:} \(\mathcal{L}_{\text{emp}}\)
\end{algorithm}

\begin{algorithm}[ht!]
\caption{Training the Acceptance Network}
\label{alg:acceptance}
\smallskip
\textbf{Input:} Acceptance Network \(a\), Score Network \(s\), dataset \(D\), number of epochs \(N\), sequence \(\{\alpha_i\}_{i=1}^N \subset [0,1]\) (increasing).

\smallskip
\textbf{For each epoch \(i = 1 \text{ to } N\):}
\begin{enumerate}
    \item Sample a batch \(b_x\) and a batch \(b_{\tilde{x}}\) from \(D\).
    \item For each \(\tilde{x} \in b_{\tilde{x}}\), sample \(v' \sim q(\cdot \mid \tilde{x})\) and set \(x' = \alpha_i v' + (1-\alpha_i) \tilde{x}\).
    \item Compute the loss as defined in Algorithm~\ref{alg:loss}.
    \item Update the Acceptance Network \(a\) using the Adam optimizer.
\end{enumerate}

\smallskip
\textbf{Output:} Trained Acceptance Network \(a\).
\end{algorithm}

\begin{figure*}[t]
    \centering
    \begin{subfigure}{0.21\textwidth}
        \includegraphics[width=\textwidth]{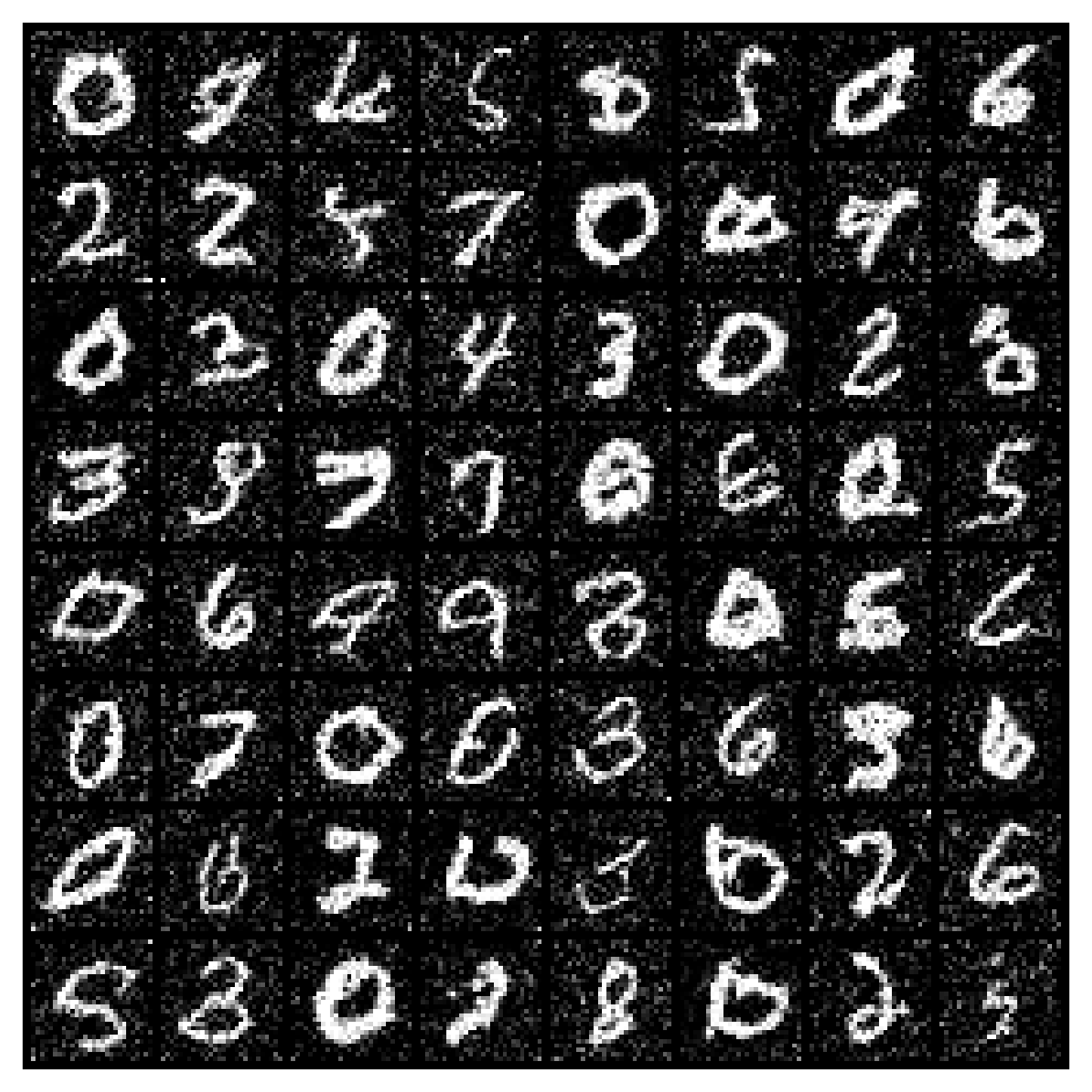}
        \caption{MALA, $\tau = 0.5$}
    \end{subfigure}
    \hfill
    \begin{subfigure}{0.21\textwidth}
        \includegraphics[width=\textwidth]{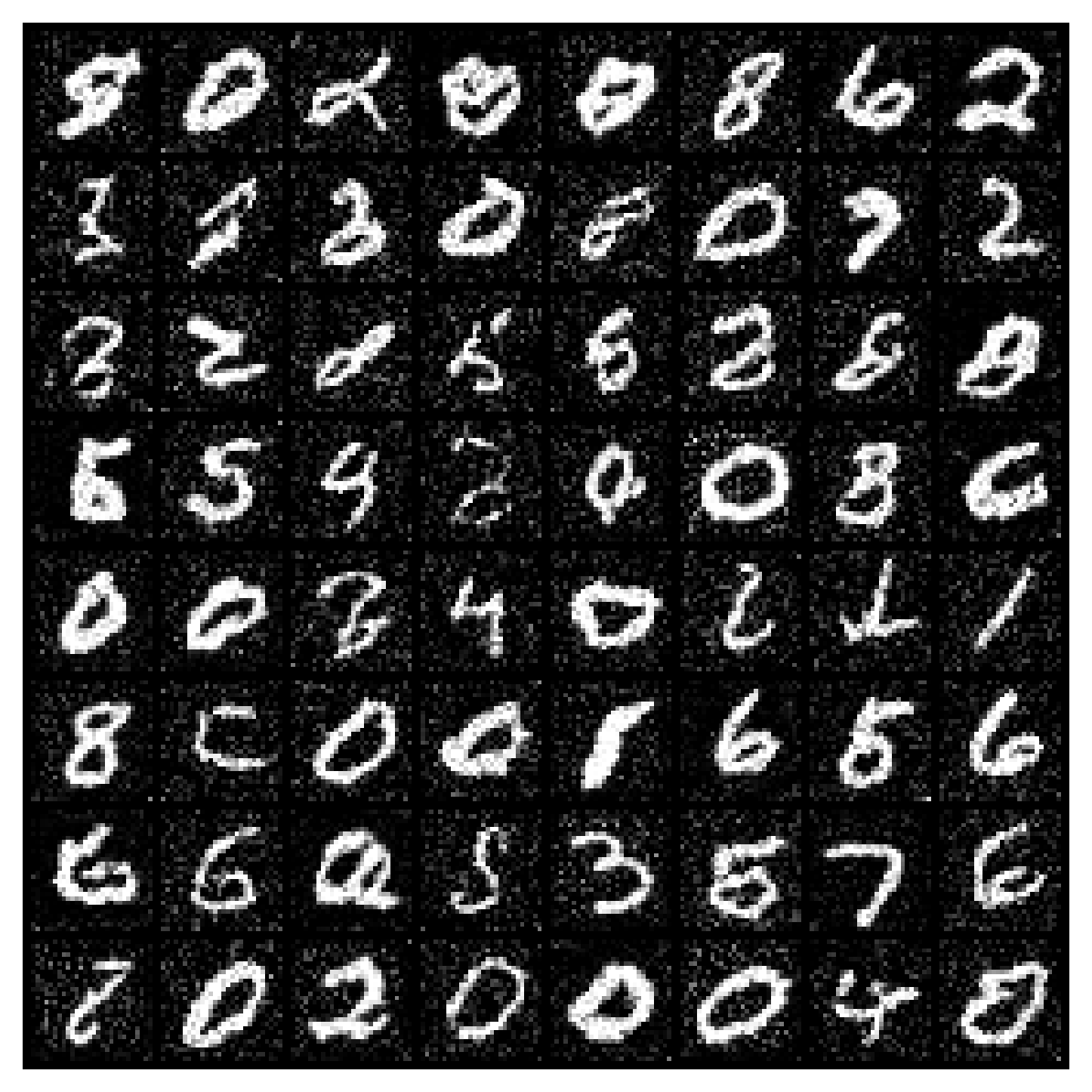}
        \caption{MALA, $\tau = 0.7$}
    \end{subfigure}
    \hfill
    \begin{subfigure}{0.21\textwidth}
        \includegraphics[width=\textwidth]{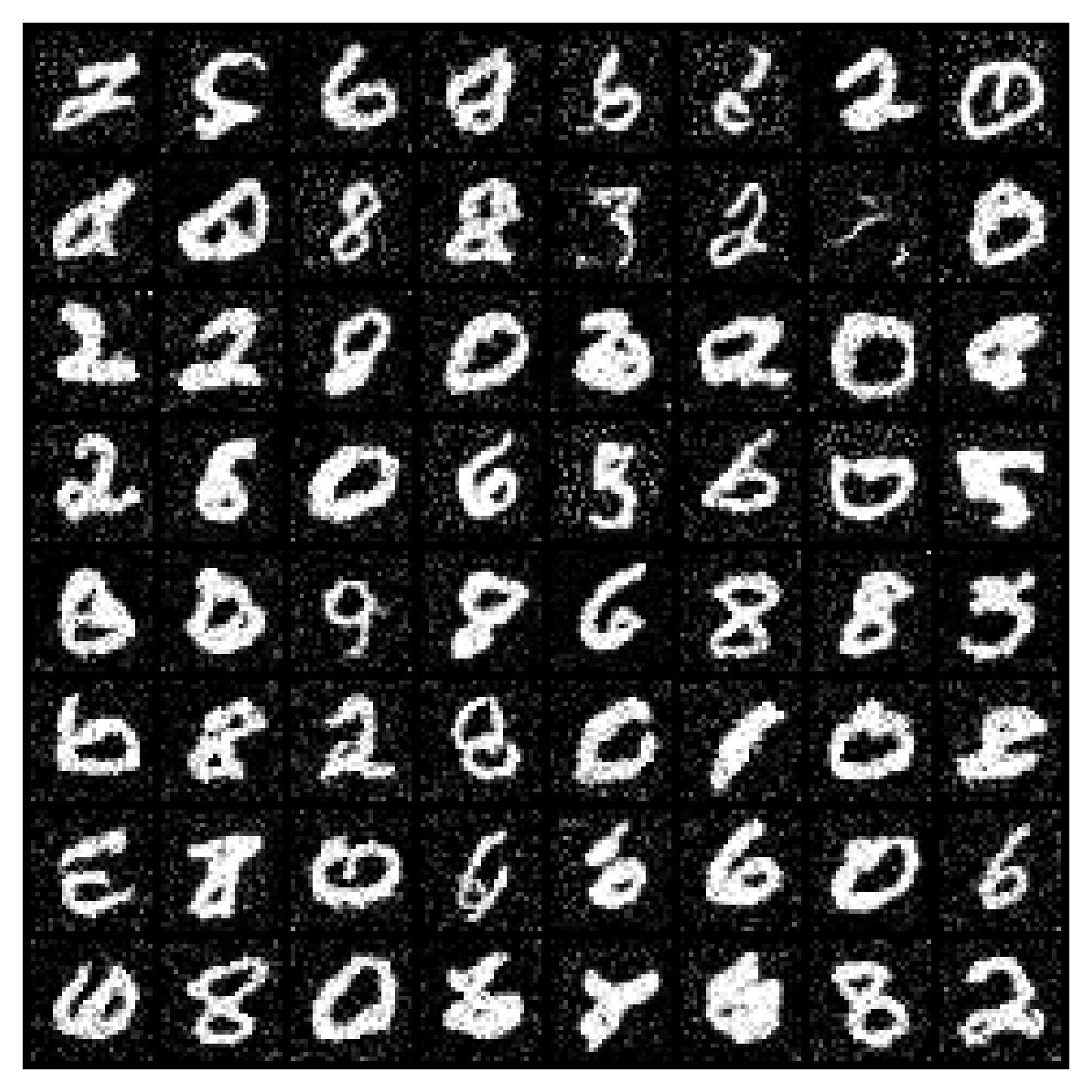}
        \caption{MALA, $\tau = 0.9$}
    \end{subfigure}
    \hfill
    \begin{subfigure}{0.21\textwidth}
        \includegraphics[width=\textwidth]{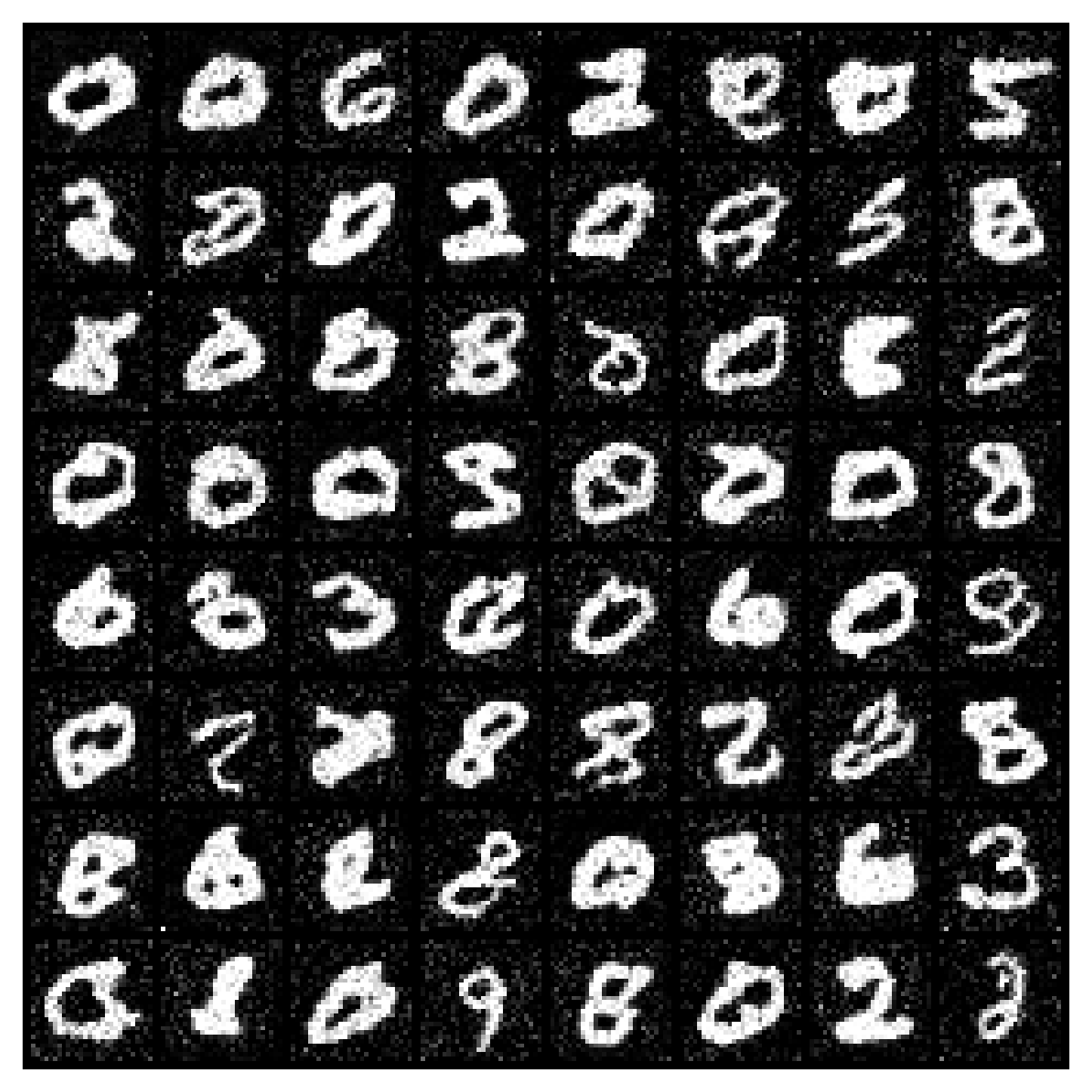}
        \caption{MALA, $\tau = 1.1$}
    \end{subfigure}

    \vspace{0.5cm}
    \begin{subfigure}{0.21\textwidth}
        \includegraphics[width=\textwidth]{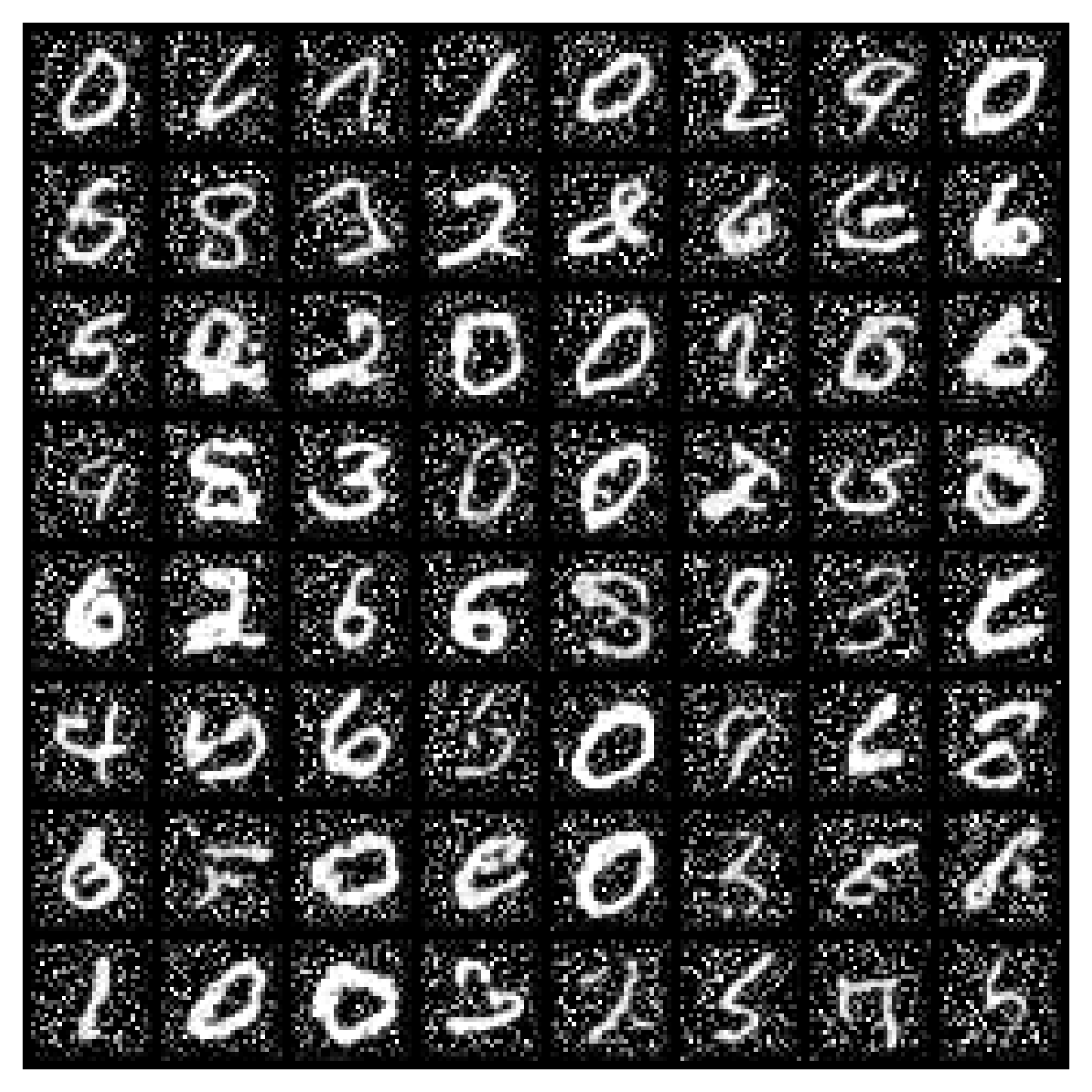}
        \caption{ULA, $\tau = 0.5$}
    \end{subfigure}
    \hfill
    \begin{subfigure}{0.21\textwidth}
        \includegraphics[width=\textwidth]{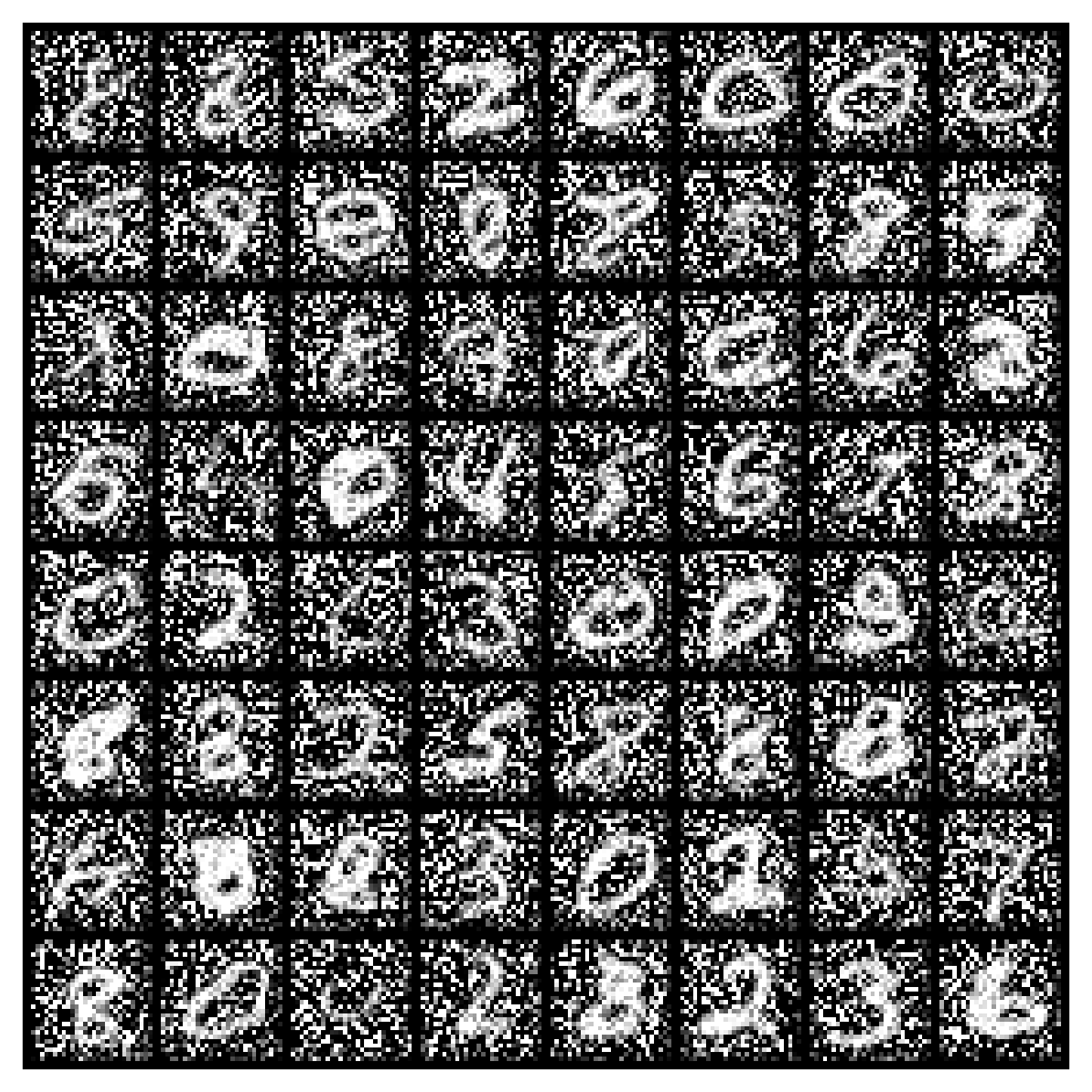}
        \caption{ULA, $\tau = 0.7$}
    \end{subfigure}
    \hfill
    \begin{subfigure}{0.21\textwidth}
        \includegraphics[width=\textwidth]{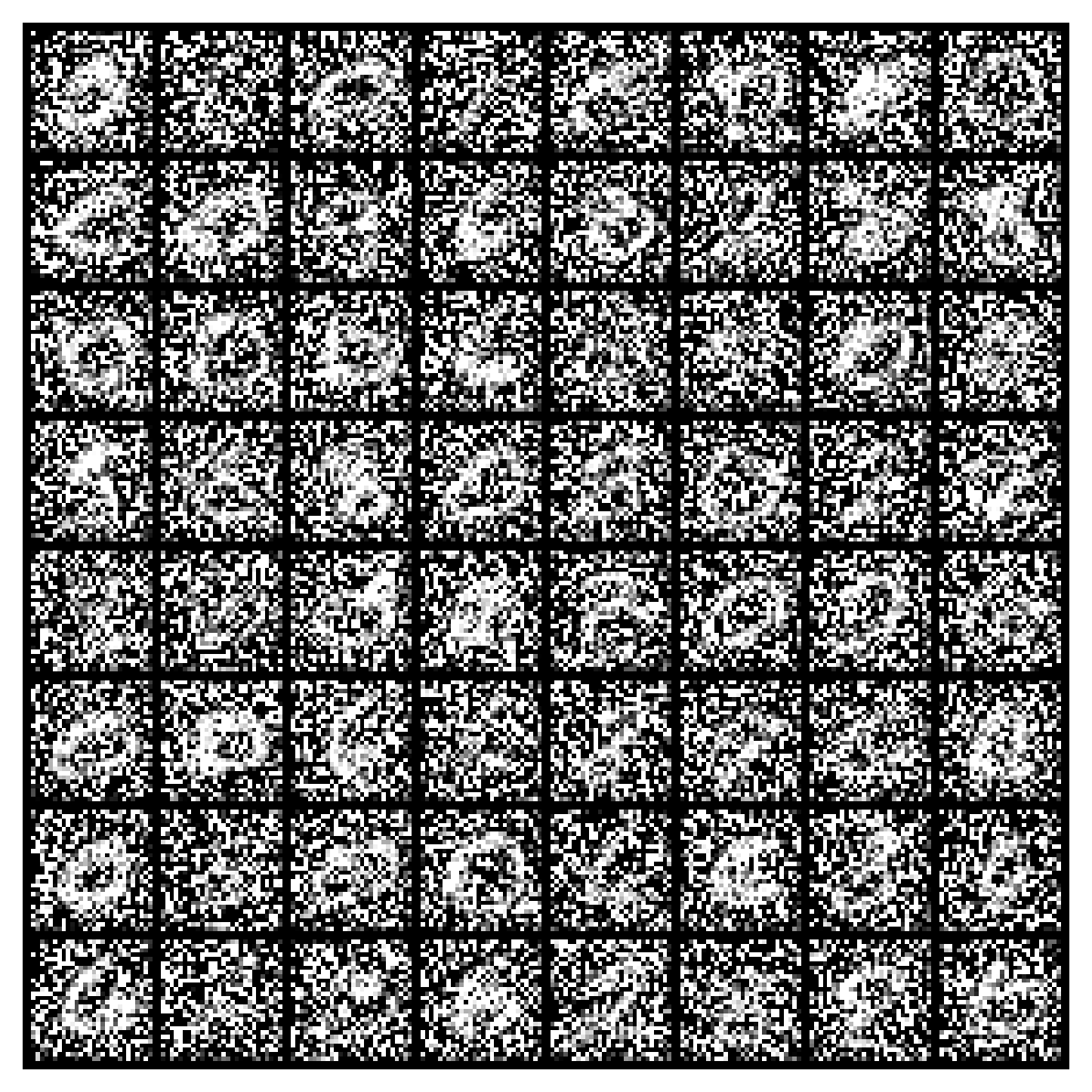}
        \caption{ULA, $\tau = 0.9$}
    \end{subfigure}
    \hfill
    \begin{subfigure}{0.21\textwidth}
        \includegraphics[width=\textwidth]{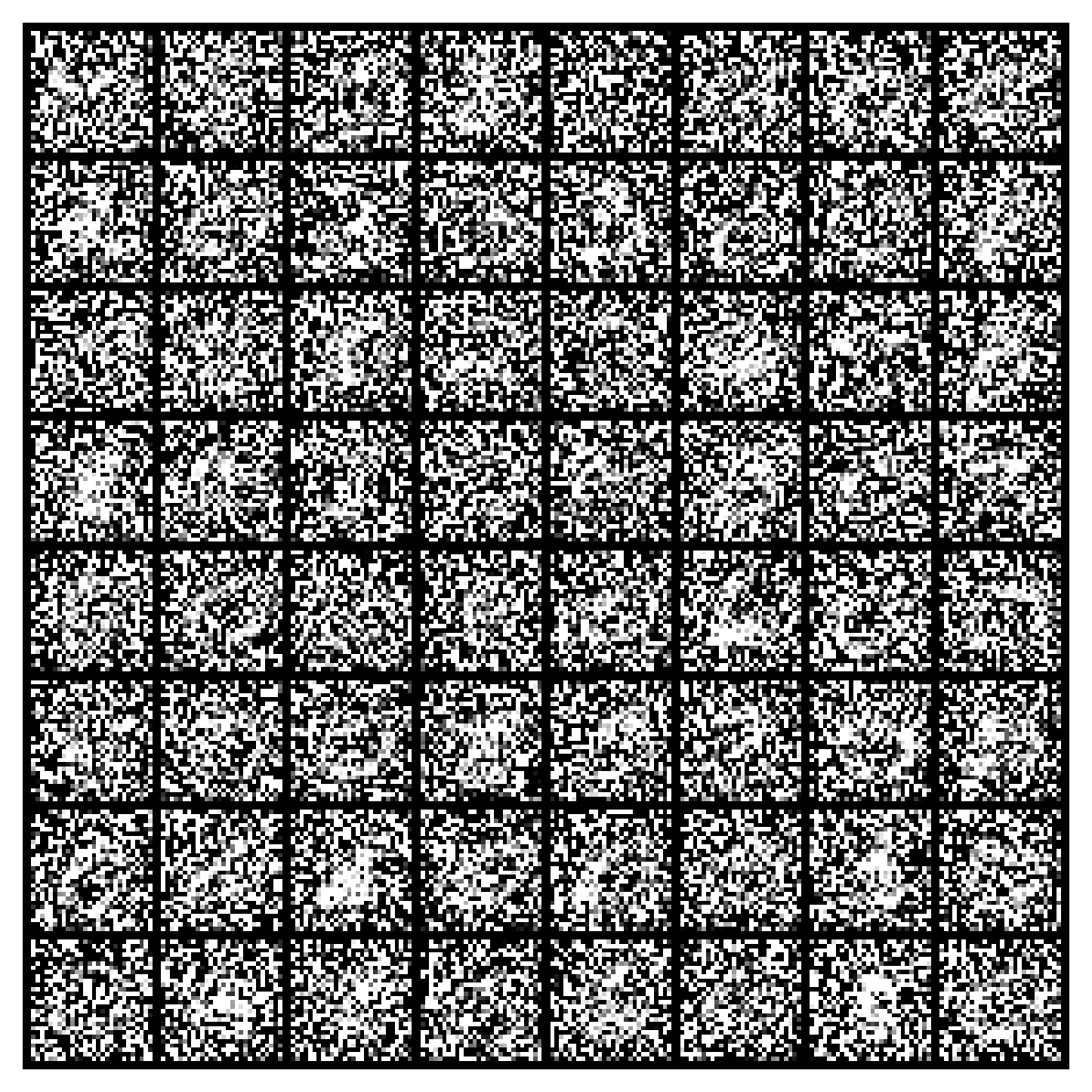}
        \caption{ULA, $\tau = 1.1$}
    \end{subfigure}

    \caption{MNIST: comparing annealed MALA (top) and annealed ULA (bottom) across $\tau$ values $0.5$, $0.7$, $0.9$, and $1.1$. Results show that MALA is more robust to step size. Both algorithms were run for $100$ denoising steps, with an additional $1000$ steps at the smallest noise level. $\tau$ is a fixed parameter that controls the adaptive step size.}
    \label{fig:mala_ula_comparison}
\end{figure*}

\begin{table*}[ht!]
    \centering
    \caption{Quantitative comparison of methods across the datasets using the following metrics: Wasserstein-1 (W1), Wasserstein-2 (W2), and Maximum Mean Discrepancy (MMD).}
    \label{tab:quantitative_results}
    \setlength\tabcolsep{4pt} 
    \begin{tabular}{@{}l|ccc|ccc|ccc|ccc@{}}
        \toprule
        \textbf{Dataset} & \multicolumn{3}{c|}{\textbf{ULA}} & \multicolumn{3}{c|}{\textbf{Score RW}} & \multicolumn{3}{c|}{\textbf{Score MALA}} & \multicolumn{3}{c}{\textbf{Score pCN}} \\
        \midrule
        \textbf{Metric} &\textbf{W1}& \textbf{W2}& \textbf{MMD}& \textbf{W1} & \textbf{W2} & \textbf{MMD} & \textbf{W1} & \textbf{W2} & \textbf{MMD} & \textbf{W1} & \textbf{W2} & \textbf{MMD} \\
        \midrule
        Moons & $0.143$ & $0.096$ & $0.054$ & $0.019$ & $0.007$ & $0.004$ & $\bm{0.018}$ & $\bm{0.004}$ & $\bm{0.002}$ & $0.039$ & $0.021$ & $0.012$ \\
        Pinwheel & $0.106$ & $0.062$ & $0.022$ & $\bm{0.012}$ & $\bm{0.001}$ & $\bm{0.001}$ & $0.086$ & $0.064$ & $0.027$ & $0.016$ & $0.007$ & $0.003$ \\
        S-curve & $0.082$ & $0.055$ & $0.016$ & $0.050$ & $0.015$ & $\bm{0.004}$ & $\bm{0.046}$ & $\bm{0.014}$ & $\bm{0.004}$ & $0.090$ & $0.051$ & $0.017$ \\
        Swiss Roll & $2.881$ & $14.409$ & $0.811$ & $2.270$ & $8.470$ & $0.400$ & $\bm{1.399}$ & $\bm{4.575}$ & $\bm{0.245}$ & $14.167$ & $113.824$ & $6.903$ \\
        \bottomrule
    \end{tabular}
\end{table*}



\begin{figure}
    \centering
    \begin{subfigure}[b]{0.4\textwidth}
        \centering
        \includegraphics[width=1.0\textwidth]{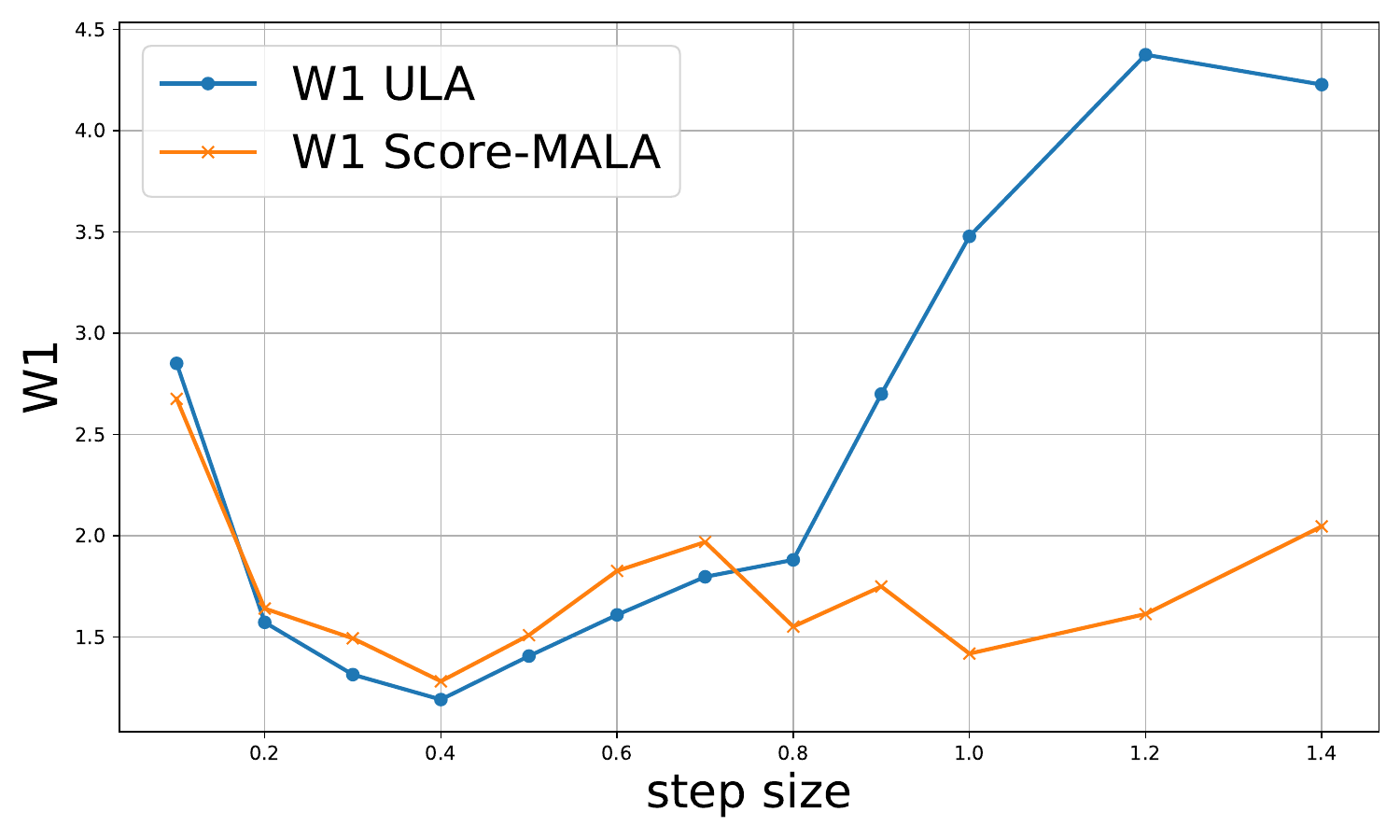}
        \caption{W1 vs step size}
    \end{subfigure}
    \caption{Influence of step size on the performance of Score-based MALA vs. ULA, evaluated on the Swiss Roll dataset.}
    \label{fig:step_size}
\end{figure}

\section{Empirical Results}


In this section, we present empirical results on various datasets (Moons, Pinwheel, S-curve, Swiss Roll, and MNIST), to demonstrate the validity of our approach. Further results on extreme value distributions are in Appendix~\ref{sec:heavy_tails}.

\begin{figure}[ht]
    \centering
    \includegraphics[width=0.72\linewidth]{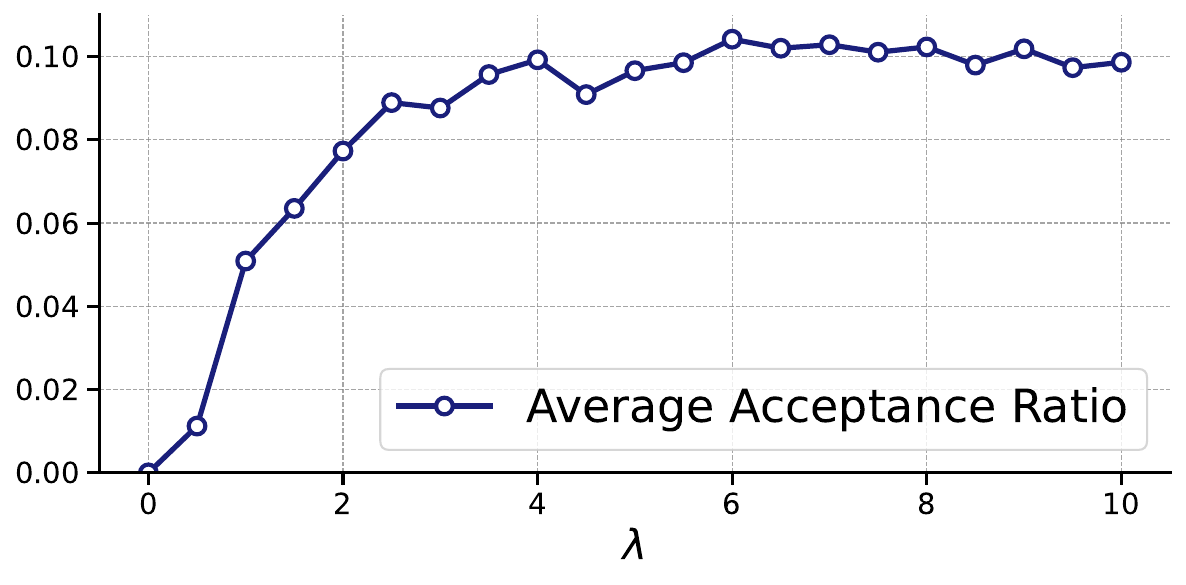}
    \caption{Average acceptance ratio as a function of $\lambda$. The average acceptance ratio is calculated by evaluating the acceptance network over a uniform grid of $x'$ values, with ranges $[-1.5, 2.25]$ for the first dimension and $[-1, 1.5]$ for the second dimension, while keeping $x = (0, 0)$ fixed. The mean acceptance ratio is then obtained by averaging uniformly over the grid.}
    \label{fig:accept_vs_lambda}
\end{figure}
\textbf{Results Analysis.} The results in Table~\ref{tab:quantitative_results} illustrate the performance of four sampling methods: ULA, Score-based RW, Score-based MALA, and Score-based pCN, across four datasets using three evaluation metrics: Wasserstein-1 (W1), Wasserstein-2 (W2), and Maximum Mean Discrepancy (MMD). Figures~\ref{fig:pinwheel_results},~\ref{fig:moons_results}, and ~\ref{fig:curve_results} illustrate the results for low-dimensional data. Overall, score-based methods achieve lower W1 and W2 values compared to ULA across all datasets. This trend is especially noticeable on the Moons and Pinwheel datasets, where the W2 metric significantly outperforms that of the ULA method.
For more complex datasets like Swiss Roll, the score MH algorithms show noticeable improvements, with Score RW and Score MALA achieving significantly lower W1 and MMD scores when compared to the other methods. 

\begin{figure}[h]
    \centering
    \includegraphics[width=0.75\linewidth]{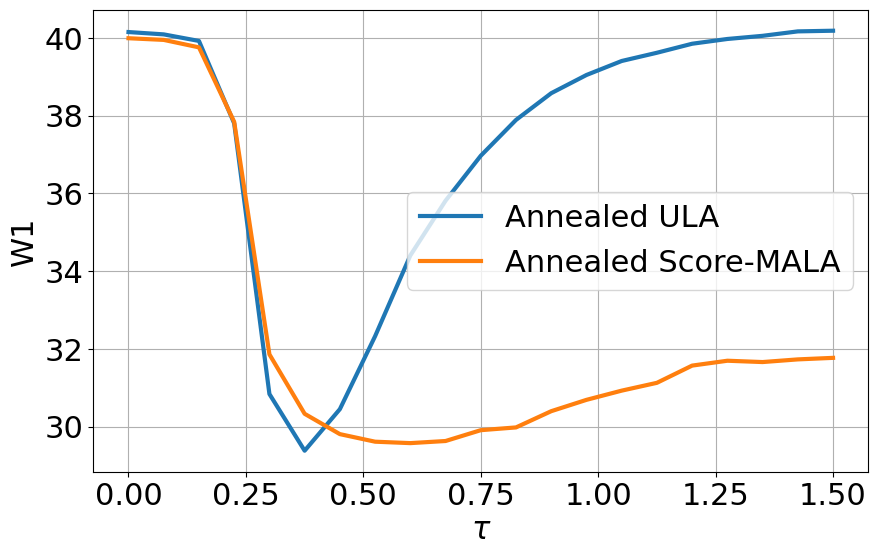}
    \caption{MNIST data generation comparing annealed ULA and annealed score-based MALA. $\tau$ is a fixed parameter that controls the adaptive step size.}
    \label{fig:mnist_snr}
\end{figure}

\paragraph{Stability with respect to Step Size: Score MALA vs. ULA.} 
The results demonstrate that Score MALA is more robust to variations in step size compared to ULA as seen in Figures~\ref{fig:mala_ula_comparison}, ~\ref{fig:step_size} and~\ref{fig:mnist_snr}.

\paragraph{Effect of Entropy Regularization.}
We study the effect of the entropy regularization term on the Moons dataset. As shown in Figure~\ref{fig:accept_vs_lambda}, the relationship between the entropy regularization parameter $\lambda$ and the average acceptance ratio indicates a significant impact on the sampling efficiency. As $\lambda$ increases from $0$ to around $4$, the average acceptance ratio rises sharply, suggesting that stronger entropy regularization enhances learning more practical acceptance functions. Beyond $\lambda = 4$, the acceptance ratio plateaus. This demonstrates that the performance remains robust for larger values of the entropy regularization term.

\paragraph{Diffusion Models Application.} We evaluate the performance of the the proposed method with denoising score matching applied to the MNIST dataset using annealed ULA and annealing with score-based MALA. Annealing ~\citep{song2019generative,song2020score} involves a sequence of noise scales, progressively refining the samples from high noise to low noise, thus improving the stability and efficiency of score-based sampling. Further details on the annealing process and its implementation are provided in the appendix. Our results presented in Figures~\ref{fig:mala_ula_comparison} and ~\ref{fig:mnist_snr} indicate that the proposed method exhibits greater stability with respect to step size variations compared to annealed ULA, as measured by the sensitivity parameter $\tau$. We present further details in Appendix~\ref{sec:experiments}.
\section{Discussion}
In this work we describe a method for estimating the acceptance function given only knowledge of samples and an estimated score function. We demonstrated the successful estimation of an acceptance function applicable across various MH sampling algorithms. While we empirically evaluate on three particular instances, this work opens the door for further implementations and techniques for sampling using acceptance functions. Additionally, we discussed a particular class of acceptance functions that are amenable for efficient sampling using entropy regularization.
The proposed methods provide new avenues for combining MH within the context of score-based generative modeling.
\paragraph{Limitations.}
The main limitation of the method is the additional computational cost required to train the acceptance network. Furthermore, more advanced architectures need to be proposed to extend this framework to higher-resolution image data. Additionally, further theoretical analysis is needed to provide more appropriate priors on which acceptance function should be chosen.

\paragraph{Acknowledgments}
Ahmed Aloui and Vahid Tarokh's work was supported in part by the Air Force Office of Scientific Research under award number FA9550-20-1-0397.

\bibliography{references}

\begin{thebibliography}{40}
\providecommand{\natexlab}[1]{#1}
\providecommand{\url}[1]{\texttt{#1}}
\expandafter\ifx\csname urlstyle\endcsname\relax
  \providecommand{\doi}[1]{doi: #1}\else
  \providecommand{\doi}{doi: \begingroup \urlstyle{rm}\Url}\fi

\bibitem[Andrieu and Thoms(2008)]{andrieu2008tutorial}
Christophe Andrieu and Johannes Thoms.
\newblock A tutorial on adaptive mcmc.
\newblock \emph{Statistics and computing}, 18:\penalty0 343--373, 2008.

\bibitem[Biron-Lattes et~al.(2024)Biron-Lattes, Surjanovic, Syed, Campbell, and Bouchard-C{\^o}t{\'e}]{biron2024automala}
Miguel Biron-Lattes, Nikola Surjanovic, Saifuddin Syed, Trevor Campbell, and Alexandre Bouchard-C{\^o}t{\'e}.
\newblock automala: Locally adaptive metropolis-adjusted langevin algorithm.
\newblock In \emph{International Conference on Artificial Intelligence and Statistics}, pages 4600--4608. PMLR, 2024.

\bibitem[Brooks et~al.(2011)Brooks, Gelman, Jones, and Meng]{brooks2011handbook}
Steve Brooks, Andrew Gelman, Galin Jones, and Xiao-Li Meng.
\newblock \emph{Handbook of markov chain monte carlo}.
\newblock CRC press, 2011.

\bibitem[Chen et~al.(2023)Chen, Chewi, Li, Li, Salim, and Zhang]{chen2023sampling}
Sitan Chen, Sinho Chewi, Jerry Li, Yuanzhi Li, Adil Salim, and Anru Zhang.
\newblock Sampling is as easy as learning the score: theory for diffusion models with minimal data assumptions.
\newblock In \emph{The Eleventh International Conference on Learning Representations}, 2023.

\bibitem[Chen et~al.(2024)Chen, He, Fu, Tian, and Tao]{chen2024adaptive}
Yuzhu Chen, Fengxiang He, Shi Fu, Xinmei Tian, and Dacheng Tao.
\newblock Adaptive time-stepping schedules for diffusion models.
\newblock In \emph{The 40th Conference on Uncertainty in Artificial Intelligence}, 2024.

\bibitem[Davies et~al.(2023)Davies, Salomone, Sutton, and Drovandi]{davies23a}
Laurence Davies, Robert Salomone, Matthew Sutton, and Chris Drovandi.
\newblock Transport reversible jump proposals.
\newblock In Francisco Ruiz, Jennifer Dy, and Jan-Willem van~de Meent, editors, \emph{Proceedings of The 26th International Conference on Artificial Intelligence and Statistics}, volume 206 of \emph{Proceedings of Machine Learning Research}, pages 6839--6852. PMLR, 25--27 Apr 2023.

\bibitem[Dinh et~al.(2016)Dinh, Sohl-Dickstein, and Bengio]{dinh2016density}
Laurent Dinh, Jascha Sohl-Dickstein, and Samy Bengio.
\newblock Density estimation using real nvp.
\newblock \emph{arXiv preprint arXiv:1605.08803}, 2016.

\bibitem[Dockhorn et~al.(2021)Dockhorn, Vahdat, and Kreis]{dockhorn2021score}
Tim Dockhorn, Arash Vahdat, and Karsten Kreis.
\newblock Score-based generative modeling with critically-damped langevin diffusion.
\newblock \emph{arXiv preprint arXiv:2112.07068}, 2021.

\bibitem[Dwivedi et~al.(2019)Dwivedi, Chen, Wainwright, and Yu]{dwivedi2019log}
Raaz Dwivedi, Yuansi Chen, Martin~J Wainwright, and Bin Yu.
\newblock Log-concave sampling: Metropolis-hastings algorithms are fast.
\newblock \emph{Journal of Machine Learning Research}, 20\penalty0 (183):\penalty0 1--42, 2019.

\bibitem[Goodfellow et~al.(2020)Goodfellow, Pouget-Abadie, Mirza, Xu, Warde-Farley, Ozair, Courville, and Bengio]{goodfellow2020generative}
Ian Goodfellow, Jean Pouget-Abadie, Mehdi Mirza, Bing Xu, David Warde-Farley, Sherjil Ozair, Aaron Courville, and Yoshua Bengio.
\newblock Generative adversarial networks.
\newblock \emph{Communications of the ACM}, 63\penalty0 (11):\penalty0 139--144, 2020.

\bibitem[Grenander and Miller(1994)]{grenander1994representations}
Ulf Grenander and Michael~I Miller.
\newblock Representations of knowledge in complex systems.
\newblock \emph{Journal of the Royal Statistical Society: Series B (Methodological)}, 56\penalty0 (4):\penalty0 549--581, 1994.

\bibitem[Haario et~al.(2001)Haario, Saksman, and Tamminen]{haario2001adaptive}
Heikki Haario, Eero Saksman, and Johanna Tamminen.
\newblock An adaptive metropolis algorithm.
\newblock \emph{Bernoulli}, 7\penalty0 (2):\penalty0 223--242, April 2001.

\bibitem[Hairer et~al.(2014)Hairer, Stuart, and Vollmer]{hairer2014spectral}
Martin Hairer, Andrew~M. Stuart, and Sebastian~J. Vollmer.
\newblock Spectral gaps for a metropolis–hastings algorithm in infinite dimensions.
\newblock \emph{The Annals of Applied Probability}, 24\penalty0 (6):\penalty0 2455--2490, December 2014.

\bibitem[Hastings(1970)]{hastings1970monte}
W.~Keith Hastings.
\newblock Monte carlo sampling methods using markov chains and their applications.
\newblock \emph{Biometrika}, 1970.

\bibitem[He et~al.(2016)He, Zhang, Ren, and Sun]{he2016deep}
Kaiming He, Xiangyu Zhang, Shaoqing Ren, and Jian Sun.
\newblock Deep residual learning for image recognition.
\newblock In \emph{Proceedings of the IEEE conference on computer vision and pattern recognition}, pages 770--778, 2016.

\bibitem[Hirt et~al.(2021)Hirt, Titsias, and Dellaportas]{hirt2021entropy}
Marcel Hirt, Michalis Titsias, and Petros Dellaportas.
\newblock Entropy-based adaptive hamiltonian monte carlo.
\newblock \emph{Advances in Neural Information Processing Systems}, 34:\penalty0 28482--28495, 2021.

\bibitem[Ho et~al.(2020)Ho, Jain, and Abbeel]{ho2020denoising}
Jonathan Ho, Ajay Jain, and Pieter Abbeel.
\newblock Denoising diffusion probabilistic models.
\newblock \emph{Advances in neural information processing systems}, 33:\penalty0 6840--6851, 2020.

\bibitem[Hyv{\"a}rinen(2005)]{hyvarinen2005estimation}
Aapo Hyv{\"a}rinen.
\newblock Estimation of non-normalized statistical models by score matching.
\newblock \emph{Journal of Machine Learning Research}, 6\penalty0 (4), 2005.

\bibitem[Kingma(2013)]{kingma2013auto}
Diederik~P Kingma.
\newblock Auto-encoding variational bayes.
\newblock \emph{arXiv preprint arXiv:1312.6114}, 2013.

\bibitem[Lew et~al.(2023)Lew, Matheos, Zhi-Xuan, Ghavamizadeh, Gothoskar, Russell, and Mansinghka]{lew23a}
Alexander~K. Lew, George Matheos, Tan Zhi-Xuan, Matin Ghavamizadeh, Nishad Gothoskar, Stuart Russell, and Vikash~K. Mansinghka.
\newblock Smcp3: Sequential monte carlo with probabilistic program proposals.
\newblock In \emph{Proceedings of The 26th International Conference on Artificial Intelligence and Statistics}, pages 7061--7088, 2023.

\bibitem[Lu et~al.(2022)Lu, Zhou, Bao, Chen, Li, and Zhu]{lu2022dpm}
Cheng Lu, Yuhao Zhou, Fan Bao, Jianfei Chen, Chongxuan Li, and Jun Zhu.
\newblock Dpm-solver: A fast ode solver for diffusion probabilistic model sampling in around 10 steps.
\newblock \emph{Advances in Neural Information Processing Systems}, 35:\penalty0 5775--5787, 2022.

\bibitem[Metropolis et~al.(1953)Metropolis, Rosenbluth, Rosenbluth, Teller, and Teller]{metropolis1953equation}
Nicholas Metropolis, Arianna~W Rosenbluth, Marshall~N Rosenbluth, Augusta~H Teller, and Edward Teller.
\newblock Equation of state calculations by fast computing machines.
\newblock \emph{The journal of chemical physics}, 21\penalty0 (6):\penalty0 1087--1092, 1953.

\bibitem[Mohri(2018)]{mohri2018foundations}
Mehryar Mohri.
\newblock Foundations of machine learning, 2018.

\bibitem[Pedregosa et~al.(2011)Pedregosa, Varoquaux, Gramfort, Michel, Thirion, Grisel, Blondel, Prettenhofer, Weiss, Dubourg, Vanderplas, Passos, Cournapeau, Brucher, Perrot, and Duchesnay]{scikit-learn}
F.~Pedregosa, G.~Varoquaux, A.~Gramfort, V.~Michel, B.~Thirion, O.~Grisel, M.~Blondel, P.~Prettenhofer, R.~Weiss, V.~Dubourg, J.~Vanderplas, A.~Passos, D.~Cournapeau, M.~Brucher, M.~Perrot, and E.~Duchesnay.
\newblock Scikit-learn: Machine learning in {P}ython.
\newblock \emph{Journal of Machine Learning Research}, 12:\penalty0 2825--2830, 2011.

\bibitem[Reu et~al.()Reu, Vargas, Kerekes, and Bronstein]{reusmooth}
Teodora Reu, Francisco Vargas, Anna Kerekes, and Michael~M Bronstein.
\newblock To smooth a cloud or to pin it down: Expressiveness guarantees and insights on score matching in denoising diffusion models.
\newblock In \emph{The 40th Conference on Uncertainty in Artificial Intelligence}.

\bibitem[Robert et~al.(2004)Robert, Casella, Robert, and Casella]{robert2004metropolis}
Christian~P Robert, George Casella, Christian~P Robert, and George Casella.
\newblock The metropolis—hastings algorithm.
\newblock \emph{Monte Carlo statistical methods}, pages 267--320, 2004.

\bibitem[Roberts and Rosenthal(2009)]{roberts2009examples}
Gareth~O Roberts and Jeffrey~S Rosenthal.
\newblock Examples of adaptive mcmc.
\newblock \emph{Journal of computational and graphical statistics}, 18\penalty0 (2):\penalty0 349--367, 2009.

\bibitem[Roberts and Stramer(2002)]{roberts2002langevin}
Gareth~O Roberts and Osnat Stramer.
\newblock Langevin diffusions and metropolis-hastings algorithms.
\newblock \emph{Methodology and computing in applied probability}, 4:\penalty0 337--357, 2002.

\bibitem[Roberts and Tweedie(1996)]{roberts1996exponential}
Gareth~O. Roberts and Richard~L. Tweedie.
\newblock Exponential convergence of langevin distributions and their discrete approximations.
\newblock \emph{Bernoulli}, 2\penalty0 (4):\penalty0 341--363, December 1996.

\bibitem[Rombach et~al.(2022)Rombach, Blattmann, Lorenz, Esser, and Ommer]{rombach2022high}
Robin Rombach, Andreas Blattmann, Dominik Lorenz, Patrick Esser, and Bj{\"o}rn Ommer.
\newblock High-resolution image synthesis with latent diffusion models.
\newblock In \emph{Proceedings of the IEEE/CVF conference on computer vision and pattern recognition}, pages 10684--10695, 2022.

\bibitem[Rosenthal et~al.(2011)]{rosenthal2011optimal}
Jeffrey~S Rosenthal et~al.
\newblock Optimal proposal distributions and adaptive mcmc.
\newblock \emph{Handbook of Markov Chain Monte Carlo}, 4\penalty0 (10.1201), 2011.

\bibitem[Song et~al.(2017)Song, Zhao, and Ermon]{song2017nice}
Jiaming Song, Shengjia Zhao, and Stefano Ermon.
\newblock A-nice-mc: Adversarial training for mcmc.
\newblock \emph{Advances in neural information processing systems}, 30, 2017.

\bibitem[Song and Ermon(2019)]{song2019generative}
Yang Song and Stefano Ermon.
\newblock Generative modeling by estimating gradients of the data distribution.
\newblock \emph{Advances in neural information processing systems}, 32, 2019.

\bibitem[Song et~al.(2020)Song, Garg, Shi, and Ermon]{song2020sliced}
Yang Song, Sahaj Garg, Jiaxin Shi, and Stefano Ermon.
\newblock Sliced score matching: A scalable approach to density and score estimation.
\newblock In \emph{Uncertainty in Artificial Intelligence}, pages 574--584. PMLR, 2020.

\bibitem[Song et~al.(2021)Song, Sohl-Dickstein, Kingma, Kumar, Ermon, and Poole]{song2020score}
Yang Song, Jascha Sohl-Dickstein, Diederik~P Kingma, Abhishek Kumar, Stefano Ermon, and Ben Poole.
\newblock Score-based generative modeling through stochastic differential equations.
\newblock In \emph{International Conference on Learning Representations}, 2021.

\bibitem[Titsias and Dellaportas(2019)]{titsias2019gradient}
Michalis Titsias and Petros Dellaportas.
\newblock Gradient-based adaptive markov chain monte carlo.
\newblock \emph{Advances in neural information processing systems}, 32, 2019.

\bibitem[Vincent(2011)]{vincent2011connection}
Pascal Vincent.
\newblock A connection between score matching and denoising autoencoders.
\newblock \emph{Neural computation}, 23\penalty0 (7):\penalty0 1661--1674, 2011.

\bibitem[Wang et~al.(2024)Wang, Liu, Smith, and Atchade]{wang2024cyclical}
Liwei Wang, Xinru Liu, Aaron Smith, and Aguemon~Y Atchade.
\newblock On cyclical mcmc sampling.
\newblock In \emph{International Conference on Artificial Intelligence and Statistics}, pages 3817--3825. PMLR, 2024.

\bibitem[Wu et~al.(2022)Wu, Schmidler, and Chen]{wu2022minimax}
Keru Wu, Scott Schmidler, and Yuansi Chen.
\newblock Minimax mixing time of the metropolis-adjusted langevin algorithm for log-concave sampling.
\newblock \emph{Journal of Machine Learning Research}, 23\penalty0 (270):\penalty0 1--63, 2022.

\bibitem[Zhang et~al.(2024)Zhang, Yin, Liang, and Liu]{pmlr-v235-zhang24bv}
Kaihong Zhang, Heqi Yin, Feng Liang, and Jingbo Liu.
\newblock Minimax optimality of score-based diffusion models: Beyond the density lower bound assumptions.
\newblock In \emph{Proceedings of the 41st International Conference on Machine Learning}, volume 235 of \emph{Proceedings of Machine Learning Research}, pages 60134--60178. PMLR, 21--27 Jul 2024.

\end{thebibliography}
\clearpage 
\onecolumn
\appendix
\textbf{\Huge{Appendix}}

\appendix

In this appendix, we first discuss an alternative approximation of the acceptance function that relies solely on the score functions, derived using the Taylor expansion of the log densities. Next, we consider sampling with score-based MALA for heavy-tailed distributions. Third, we provide the proofs for the theoretical results. Finally, we present additional details of the experiments and further empirical results.

\section{Taylor Score-based Metropolis-Hastings}
\label{sec:taylor}
In this section, we propose an approximation of the acceptance function  based on the Taylor expansion of the log densities.

For an acceptance function $a(x'x)$ as defined in Equation \ref{eqn:ratio_acceptance}, the ratio $r(x',x) = \frac{p(x') q(x | x')}{p(x) q(x' | x)}$ can be rewritten as:
\begin{equation}
\label{eqn:log_acceptance_ratio}
\log \left( r(x',x)\right) = \log p(x') - \log p(x) + \log q(x \mid x') - \log q(x' \mid x)
\end{equation}

Let \( \tau = x' - x \). Assuming that \( \|\tau\| \) is sufficiently small, we perform a second-order Taylor series expansion of the logarithmic terms around the point \( x \).

\paragraph{Taylor Expansion of \( \log p(x') \) Around \( x \):}
Expanding \( \log p(x') = \log p(x + \tau) \) using a second-order Taylor series:
\begin{equation}
\label{eqn:second_order_taylor}
\log p(x + \tau) = \log p(x) + \nabla \log p(x) \cdot \tau + \frac{1}{2} \tau^\top \nabla^2 \log p(x) \, \tau + o(\|\tau\|^2)
\end{equation}
Therefore,
\begin{equation}
\label{eqn:second_order_taylor_rearranged}
\log p(x') - \log p(x) = \nabla \log p(x) \cdot \tau + \frac{1}{2} \tau^\top \nabla^2 \log p(x) \, \tau + o(\|\tau\|^2)
\end{equation}

\paragraph{Taylor Expansion of \( \log p(x) \) Around \( x' \):}
Similarly, expanding \( \log p(x) = \log p(x' - \tau) \):
\begin{equation}
\label{eqn:second_order_taylor_prime}
\log p(x' - \tau) = \log p(x') - \nabla \log p(x') \cdot \tau + \frac{1}{2} \tau^\top \nabla^2 \log p(x') \, \tau + o(\|\tau\|^2)
\end{equation}
Rearranging gives:
\begin{equation}
\label{eqn:second_order_taylor_prime_rearranged}
\log p(x) - \log p(x') = -\nabla \log p(x') \cdot \tau + \frac{1}{2} \tau^\top \nabla^2 \log p(x') \, \tau + o(\|\tau\|^2)
\end{equation}

\paragraph{Final Result}
Therefore, we have that,
\[
\begin{aligned}
\log p(x') - \log p(x) &= \frac{1}{2} \left( \log p(x') - \log p(x) \right) + \frac{1}{2} \left( \log p(x') - \log p(x) \right) \\
&= \frac{1}{2} \left[ \nabla \log p(x) \cdot \tau + \nabla \log p(x') \cdot \tau \right] \\
&\quad + \frac{1}{4} \tau^\top \left( \nabla^2 \log p(x) - \nabla^2 \log p(x') \right) \tau + o(\|\tau\|^2)
\end{aligned}
\]
Simplifying, we obtain:
\begin{equation}
\label{eqn:averagin}  
\log p(x') - \log p(x) = \frac{1}{2} \left( \nabla \log p(x) + \nabla \log p(x') \right) \cdot \tau + \frac{1}{4} \tau^\top \left( \nabla^2 \log p(x) - \nabla^2 \log p(x') \right) \tau + o(\|\tau\|^2)
\end{equation}

\begin{figure*}[ht]
    \centering
    \begin{subfigure}[b]{0.32\textwidth}
        \includegraphics[width=\textwidth]{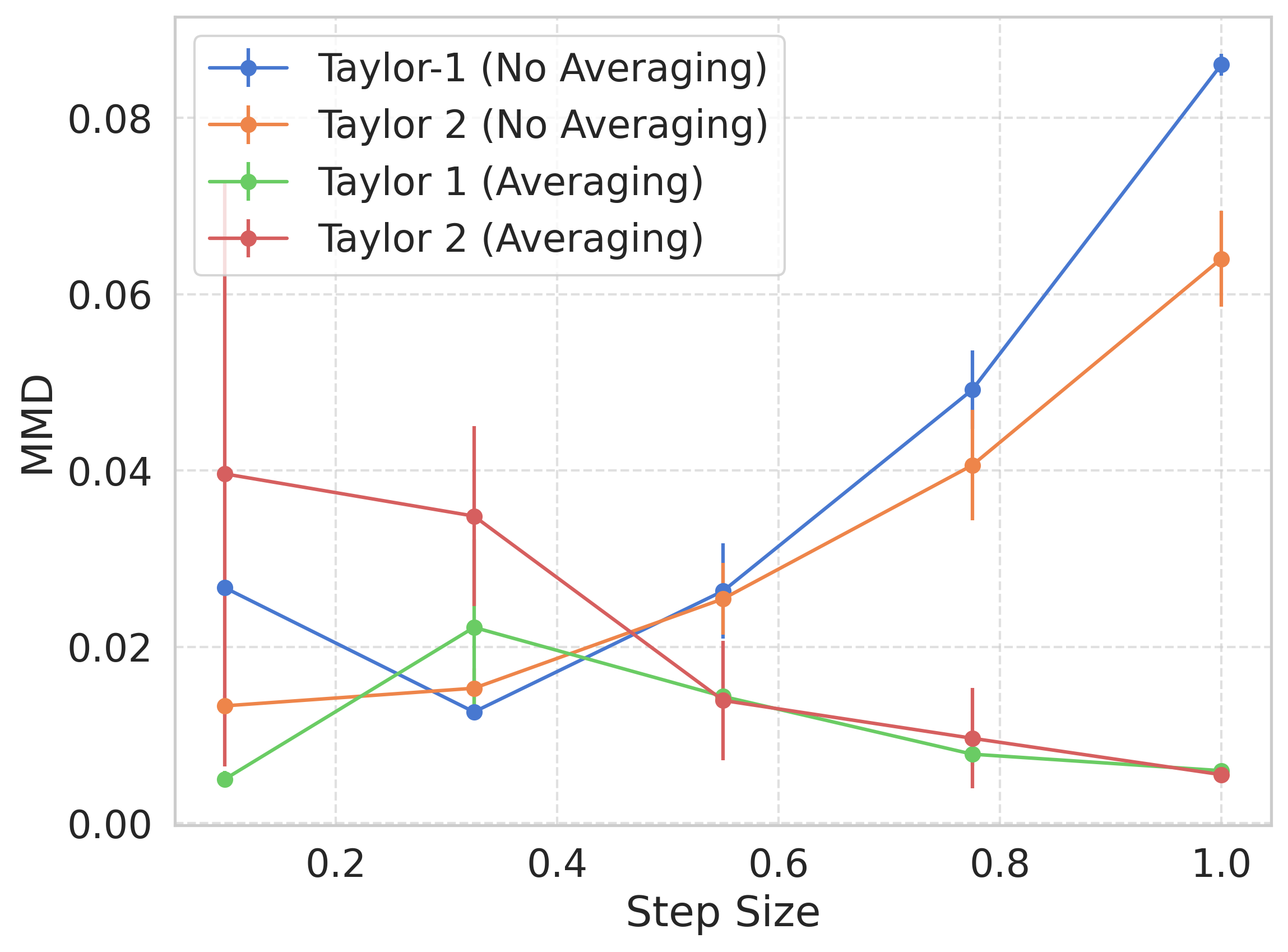}
        \caption{MMD Comparison}
        \label{fig:mmd_comparison}
    \end{subfigure}
    \hfill
    \begin{subfigure}[b]{0.32\textwidth}
        \includegraphics[width=\textwidth]{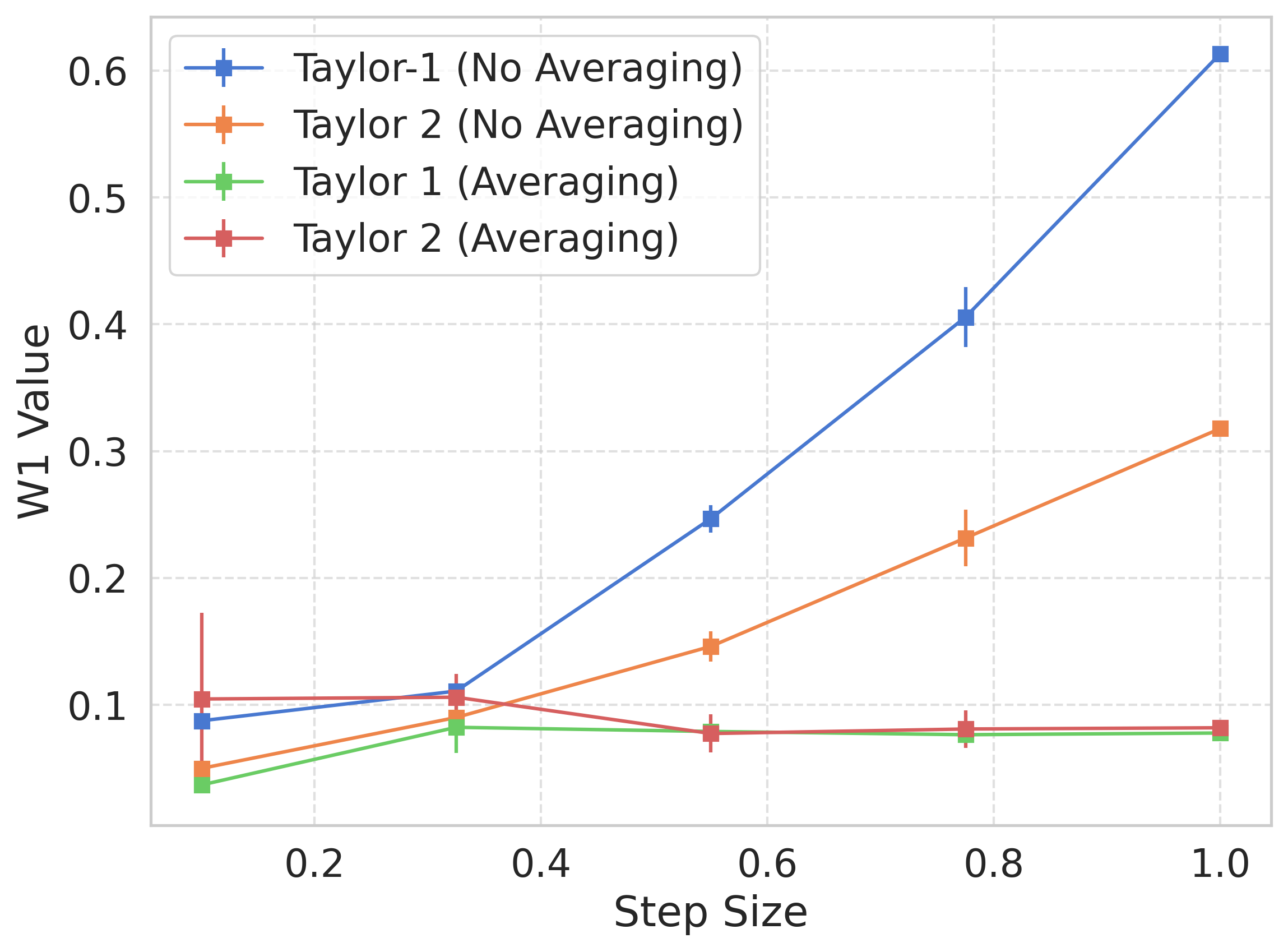}
        \caption{W1 Comparison}
        \label{fig:w1_comparison}
    \end{subfigure}
    \hfill
    \begin{subfigure}[b]{0.32\textwidth}
        \includegraphics[width=\textwidth]{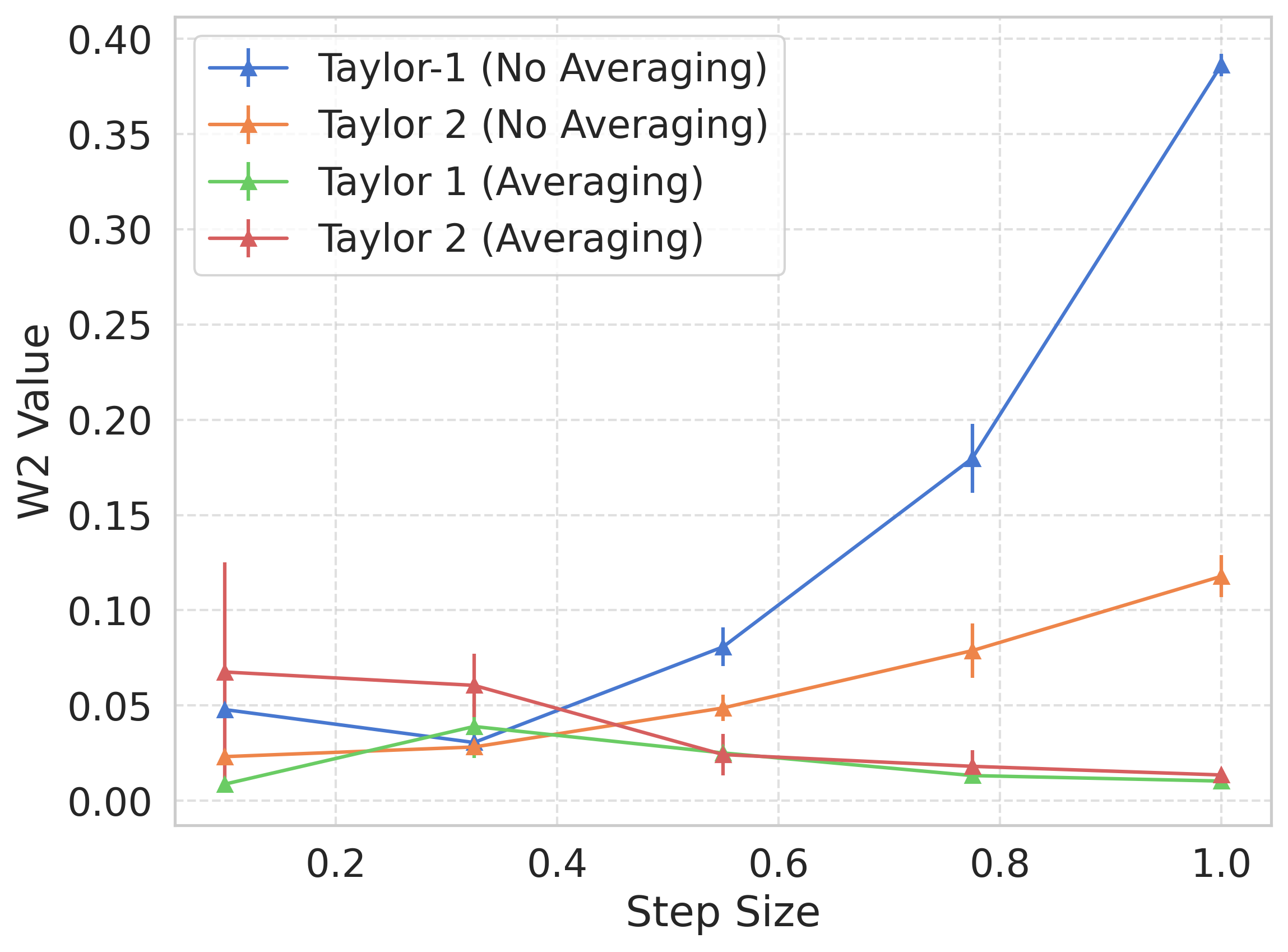}
        \caption{W2 Comparison}
        \label{fig:w2_comparison}
    \end{subfigure}
    \caption{Comparison of Taylor approximations: Taylor-1 and Taylor-2 methods (with and without averaging) evaluated using MMD, W1, and W2 metrics. The experiments were conducted on the Moons dataset.}
    \label{fig:taylor_comparison}
\end{figure*}

Combining the above results, we have:
\[
\log r(x',x) = \frac{1}{2} \left( \nabla \log p(x) + \nabla \log p(x') \right) \cdot \tau + \frac{1}{4} \tau^\top \left( \nabla^2 \log p(x) - \nabla^2 \log p(x') \right) \tau + o(\|\tau\|^2) + \log q(x \mid x') - \log q(x' \mid x) 
\]
If the distance between the Hessian is negligible, we can approximate the acceptance function as
\[
\log r(x',x) \approx \frac{1}{2} \left( \nabla \log p(x) + \nabla \log p(x') \right) \cdot \tau  + \left[ \log q(x \mid x') - \log q(x' \mid x) \right]
\]
We denote this approximation of the acceptance function as Taylor-1 (Averaging). 
Figure \ref{fig:taylor_comparison} illustrates how this approximation can provide a better result compared to the first-order and second-order Taylor series from Equation \ref{eqn:second_order_taylor_rearranged}, while achieving very similar performance to the original Equation \ref{eqn:averagin}, which utilizes the Hessian terms. This is particularly important because computing the Hessian numerically (by backpropagation through a score model) is computationally expensive and unstable.

\begin{figure}[ht]
    \centering
    \includegraphics[width=0.5\linewidth]{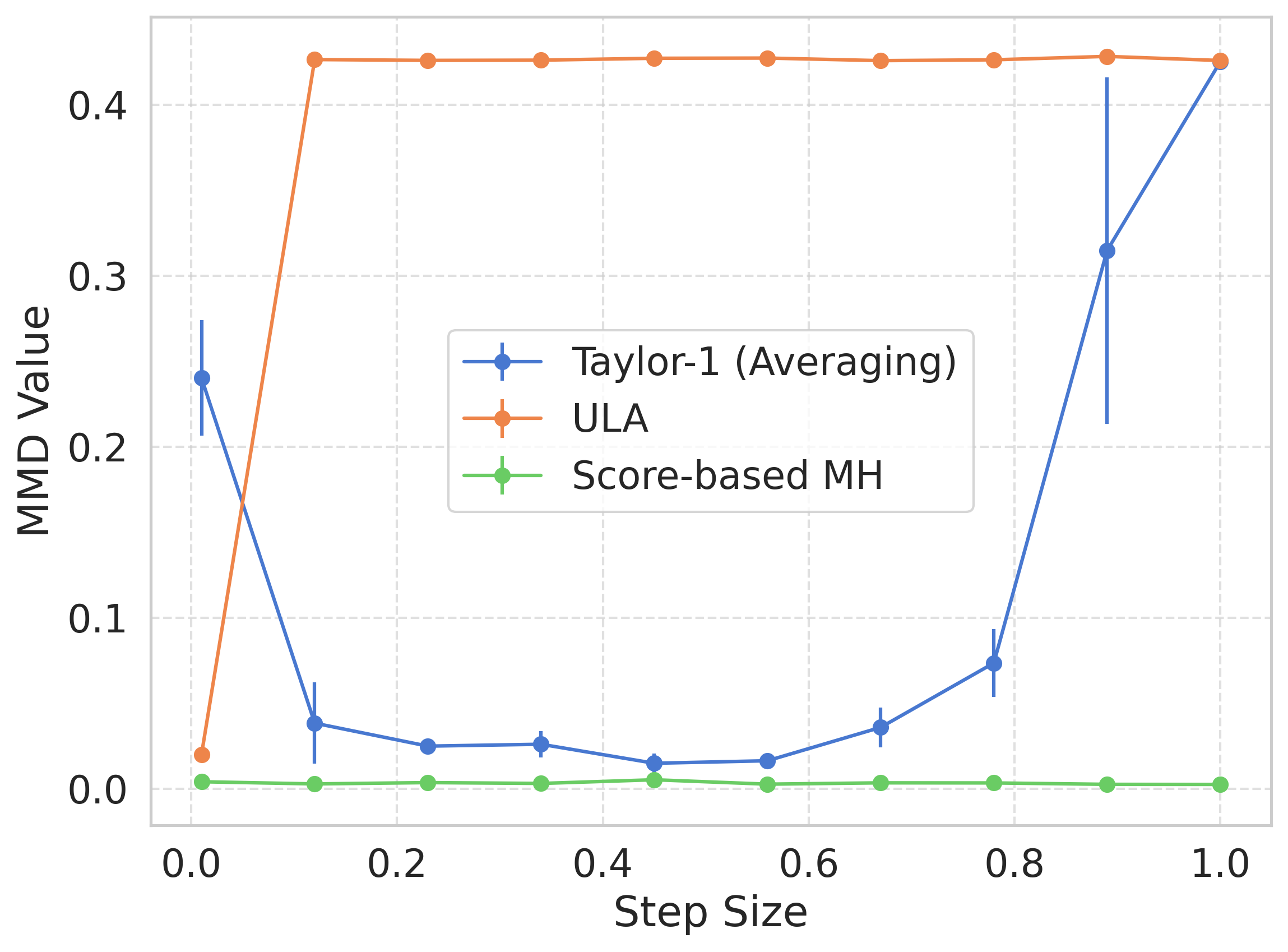}
    \caption{MMD distance comparison between ULA, Taylor-1 (Averaging), and Score-based MH.}
    \label{fig:acceptance_vs_taylor}
\end{figure}

Additionally, Figure \ref{fig:acceptance_vs_taylor} compare the performance of Taylor-1 (Averaging) to score-based Metropolis-Hastings. We can see that for some range the Taylor approximation provides a very good approximation of the Acceptance function as it yields good sampling quality. However, even though some robustness is observed score-based MALA still outperforms the Taylor approximation. 

\section{Metropolis-Hastings and Generalized Extreme Value Distributions}
\label{sec:heavy_tails}
In this section, we study the effect of sampling using ULA, MALA, and score-based MALA algorithms. We analyze the impact of the adjustment step on distributions with heavy tails and hypothesize that as the tail becomes heavier, the adjustment step becomes increasingly more important. This observation suggests that, to accurately model and generate extreme events using score-based models, incorporating the adjustment step is essential.

\paragraph{Generalized Extreme Value (GEV) Distribution}

The Generalized Extreme Value (GEV) distribution is commonly used to model the maxima of datasets and is defined by its probability density function (PDF):

\[
f(x; \xi, \mu, \sigma) = 
\begin{cases} 
\frac{1}{\sigma} \left( 1 + \xi \frac{x - \mu}{\sigma} \right)^{-1 - \frac{1}{\xi}} \exp \left( - \left( 1 + \xi \frac{x - \mu}{\sigma} \right)^{-\frac{1}{\xi}} \right), & \text{if } \xi \neq 0, \\
\frac{1}{\sigma} \exp \left( -\frac{x - \mu}{\sigma} \right) \exp \left( - \exp \left( -\frac{x - \mu}{\sigma} \right) \right), & \text{if } \xi = 0,
\end{cases}
\]

where:
- \( \mu \) is the location parameter,
- \( \sigma > 0 \) is the scale parameter,
- \( \xi \) is the shape parameter (controls the tail heaviness).

The GEV distribution unifies three types of distributions: Gumbel (\(\xi = 0\)), Fr\'echet (\(\xi > 0\)), and Weibull (\(\xi < 0\)), making it highly flexible for extreme value analysis.

For the Fr\'echet distribution, moments of order \( k \) exist only if \( k < \frac{1}{\xi} \). If \( k \geq \frac{1}{\xi} \), the moment diverges, leading to infinite values. This property makes the Fr\'echet distribution particularly suited for modeling heavy-tailed distributions, as it captures the behavior of distributions with heavy tails and undefined higher-order moments.

\begin{figure}[ht]
    \centering
    \includegraphics[width=0.8\textwidth]{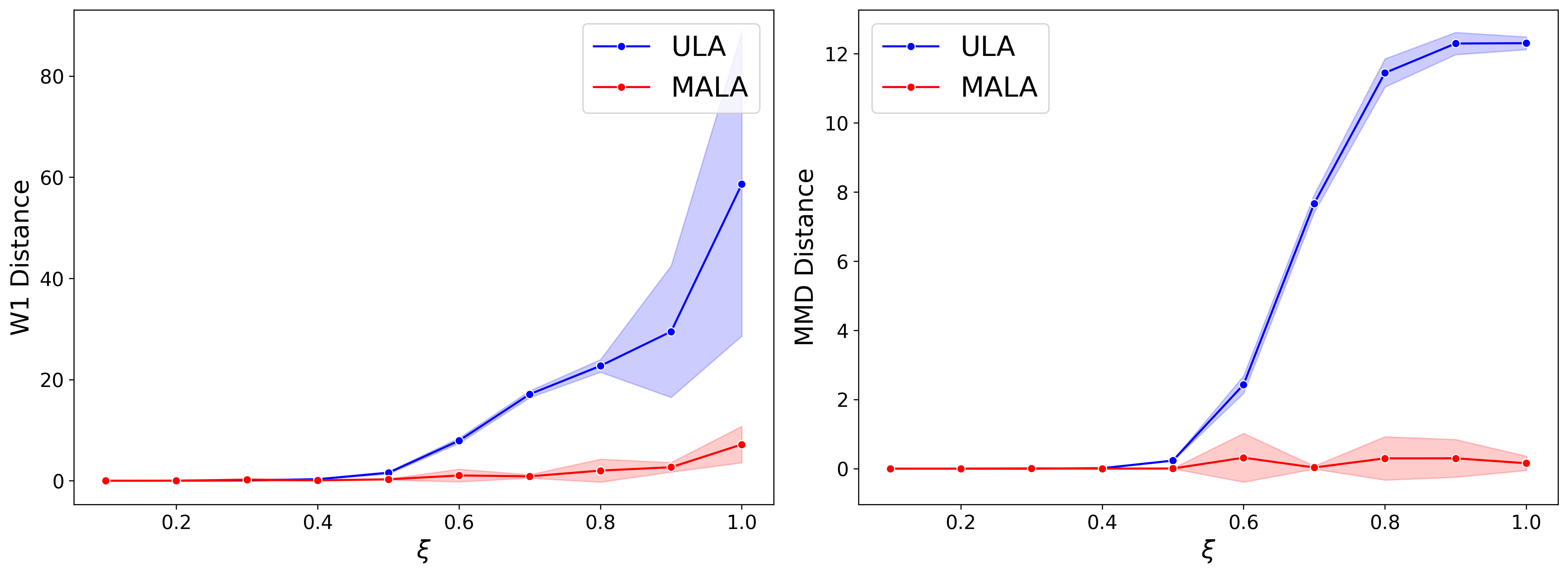}
    \caption{Comparison between ULA and MALA sampling methods for GEV$(0,1,\xi)$, highlighting the effect of the adjustment step on heavy-tailed distributions. We fix the step size of both ULA and MALA to $0.1$.} 
    \label{fig:ula_mala_comparison}
\end{figure}

\begin{figure}[ht]
    \centering
    \begin{subfigure}[b]{0.49\textwidth}
        \centering
        \includegraphics[width=\textwidth]{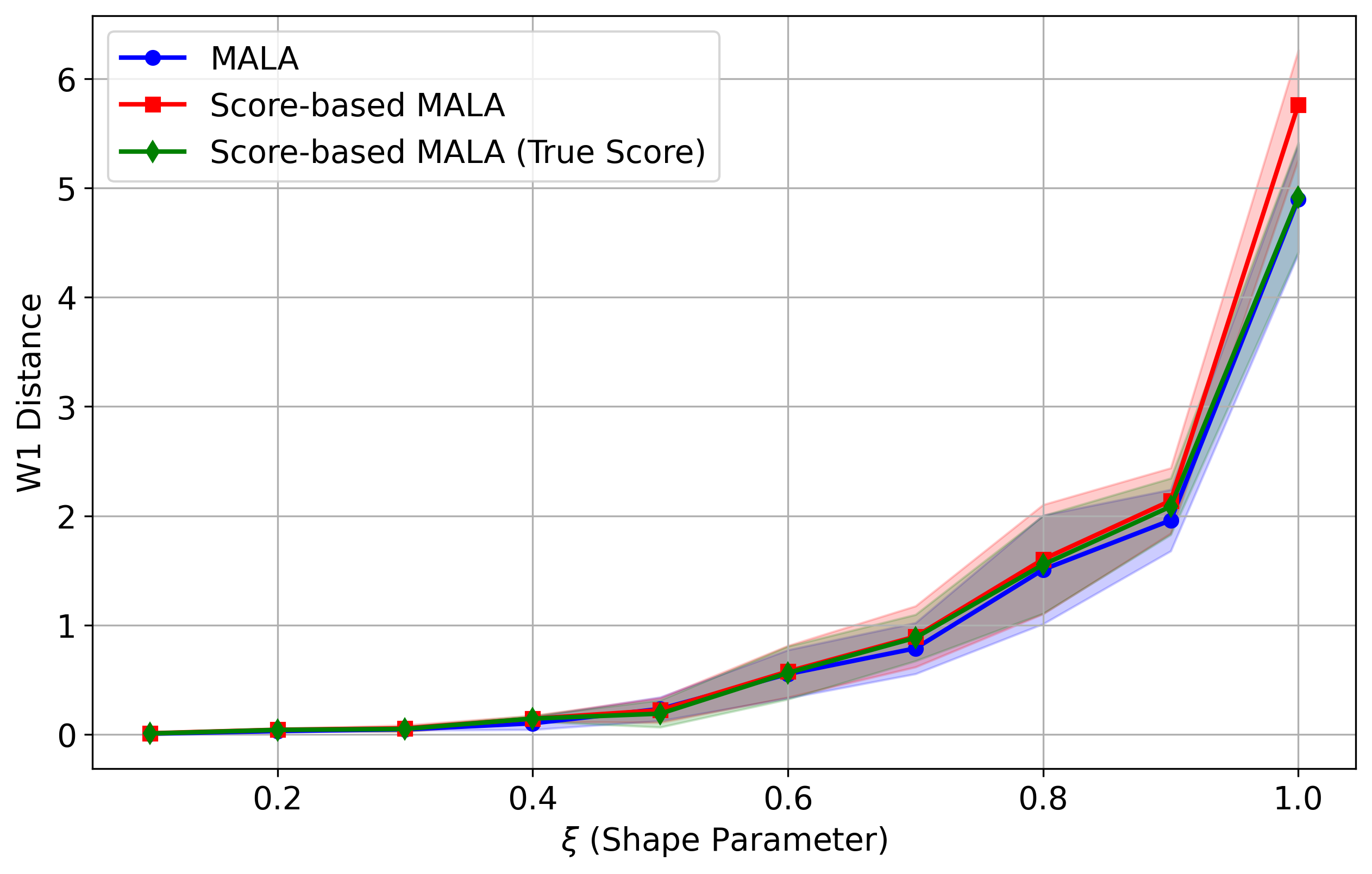}
        \caption{W1 Distance comparison.}
    \end{subfigure}
    \hfill
    \begin{subfigure}[b]{0.49\textwidth}
        \centering
        \includegraphics[width=\textwidth]{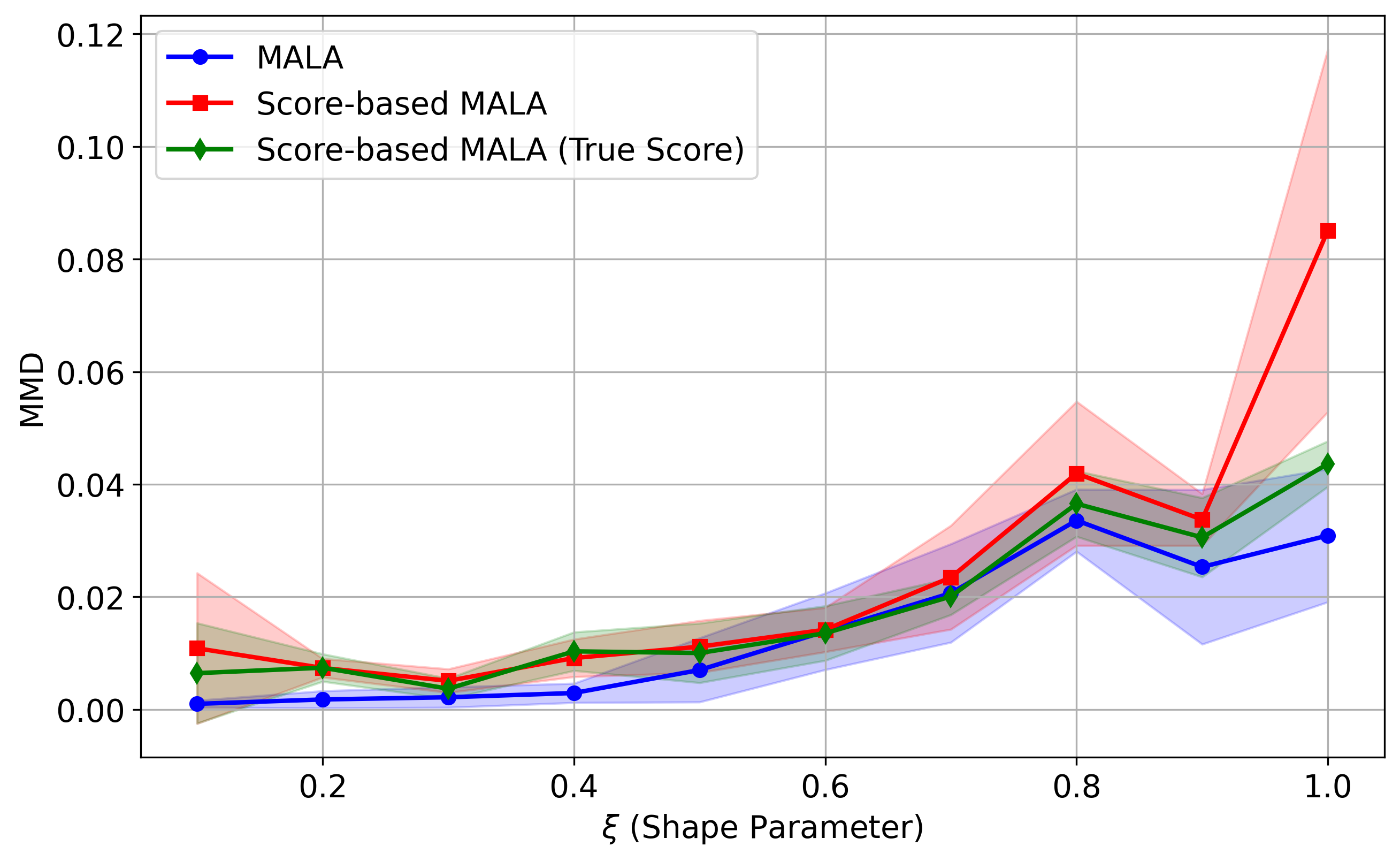}
        \caption{MMD comparison.}
    \end{subfigure}
    \caption{Comparison between score-based MALA (the score network is trained from data), score-based MALA (True Score) where the acceptance network is trained with the true score function, and the original MALA algorithm. Left: W1 distance. Right: MMD.}
    \label{fig:score_based_mala_comparison}
\end{figure}

In Figure\ref{fig:score_based_mala_comparison}, we illustrate that score-based MALA achieves competitive performance compared to the ground truth MALA. While MALA uses the standard acceptance function defined in Equation~\ref{eqn:ratio_acceptance}, score-based MALA learns a potentially different acceptance function. 

\section{Proofs}
\label{sec:proofs}
\begin{proof}[Proof of Proposition 1]
We aim to prove that 
$$\mathcal{L}(a) = 0 \iff a \in \mathcal{A}.
$$

\textbf{(Reverse Direction:)} 
Suppose \( a \in \mathcal{A} \), meaning that for every $x,x' \in \mathcal{X}$, \( a(x', x) \) satisfies
\[
\frac{a(x', x)}{a(x, x')} = \frac{p(x') q(x \mid x')}{p(x) q(x' \mid x)}.
\]
Taking the logarithm on both sides,
\[
\log a(x', x) - \log a(x, x') = \log p(x') - \log p(x) + \log q(x \mid x') - \log q(x' \mid x).
\]
Differentiating both sides with respect to \( x \) and \( x' \), we obtain
\[
\nabla \log a(x', x) - \nabla \log a(x, x') = \nabla \log p(x') - \nabla \log p(x) + \nabla \log q(x \mid x') - \nabla \log q(x' \mid x).
\]
as we have that:     
$$\mathcal{L}(a) = \mathbb{E}_{x \sim p,x'\sim q(\cdot|x)}\bigg[ \left\| \nabla \log a(x', x) - \nabla \log a(x, x')  -  \nabla \log p(x') + \nabla \log p(x) - \nabla \log q(x|x') + \nabla \log q(x'|x) \right\|^2 \bigg]
$$
This implies \( \mathcal{L}(a) = 0 \).

\textbf{(Forward Direction:)}

Suppose that $\mathcal{L}(a) = 0$, i.e., as the expectation is taken over a positive variable we have that almost surely, for every $x,x' \in \mathcal{X}$
$$
\begin{aligned}
& \nabla \log a(x', x) - \nabla \log a(x, x') = \nabla \log p(x') - \nabla \log p(x) + \nabla \log q(x \mid x') - \nabla \log q(x' \mid x),
\end{aligned}
$$
then integrating this equality gives:
$$
\begin{aligned}
& \log a(x', x) - \log a(x, x')= \log p(x') - \log p(x) + \log q(x \mid x') - \log q(x' \mid x) + C,
\end{aligned}
$$
where $C \in \mathbb{R}$ is a constant. Therefore, the acceptance function $a(x, x')$ can be expressed as:
$$
\frac{a(x', x)}{a(x,x')} = e^C \cdot \frac{p(x') q(x \mid x')}{p(x) q(x' \mid x)}.
$$
as $e^C $ does not depend on $x$, setting $x=x'$, we get that 
\[1 = e^C\]
Therefore we have that $C=0$. 
Hence, $a \in \mathcal{A}$, which concludes the proof.
\end{proof}

\begin{proof}[Proof of Proposition 2]
Let \( \hat{\mathcal{L}}(a) \) be the empirical SBM loss computed over \( N \) i.i.d. training samples:
\[
\hat{\mathcal{L}}(a) = \frac{1}{N} \sum_{i=1}^{N} \left\| \nabla \log a(x_i', x_i) - \nabla \log a(x_i, x_i') - \nabla \log p(x_i') + \nabla \log p(x_i) - \nabla \log q(x_i \mid x_i') + \nabla \log q(x_i' \mid x_i) \right\|^2.
\]
The true expected loss is
\[
\mathcal{L}(a) = \mathbb{E}_{x \sim p, x' \sim q(\cdot|x)} \left[ \left\| \nabla \log a(x', x) - \nabla \log a(x, x') - \nabla \log p(x') + \nabla \log p(x) - \nabla \log q(x \mid x') + \nabla \log q(x' \mid x) \right\|^2 \right].
\]

All we need to prove is that the loss function is bounded in order to apply the standard uniform convergence bound from statistical learning theory~\citep{mohri2018foundations}.

\textbf{Proving the boundedness of the loss function.}  
By Assumption~\ref{ass:compact}, \(\mathcal{X}\) is compact, which implies that \(\mathcal{X} \times \mathcal{X}\) is also compact. Furthermore, by Assumptions~\ref{ass:cont} and~\ref{ass:hyp_bound}, the gradient terms \(\nabla \log a\), \(\nabla \log p\), and \(\nabla \log q\) are continuous over a compact domain. By the Extreme Value Theorem, they attain their maximum absolute values, denoted by constants \(C_1, C_2,\) and \(C_3\), respectively:
\[
\sup_{x, x' \in \mathcal{X}} \|\nabla \log a(x', x)\| \leq C_1, \quad \sup_{x, x' \in \mathcal{X}} \|\nabla \log p(x)\| \leq C_2, \quad \sup_{x, x' \in \mathcal{X}} \|\nabla \log q(x' \mid x)\| \leq C_3.
\]

Thus, the SBM loss is bounded as follows:
\[
\mathcal{L}(a) \leq \mathbb{E}_{x \sim p, x' \sim q(\cdot|x)} \left[ (2C_1 + 2C_2 + 2C_3)^2 \right] = C_4,
\]
where \( C_4 = (2C_1 + 2C_2 + 2C_3)^2 \) is a finite constant.

Applying standard uniform convergence bounds~\citep{mohri2018foundations}, for all \( a \in \mathcal{F} \), with probability at least \( 1 - \delta \), we obtain
\[
\sup_{a \in \mathcal{F}} \left| \mathcal{L}(a) - \hat{\mathcal{L}}(a) \right| \leq 2 \mathcal{R}_N(\mathcal{F}) + \mathcal{O}\left(\sqrt{\frac{\log(1/\delta)}{N}}\right),
\]
where \( \mathcal{R}_N(\mathcal{F}) \) is the Rademacher complexity of the function class \( \mathcal{F} \).
\end{proof}

\begin{proof}[Proof of Proposition 3]
Suppose that $\mathcal{L}(a) = 0$, Let $a_M = \frac{a}{M}$, we prove in Proposition 1 that $\mathcal{L}(a) = 0$ is equivalent to $a\in \mathcal{A}$, hence we will prove that $a_M\in \mathcal{A}$.

    To prove that the acceptance function \(a_M(x, x')\) satisfies the detailed balance condition, we need to show that:
    \[
    p(x) q(x' \mid x) a_M(x', x) = p(x') q(x \mid x') a_M(x, x').
    \]

    Substituting the definition of the acceptance function \(a_M\), we consider two cases based on the value of the expression \(\frac{p(x') q(x \mid x')}{p(x) q(x' \mid x)}\). Without loss of generality we consider the case when \(\frac{p(x') q(x \mid x')}{p(x) q(x' \mid x)} \leq 1\), then by definition:
    \[
    a_M(x', x) = \frac{1}{M} \frac{p(x') q(x \mid x')}{p(x) q(x' \mid x)}.
    \]
    Substituting this into the detailed balance equation, we have:
    \[
    p(x) q(x' \mid x) \cdot \frac{1}{M} \frac{p(x') q(x \mid x')}{p(x) q(x' \mid x)} = \frac{1}{M} p(x') q(x \mid x').
    \]

    Similarly, since \(\frac{p(x) q(x' \mid x)}{p(x') q(x \mid x')} \leq 1\) in this case, we have:
    \[
    a_M(x, x') = \frac{1}{M}.
    \]
    Thus, the right-hand side of the detailed balance condition becomes:
    \[
    \frac{1}{M}p(x') q(x \mid x')
    \]
    Therefore, both sides are equal:
    \[
    \frac{1}{M} p(x') q(x \mid x') = \frac{1}{M} p(x') q(x \mid x'),
    \]
    which confirms that the detailed balance condition holds in this case.

Therefore, $a_M \in \mathcal{A}$. Hence, $\mathcal{L}(\frac{a}{M}) = 0$, which concludes the proof.
\end{proof}

\section{Experiments Details}
In this section, we provide descriptions of the datasets used to generate the empirical results from the main text, and outline the neural network architectures and hyperparameter choices. All presented empirical results were compiled using an NVIDIA RTX 3090 GPU.

\subsection{Dataset Descriptions} \label{sec:dataset_descriptions}

We use four datasets generated using \texttt{scikit-learn}~\citep{scikit-learn} and custom code. The parameters provided ensure reproducibility of the results:

\begin{itemize}
    \item \textbf{Moons}: This dataset consists of two interlocking crescent-shaped clusters. We generate \(10000\) samples with a noise level of \(0.1\) using \texttt{sklearn.datasets.make\_moons(n\_samples=10000, noise=0.1)}.
    
    \item \textbf{Pinwheel}: Data points are arranged in six spiral arms. We generate $10000$ samples with a radial standard deviation of $0.5$, tangential standard deviation of \(0.05\), and a rate of \(0.25\) (which controls the spread of the arms). The dataset is generated using custom code and the following parameters: 
    \begin{itemize}
        \item \texttt{num\_classes=5}, 
        \item \texttt{radial\_std=0.5}, 
        \item \texttt{tangential\_std=0.05}, 
        \item \texttt{rate=0.25}.
    \end{itemize}
\setcounter{algorithm}{2}
\begin{algorithm}
\caption{Pinwheel Dataset Generation}
\begin{algorithmic}[1]
    \State \textbf{Input:} Number of samples $n$, number of classes $K$, radial standard deviation $\sigma_r$, tangential standard deviation $\sigma_t$, rotation rate $\alpha$
    \State \textbf{Output:} Dataset $\mathbf{X} \in \mathbb{R}^{n \times 2}$
    
    \State Generate random class labels $\mathbf{y} \in \{0, 1, \dots, K-1\}$ for $n$ samples
    \State Compute angles $\theta_k = \frac{2\pi k}{K}$ for each class $k \in \{0, 1, \dots, K-1\}$
    \State Generate radial components $r_i \sim \mathcal{N}(1, \sigma_r^2)$ for each sample $i$
    \State Compute tangential noise $\delta_i \sim \mathcal{N}(0, \sigma_t^2)$ for each sample $i$
    \State Compute angle for each point: $\phi_i = \theta_{y_i} + \alpha \cdot r_i$
    \State Compute Cartesian coordinates:
    \[
    x_i = r_i \cdot \cos(\phi_i) + \delta_i
    \]
    \[
    y_i = r_i \cdot \sin(\phi_i) + \delta_i
    \]
    \State Return the dataset $\mathbf{X} = \{(x_i, y_i)\}_{i=1}^{n}$
\end{algorithmic}
\end{algorithm}

    \item \textbf{S-curve}: This dataset consists of \(10000\) points distributed along an "S"-shaped 3D manifold with added Gaussian noise of \(0.1\). It is generated using \texttt{sklearn.datasets.make\_s\_curve(n\_samples=10000, noise=0.1)}.

    \item \textbf{Swiss Roll}: This dataset contains \(10000\) points arranged along a 3D spiral-shaped surface, with Gaussian noise of \(0.5\) added. It is generated using \texttt{sklearn.datasets.make\_swiss\_roll(n\_samples=10000, noise=0.5)}.
    \item \textbf{MNIST}: MNIST consists of grayscale images of handwritten digits (0-9), each of size \(28 \times 28\). The dataset contains \(60,000\) training samples and \(10,000\) test samples. For our experiments, we normalize pixel values to \([-1,1]\) and convert the images into PyTorch tensors.

\end{itemize}

All datasets are converted to PyTorch tensors for further processing. The experiments were performed using an NVIDIA RTX 3090 GPU.

\subsection{Neural Network Architectures} \label{sec:network_architectures}
We now provide detailed descriptions of several different neural network architectures designed for learning the score function (Score Nets), and the acceptance networks (Acceptance Nets).

\paragraph{Score Nets.}
The Score Nets architecture is a simple feed-forward neural network consisting of an input layer, two hidden layers, and an output layer. The input dimension is the same as the output dimension, ensuring that the output matches the shape of the input data. The \texttt{Softplus} activation function is applied after each hidden layer to introduce non-linear transformations. The output layer does not use an activation function, as it directly produces the estimated score of the input. 

\paragraph{Acceptance Nets.}
The Acceptance Nets architecture is a feed-forward neural network designed to compute the acceptance function for a pair of inputs. The network takes two input vectors, concatenates them, and passes the result through an initial fully connected layer with a customized hidden dimension. Following this, the network consists of three residual blocks, each containing fully connected layers with the same hidden dimension, enhanced by \texttt{GELU} activation functions to improve non-linearity and gradient flow. Finally, the output is passed through a fully connected layer, followed by \texttt{GELU} and a \texttt{Sigmoid} activation to ensure the output is between 0 and 1, representing the acceptance probability.

\paragraph{MNIST Architectures.} For the MNIST Score Net, we employ a time-dependent UNet architecture with an encoder and decoder blocks and we use Softplus activations functions. For further details on the Score Network, please refer to the MNIST tutorial available at the following GitHub repository, as we use the same architecture \url{https://github.com/yang-song/score_sde}. For its acceptance network we use a Siamese like and that takes two inputs \((x, x')\), embeds them via convolutional and residual layers, and processes them alongside learned time and step-size embeddings. We use GeLU activation functions. These branches do not share weights. The representation are then fed to a common branch. The acceptance network also encodes the time and step size parameters. 
 
\paragraph{Annealed ULA and MALA for MNIST.}
We evaluate annealed ULA and annealed MALA for sampling high-quality image reconstructions. We employ a denoisnig score matchign appraoch as presented in ~\citep{song2019generative,song2020score}.

\subsection{Hyperparameters}
\paragraph{Score Nets.}
We summarize the huyperparameters used to train the Score Nets in Table \ref{tab:scorenet_hyperparameters}.
\begin{table*}[h]
\caption{Score-Net Training Hyperparameters}
\label{tab:scorenet_hyperparameters}
\begin{center}
\begin{tabular}{c|c|c|c|c}
\hline
\textbf{Dataset} & \textbf{Optimizer} & \textbf{Learning Rate (LR)} & \textbf{Epochs} & \textbf{Hidden Dimension} \\
\hline
Moons & Adam & $1 \times 10^{-3}$ & 5000 & 64 \\
Pinwheel & Adam & $5 \times 10^{-4}$ & 2000 & 512 \\
S-curve & Adam & $5 \times 10^{-4}$ & 2000 & 512\\
Swiss Roll & Adam & $5 \times 10^{-4}$ & 2000 & 512\\
\hline
\end{tabular}
\end{center}
\end{table*}

\paragraph{Acceptance Nets.} We provdie below the detailed for training the Acceptance network for the various score-based Metropolis-Hastings algorithms. 

\begin{table*}[h]
\caption{Acceptance-Net Training Hyperparameters}
\label{tab:acceptancenet_hyperparameters}
\begin{center}
\begin{tabular}{c|c|c|c|c|c|c}
\hline
\textbf{Dataset} & \textbf{Optimizer} & \textbf{LR} & \textbf{Hidden Dimension} & \textbf{Residual Layers} &\textbf{Epochs}&\(\lambda\) \\
\hline
 Moons & Adam & $5 \times 10^{-4}$ & 256 & 3 &1000 & 2\\
 Pinwheel & Adam & $5 \times 10^{-4}$ & 256 & 4 &200&2 \\
S-curve & Adam & $5 \times 10^{-4}$ & 512 & 4 &200&2 \\
Swiss Roll & Adam & $5 \times 10^{-4}$ & 512 & 4 &200&1 \\
\hline
\end{tabular}
\end{center}
\end{table*}

\section{Additional Empirical Results}
\label{sec:experiments}
\begin{figure*}[t]
    \centering
    \begin{subfigure}[b]{0.32\textwidth}
        \includegraphics[width=\textwidth]{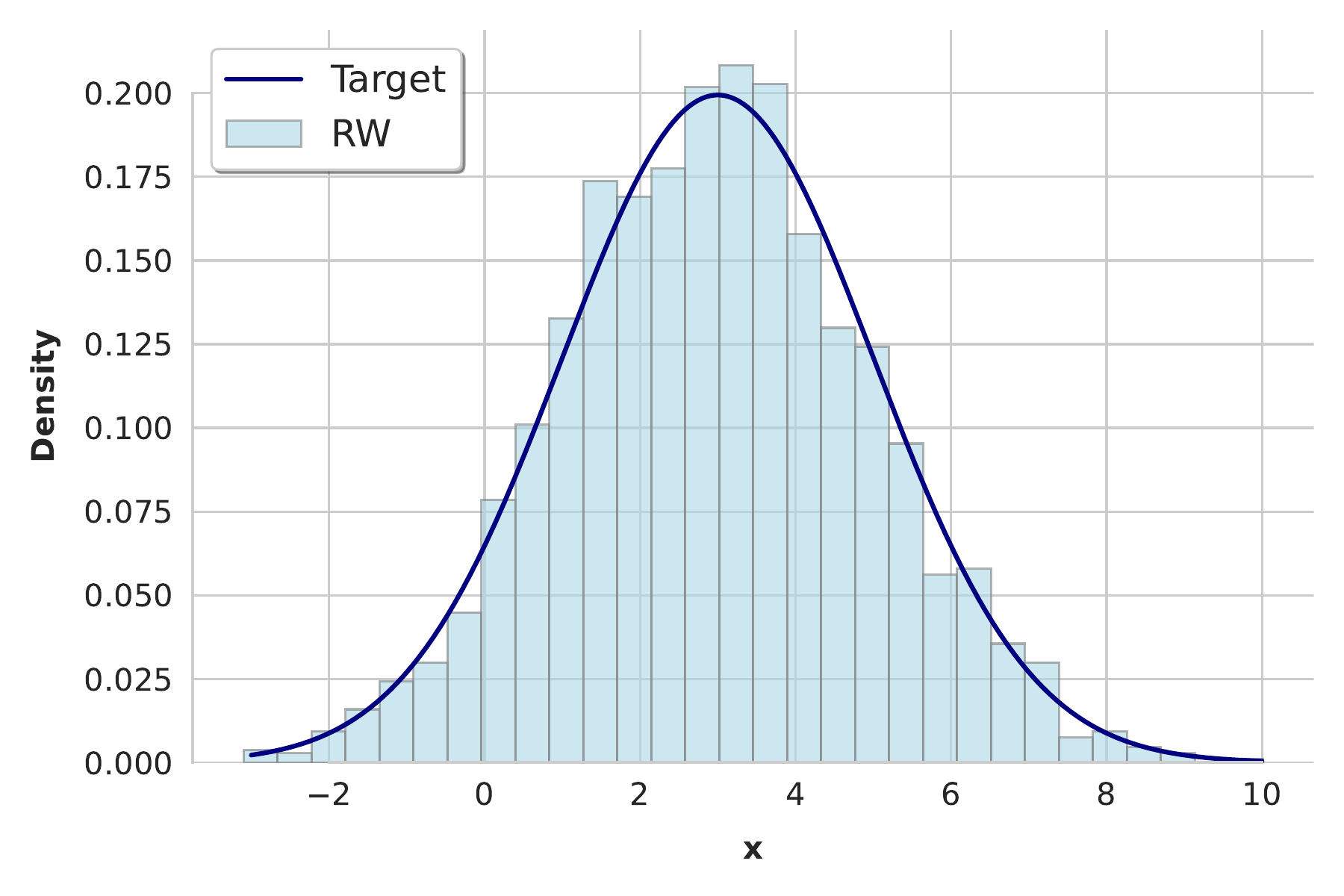}
        \caption{Standard Random Walk Sampling.}
        \label{fig:gaussian_mh}
    \end{subfigure}
    \hfill
    \begin{subfigure}[b]{0.32\textwidth}
        \includegraphics[width=\textwidth]{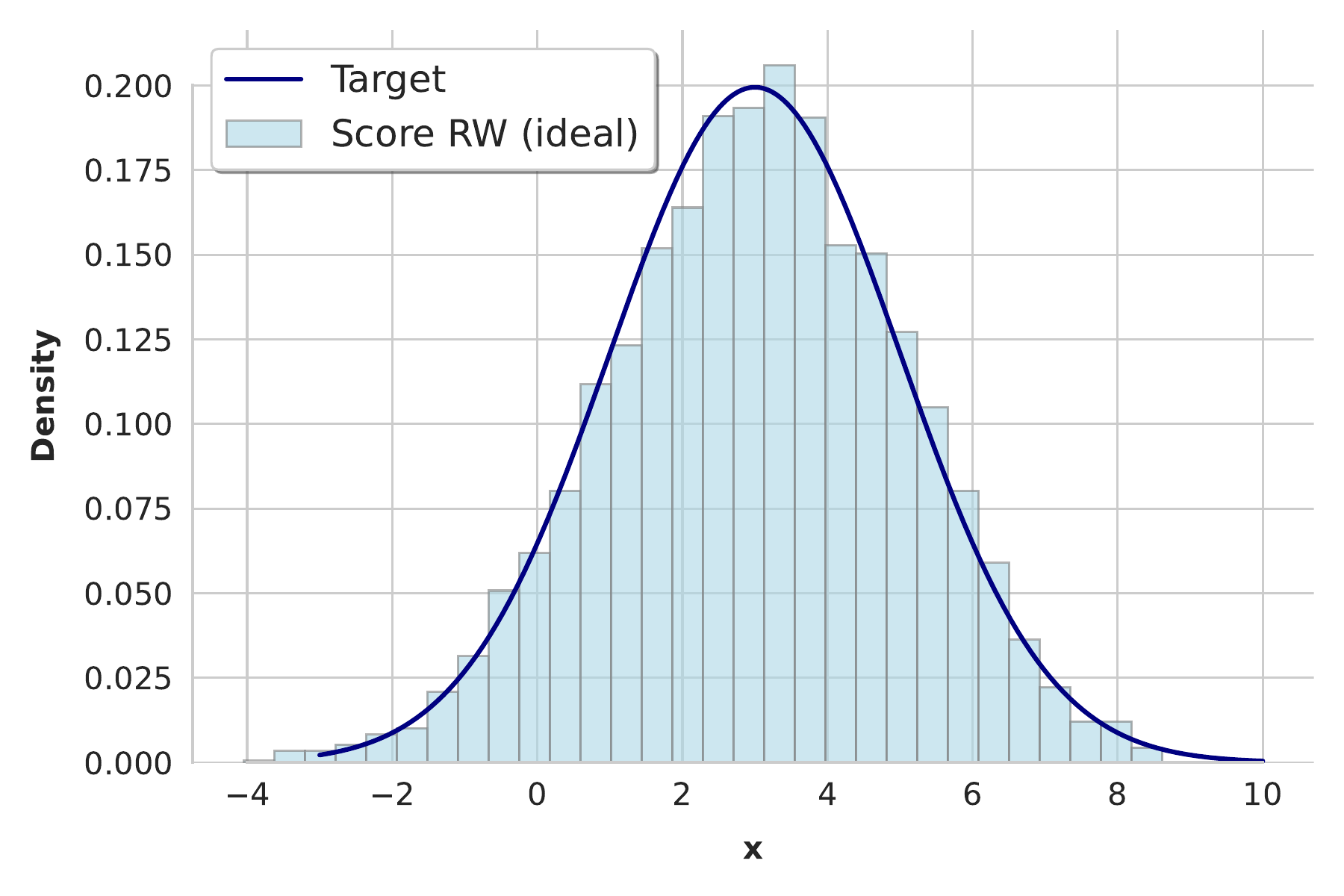}
        \caption{Score-based RW with true scores.}
        \label{fig:gaussian_accept_mh}
    \end{subfigure}
    \hfill
    \begin{subfigure}[b]{0.32\textwidth}
        \includegraphics[width=\textwidth]{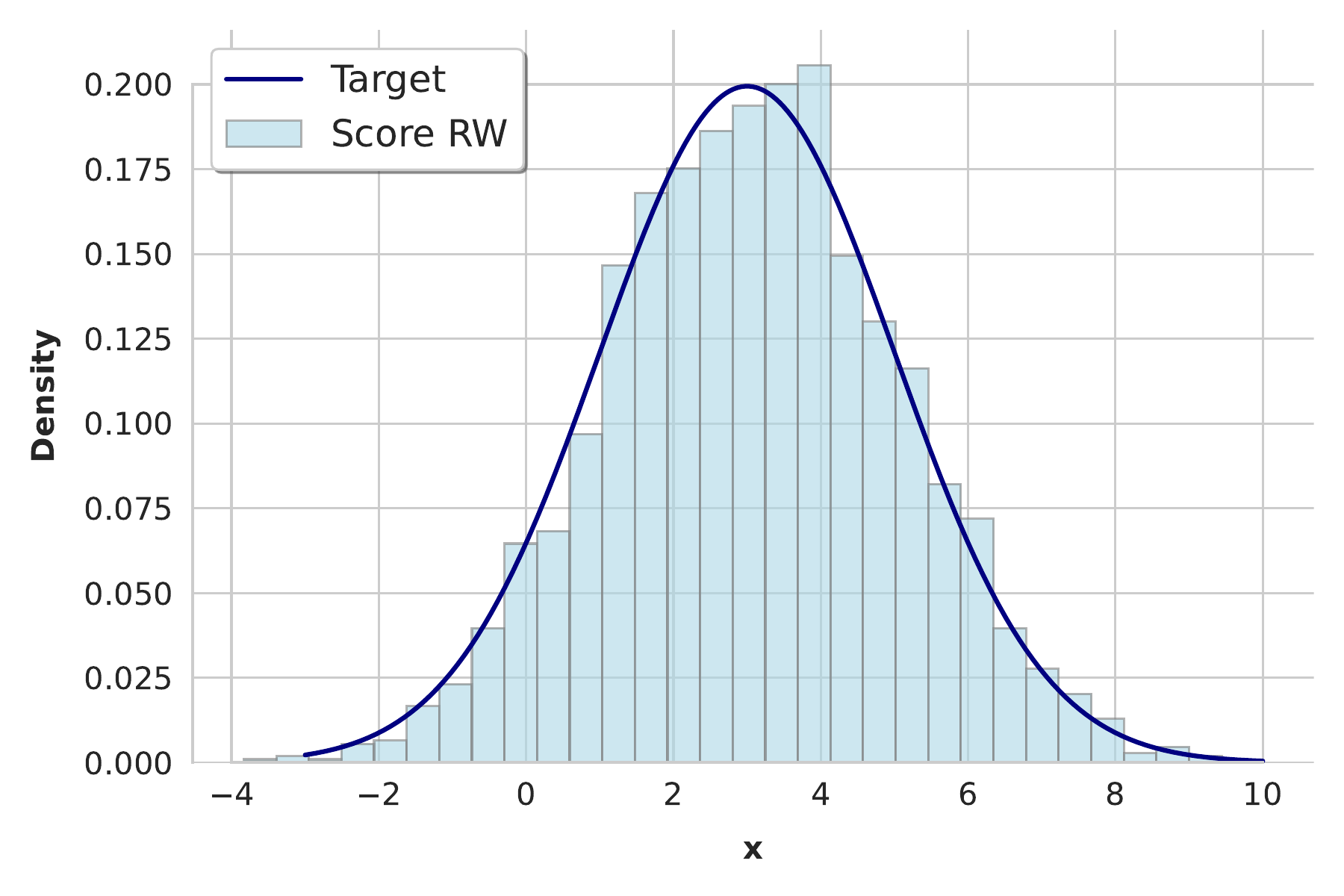}
        \caption{Score-based RW with learned score.}
        \label{fig:gaussian_accept_score_mh}
\end{subfigure}

\caption{Comparison of sampling methods from a Gaussian distribution $\mathcal{N}(3, 2)$ using three approaches: (a) Standard Random Walk Sampling with the acceptance function defined in~\eqref{eqn:ratio_acceptance}, (b) Score-based RW with an acceptance network trained using true scores, and (c) Score-based RW with an acceptance network trained using a learned score.}
\label{fig:gaussian_sampling_methods}
\end{figure*}

\begin{figure*}[ht]
    \centering
    \begin{subfigure}[b]{0.18\textwidth}
        \includegraphics[width=\textwidth]{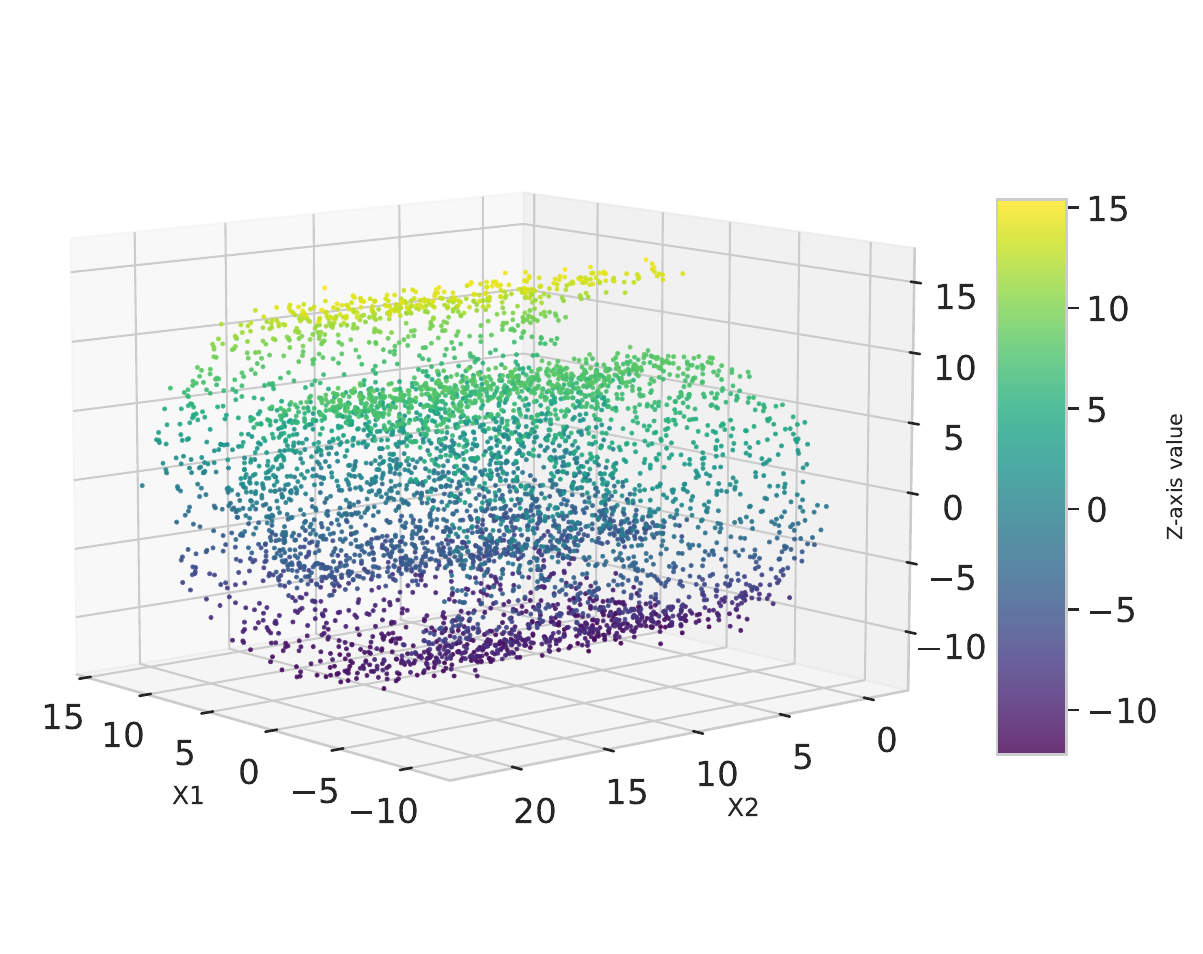}
        \caption{Original samples.}
    \end{subfigure}
    \hfill
    \begin{subfigure}[b]{0.18\textwidth}
        \includegraphics[width=\textwidth]{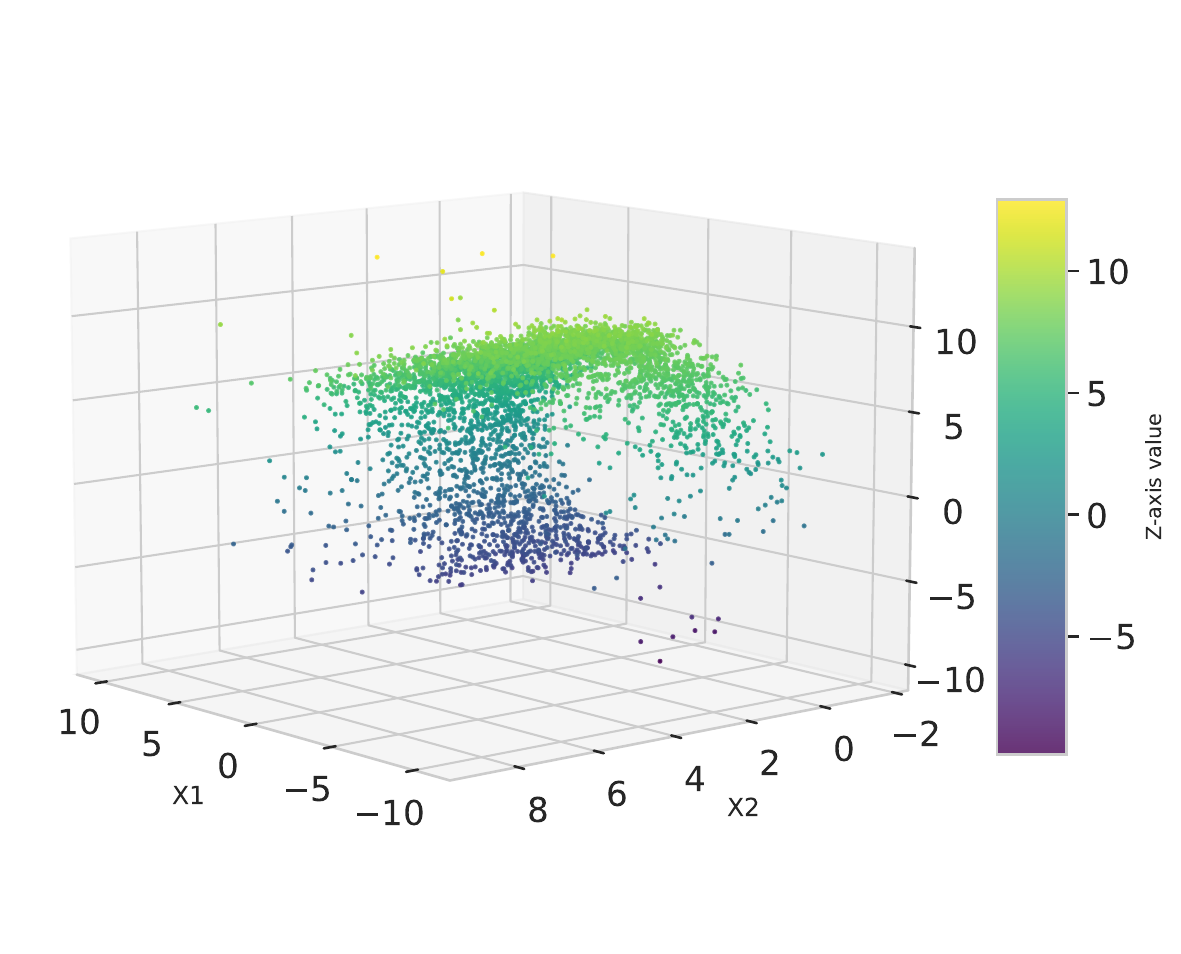}
        \caption{ULA.}
    \end{subfigure}
    \hfill
    \begin{subfigure}[b]{0.18\textwidth}
        \includegraphics[width=\textwidth]{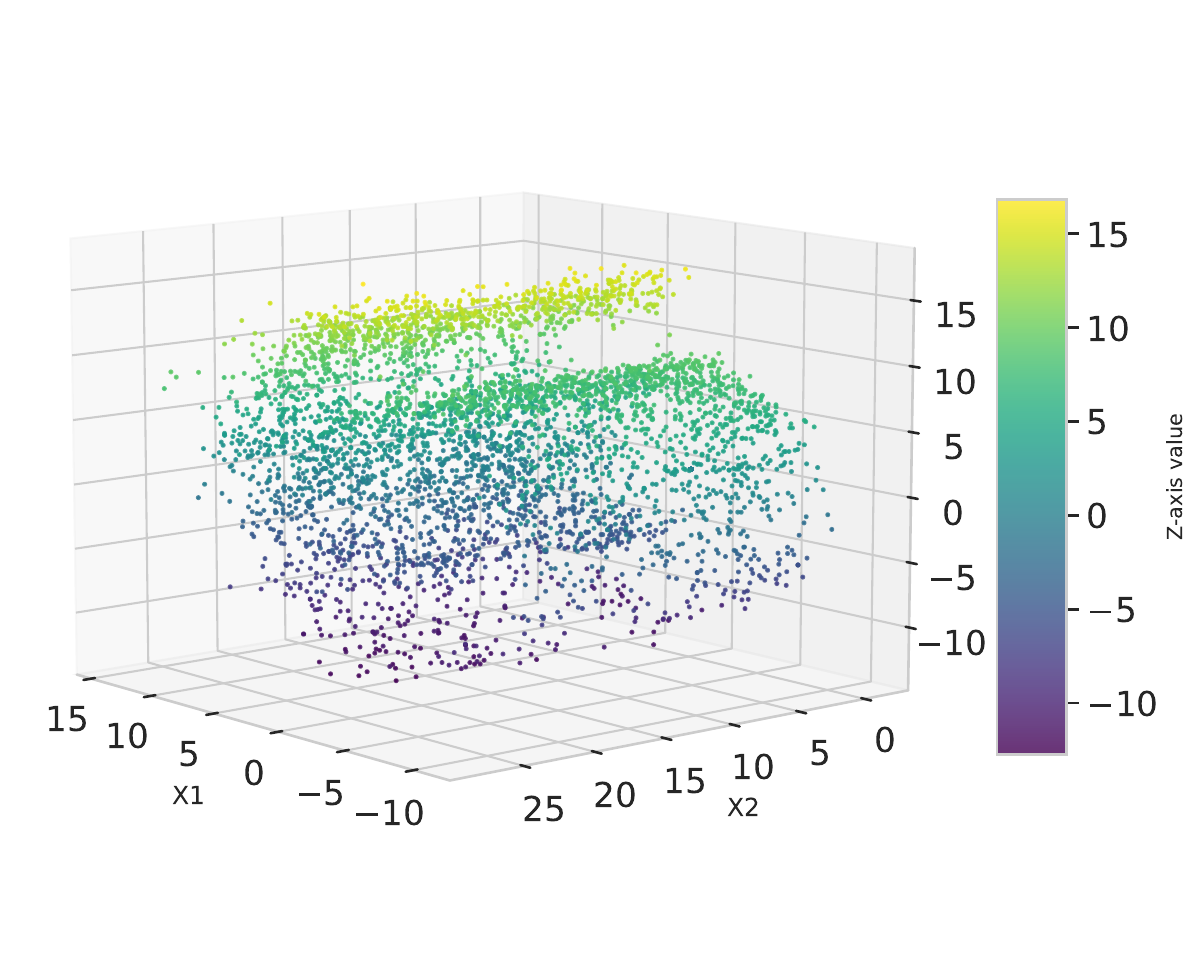}
        \caption{Score RW.}
    \end{subfigure}
    \hfill
    \begin{subfigure}[b]{0.18\textwidth}
        \includegraphics[width=\textwidth]{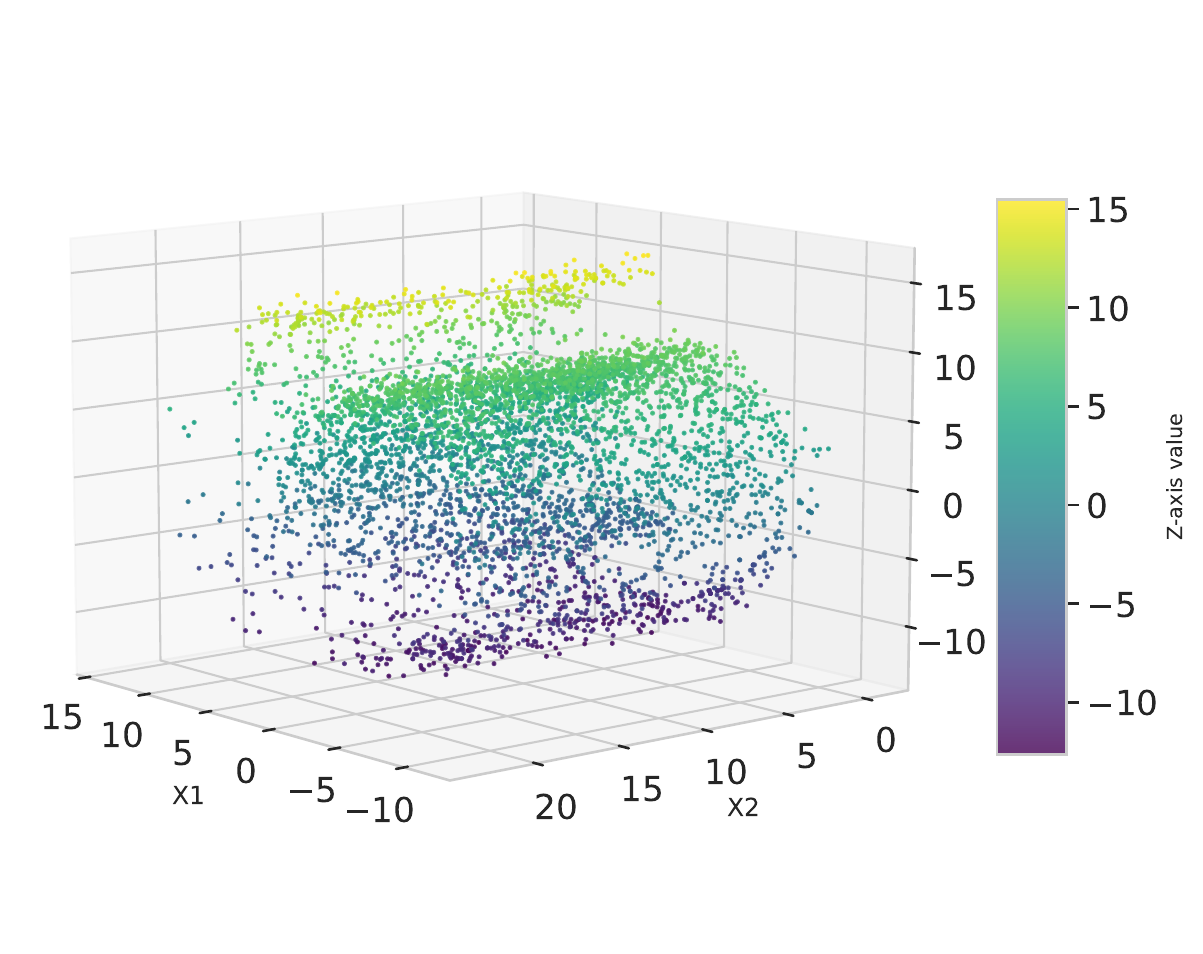}
        \caption{Score MALA.}
    \end{subfigure}
    \hfill
    \begin{subfigure}[b]{0.18\textwidth}
        \includegraphics[width=\textwidth]{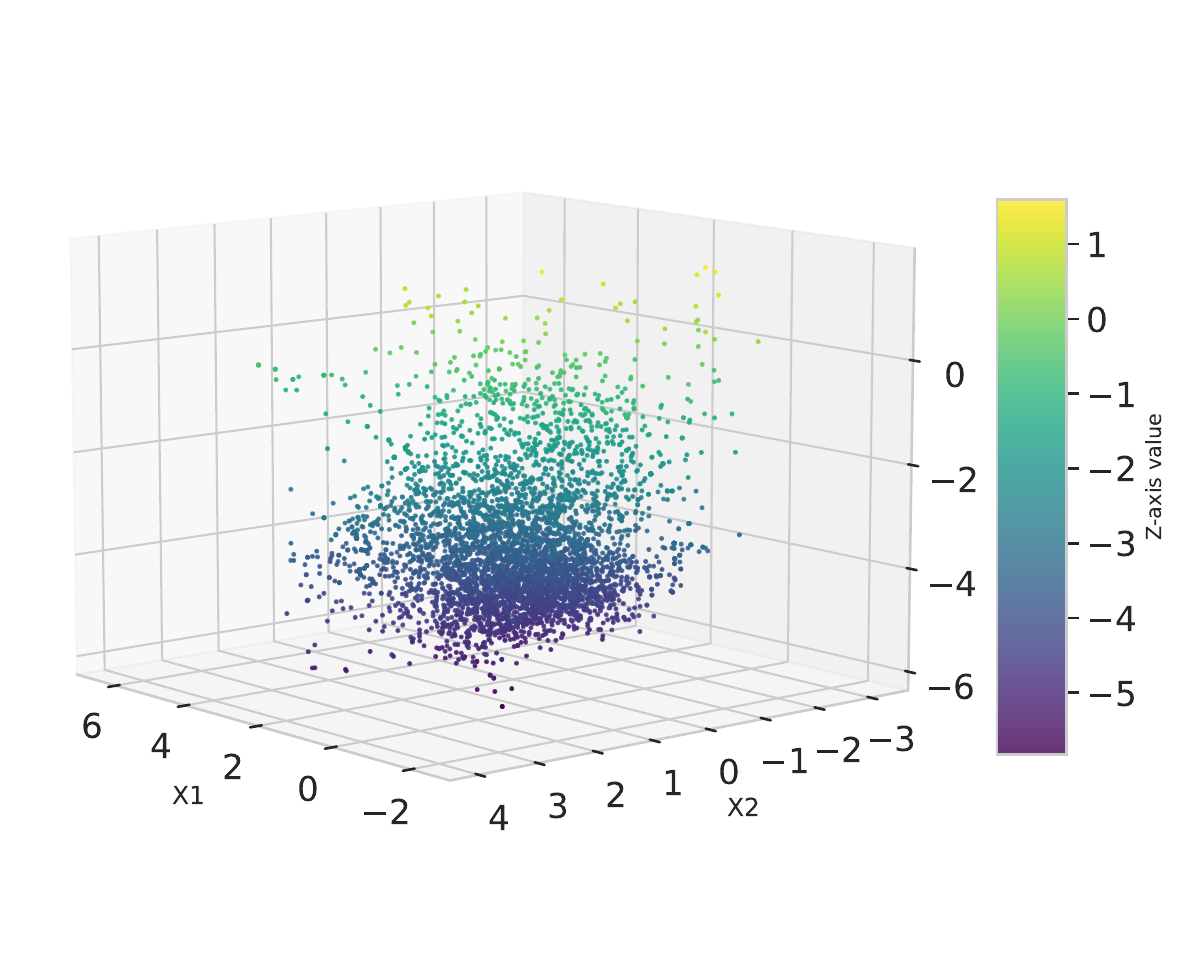}
        \caption{Score pCN.}
    \end{subfigure}
    \caption{Comparison of different methods on the Swiss Roll dataset.}
    \label{fig:roll_results}
\end{figure*}

\begin{figure*}
    \centering
    \includegraphics[width=0.95\linewidth]{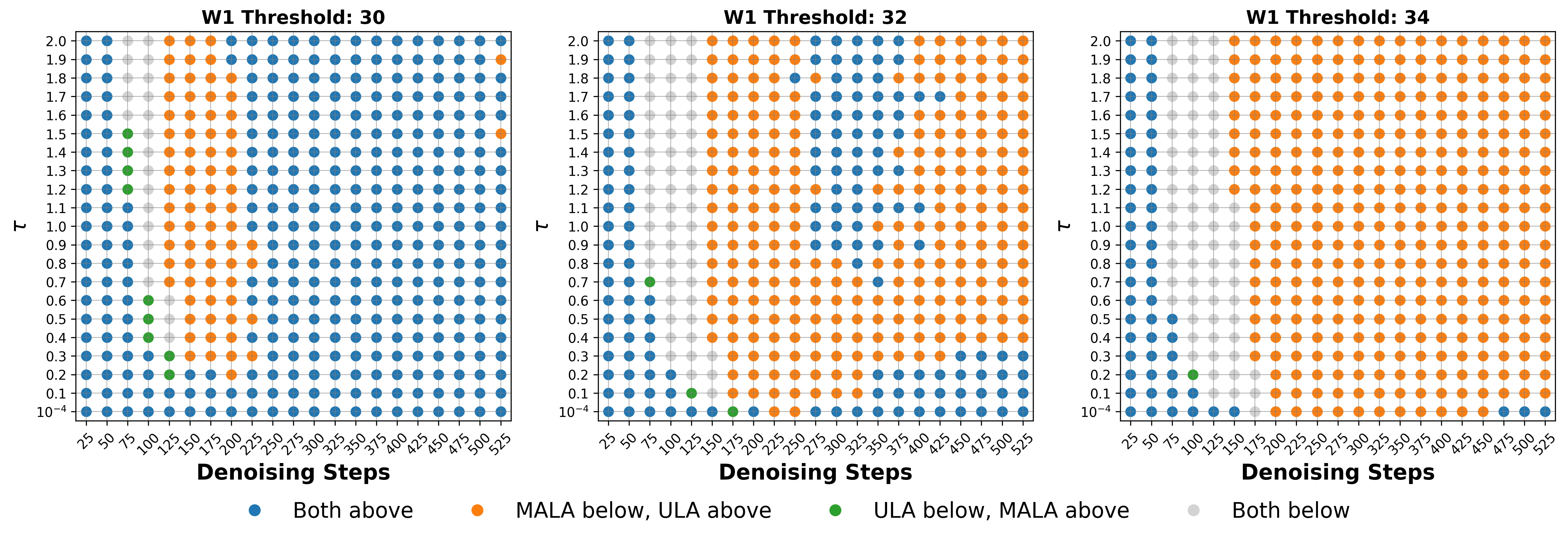}
    \caption{Comparison of Wasserstein-1 (W1) values for Annealed MALA and Annealed ULA across different $\tau$ and step size settings. The plots classify regions based on whether the W1 values for both methods exceed or fall below a given threshold. Each subplot corresponds to a different threshold value (\(30, 32, 34\)). Colors indicate classification: (i) both methods above the threshold (blue), (ii) MALA below and ULA above (orange), (iii) ULA below and MALA above (green), and (iv) both methods below the threshold (gray). The x-axis represents the number of denoising steps, while the y-axis corresponds to the adaptive step size parameter \( \tau \). We can see that annealed MALA exhibits a more robust behavior across different $\tau$ levels.}
    \label{fig:mala_robust}
\end{figure*}

In this section, we include additional empirical results for three datasets. First, we perform a sanity check using a synthetic dataset, demonstrating that training the acceptance network with a learned score achieves similar performance to training with the true score, as shown in Figure~\ref{fig:gaussian_sampling_methods}. Furthermore, we present results for generating the Swiss Roll dataset in Figure~\ref{fig:roll_results}. Finally, we provide additional results for the MNIST dataset, illustrating the robustness of Annealed MALA to the step size in Figure~\ref{fig:mala_robust}.

\end{document}